\documentclass[10pt,journal,compsoc]{IEEEtran} 

\usepackage{authblk} 
\usepackage[nocompress]{cite} 

\usepackage[pdftex]{graphicx}
\graphicspath{{./img/}}
\DeclareGraphicsExtensions{.pdf,.jpeg,.png}
\usepackage{amsmath} 
\usepackage{amssymb}
\usepackage{mdwmath} 
\usepackage{dsfont} 
\usepackage{mdwtab} 
\usepackage{array} 
\usepackage[caption=false,font=footnotesize,labelfont=sf,textfont=sf]{subfig} 
\usepackage{url}
\usepackage{xr-hyper}
\usepackage[colorlinks=true,backref=page]{hyperref}
\usepackage{threeparttable}
\usepackage{multicol}
\usepackage{multirow}
\usepackage{subfig}

\usepackage{cleveref} 
\crefname{figure}{Fig.}{Figs.}
\crefname{equation}{Eq.}{Eqs.}
\crefname{section}{Sec.}{Secs.}
\crefname{subsection}{Sec.}{Secs.}
\crefname{paragraph}{Sec.}{Secs.}
\crefname{table}{Tab.}{Tabs.}
\usepackage[table,xcdraw]{xcolor}

\usepackage{pdflscape} 
\usepackage{adjustbox} 

\makeatletter
\newcommand*{\addFileDependency}[1]{
  \typeout{(#1)}
  \@addtofilelist{#1}
  \IfFileExists{#1}{}{\typeout{No file #1.}}
}
\makeatother

\usepackage{tikz}
\usetikzlibrary{decorations.text}

\newcommand\copyrighttext{%
  \tiny \textcopyright 2023 IEEE. Personal use of this material is permitted. Permission from IEEE must be obtained for all other uses, in any current or future media, including reprinting/republishing this material for advertising or promotional purposes, creating new collective works, for resale or redistribution to servers or lists, or reuse of any copyrighted component of this work in other works. DOI: \href{https://doi.org/10.1109/TPAMI.2023.3243465}{10.1109/TPAMI.2023.3243465}}
\newcommand\copyrightnotice{%
\begin{tikzpicture}[remember picture,overlay]
\node[anchor=south,yshift=10pt] at (current page.south) {\fbox{\parbox{\dimexpr\textwidth-\fboxsep-\fboxrule\relax}{\copyrighttext}}};
\end{tikzpicture}%
}

\newcommand{\generaltable}{\textit{Table S1}~}

\makeatletter
\DeclareRobustCommand{\doublewidetilde}[1]{{%
\mathpalette\double@widetilde{#1}%
}}
\DeclareRobustCommand{\double@widetilde}[2]{%
\sbox\z@{$\m@th#1\widetilde{#2}$}%
\ht\z@=.95\ht\z@
\widetilde{\box\z@}%
}
\renewcommand\AB@affilsepx{\space\protect\Affilfont}
\makeatother

\usepackage{pdfrender}

\author[1,3]{Javier Selva}
\author[2,4]{Anders S. Johansen}
\author[1,2,3]{Sergio Escalera}
\author[2,4]{\linebreak Kamal Nasrollahi}
\author[2,5]{Thomas B. Moeslund}
\author[2,3]{Albert Clap\'{e}s}

\affil[1]{\normalsize Universitat de Barcelona}
\affil[2]{Aalborg University}
\affil[3]{Computer Vision Center}
\affil[4]{Milestone Systems}
\affil[5]{Pioneer Center for AI}
\affil[ ]{\linebreak \tt\small jselvaca21@alumnes.ub.edu, \{alcl,asjo,kn,tbm\}@create.aau.dk, sergio@maia.ub.es}

\begin{document}
\title{Video Transformers: A Survey}

\IEEEtitleabstractindextext{
\begin{abstract}
Transformer models have shown great success handling long-range interactions, making them a promising tool for modeling video. However, they lack inductive biases and scale quadratically with input length. These limitations are further exacerbated when dealing with the high dimensionality introduced by the temporal dimension. While there are surveys analyzing the advances of Transformers for vision, none focus on an in-depth analysis of video-specific designs. In this survey, we analyze the main contributions and trends of works leveraging Transformers to model video. Specifically, we delve into how videos are handled at the input level first. Then, we study the architectural changes made to deal with video more efficiently, reduce redundancy, re-introduce useful inductive biases, and capture long-term temporal dynamics. In addition, we provide an overview of different training regimes and explore effective self-supervised learning strategies for video. Finally, we conduct a performance comparison on the most common benchmark for Video Transformers (i.e., action classification), finding them to outperform 3D ConvNets even with less computational complexity.
\end{abstract}

}

\maketitle
\copyrightnotice
\vspace{-0.25cm}
\ifCLASSOPTIONcompsoc
  \IEEEraisesectionheading{\section{Introduction}\label{sec:introduction}}
\else
  \section{Introduction}
  \label{sec:introduction}
\fi

\IEEEPARstart{V}{ideo} is increasingly becoming a popular medium to convey audio-visual information. Video provides the visual appeal of images while introducing motion and deformations through the additional time dimension. As such, processing video data is partially akin to both images (continuous visual signals) and natural language processing (structured as a sequence). The video domain further introduces its own challenges, namely a large increase in dimensionality linked with a high level of information redundancy and the need to model motion dynamics. 

Transformers~\cite{vaswani2017attention} are a recent family of models, originally designed to replace recurrent layers in a machine translation setting. Its purpose was to remedy limitations of sequence modeling architectures by handling whole sequences at once (as opposed to RNNs, which are sequential in nature), allowing further parallelization. Besides, it removes the locality bias of traditional architectures, such as CNNs, and instead learns interactions of non-local contexts of the input. This lack of inductive biases makes Transformers very versatile, as seen by the quick adoption for modeling many data types~\cite{devlin2019bert,plizzari2021skeleton,gong21b_interspeech,dosovitskiy2021an,carion2020end}, including videos~\cite{bertasius2021spacetime,girdhar2019video,arnab2021vivit,liu2021swinvideo,zhu2020actbert,li2020hero}. The Transformer evolves input representations based on interactions among all sequence elements. These interactions are modulated through a pair-wise affinity function that weighs the contribution that every element should have on any other. The ability to model all-to-all relationships can be especially beneficial to understand motion cues, long-range temporal interactions, and dynamic appearance changes in video data. However, Transformers scale quadratically with sequence length $T$ (i.e., $\mathcal{O}(T^2)$, due to the pair-wise affinity computation) which is exacerbated by the high dimensionality of video. Furthermore, the lack of inductive biases makes Transformers require large amounts of data or several modifications to adapt to the highly redundant spatiotemporal structure of video. 

The recent surge in \textit{Video Transformer} (VT) works makes it convoluted to keep track of the latest advances and trends. Existing surveys focus on design choices for Transformers in general~\cite{lin2021survey}, NLP~\cite{surveyNLP}, images~\cite{surveyVisual4, surveyVisual3}, or efficient designs~\cite{surveyEfficient1, surveyEfficient3}. Given the sequential nature of video, as well as the large dimensionality and redundancy introduced by the temporal dimension, adopting image-based solutions or NLP-based designs for long-term modeling 
will not suffice. While other existing surveys include video, they are limited to superficial comments of a few VTs in the broader context of vision Transformers~\cite{surveyVision1, surveyVision2, yang2022transformers}, techniques to integrate visual data with other modalities~\cite{multimodalSurvey1,xu2022multimodal}, or video-language pre-training~\cite{multimodalSurvey2}. In this sense, they miss an in-depth analysis that properly captures the challenges of modeling raw image sequences or highly redundant spatiotemporal visual features through Transformers. 

In this survey, we comprehensively analyze advances and limitations of Transformers when considering the particularities of modeling video data. To do so, we review over 100 VT works and provide detailed taxonomies of the various design choices throughout the VT pipeline (namely input, architecture, and training). Finally, we extensively compare performance on the task of video classification based on self-reported results from the state-of-the-art on Kinetics 400~\cite{carreira2017quo} and Something-Something-v2~\cite{mahdisoltani2018effectiveness}.

The structure of the paper is as follows: \cref{sec:transformers} introduces the original Transformer; in~\cref{sec:input} we explore how videos are handled prior to the Transformer; \cref{sec:architecture} describes Transformer design adaptations to video; \cref{sec:training} investigates common training strategies; \cref{sec:application} discusses VTs performance on action classification; and in \cref{sec:discussion} we discuss the main trends, limitations, and future work. For an extensive list of all VT works reviewed in this survey, and details on how each section in this survey relates to a given work, see \generaltable in the supplementary.

\vspace{-0.25cm}
\section{The Transformer}
\label{sec:transformers}
\begin{figure}[t!]
    \centering
    \includegraphics[width=\linewidth]{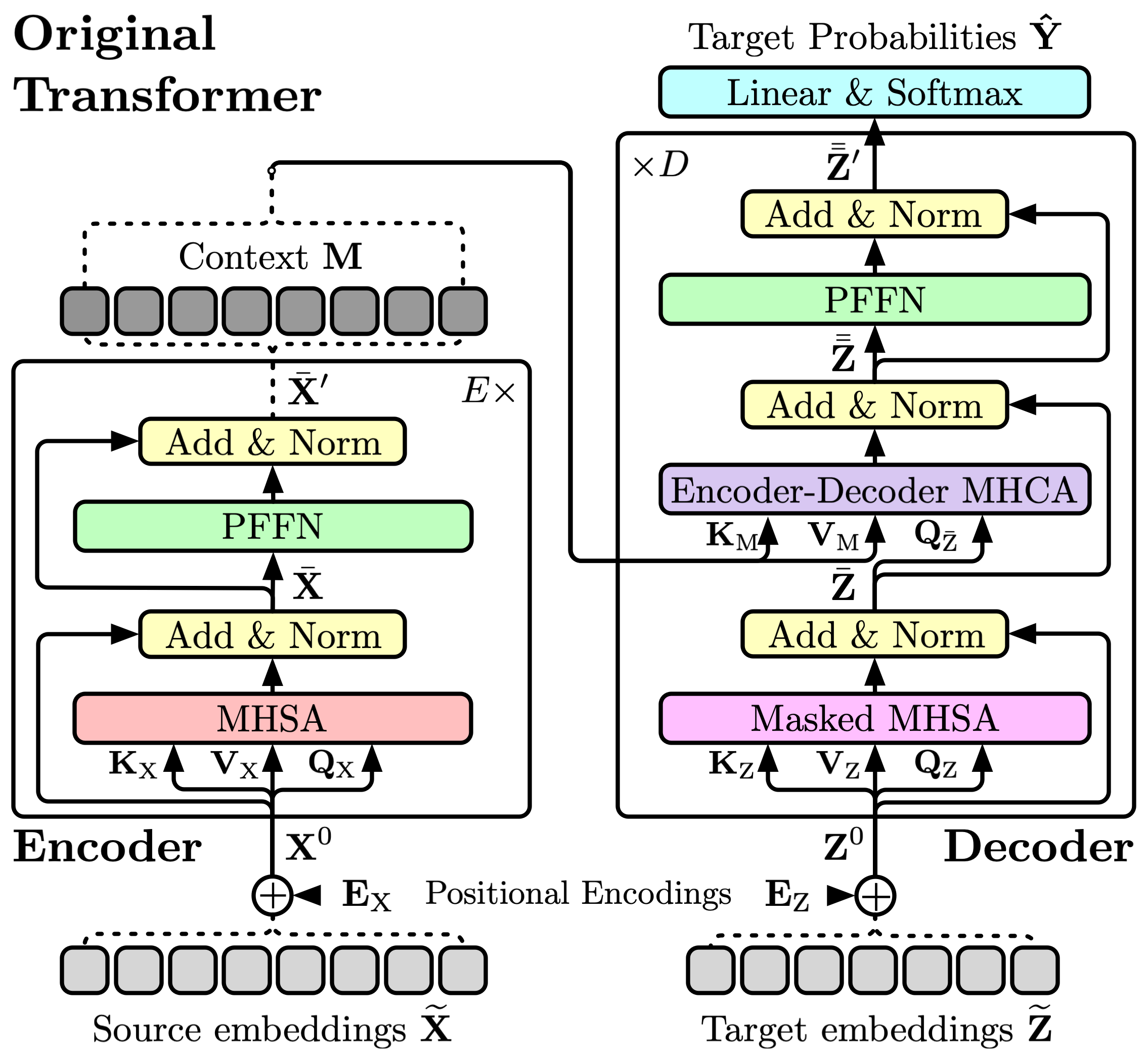}
    \caption{Visualization of the original Transformer proposed in \cite{vaswani2017attention}.}
    \label{fig:transformer}
\end{figure}

Originally proposed for language translation~\cite{vaswani2017attention}, the Transformer consists of two distinct modules: encoder and decoder, each composed of several stacked Transformer layers (see~\cref{fig:transformer}). The \textit{encoder} was designed to produce a representation of the source language sentence that is then attended by the \textit{decoder}, which will eventually translate it into the target language. We first introduce a few necessary concepts (input pre-processing and the self-attention operation) to then follow the flow of the Transformer while explaining its components and functioning.

\noindent\textbf{Input pre-processing: tokenization, linear embedding, and positional encodings}. The \textit{tokenization} converts the input source and target language sentences into sequences of words (or subwords), namely ``tokens''. Let $\doublewidetilde{\mathbf{X}} = (\doublewidetilde{\mathbf{x}}_1, \ldots, \doublewidetilde{\mathbf{x}}_{N_\mathrm{x}})$ and $\doublewidetilde{\mathbf{Z}} = (\doublewidetilde{\mathbf{z}}_1, \ldots, \doublewidetilde{\mathbf{z}}_{N_\mathrm{z}})$ be, respectively, the source and target sequences of one-hot encoded tokens over their respective word vocabularies $\mathcal{X}$ and $\mathcal{Z}$ (i.e., $\doublewidetilde{\mathbf{x}} \in \mathbb{R}^\mathcal{|X|}$ and $\doublewidetilde{\mathbf{z}} \in \mathbb{R}^\mathcal{|Z|}$). Then, \textit{linear embedding} is simply the step of projecting one-hots to a continuous embedding space via a learned linear transformation: $f_\mathcal{X}: \mathbb{R}^{|\mathcal{X}|} \mapsto \mathbb{R}^{d_m}$ (analogously $f_\mathcal{Z}$), where $d_m$ will be the dimensionality handled internally by the Transformer. This way, we obtain the source embeddings $\widetilde{\mathbf{X}} = (f_\mathcal{X}(\doublewidetilde{\mathbf{x}}_1), \ldots, f_\mathcal{X}(\doublewidetilde{\mathbf{x}}_{N_\mathrm{x}}))$ and target embeddings $\widetilde{\mathbf{Z}} = (f_\mathcal{Z}(\doublewidetilde{\mathbf{z}}_1), \ldots, f_\mathcal{Z}(\doublewidetilde{\mathbf{z}}_{N_\mathrm{z}}))$. Finally, \textit{positional encodings} are added to signal the position of the tokens in the sequence to the later (otherwise permutation invariant) attention operations. Defined using a set of (non-learnable) sinusoidal encodings (see~\cite{vaswani2017attention} for details), these are added to the source/target embeddings before being input to encoder/decoder (as depicted in~\cref{fig:transformer}): $\mathbf{X}^0 = (\widetilde{\mathbf{x}}_1 + \mathbf{e}_1^{\mathrm{x}}, \ldots, \widetilde{\mathbf{x}}_{N_\mathrm{x}} + \mathbf{e}_{N_\mathrm{x}}^{\mathrm{x}})$ and $\mathbf{Z}^0 = (\widetilde{\mathbf{z}}_1 + \mathbf{e}_1^{\mathrm{z}}, \ldots, \widetilde{\mathbf{z}}_{N_\mathrm{z}} + \mathbf{e}_{N_\mathrm{z}}^{\mathrm{z}})$, where $\mathbf{e}_{\cdot}^{\mathrm{x}}, \mathbf{e}_{\cdot}^{\mathrm{z}} \in \mathbb{R}^{d_\mathrm{m}}$.

\noindent\textbf{Self-attention (SA)}. It is the core operation of the Transformer. Given an arbitrary sequence of token embeddings $\mathbf{X} \in \mathbb{R}^{N_\mathrm{x} \times d_\mathrm{m}}$ (e.g., $\mathbf{X}^0$), it augments (contextualizes) each of the embeddings $\mathbf{x}_i \in \mathbb{R}^{d_\mathrm{m}}$ with information from the rest of embeddings. For that, the embeddings in $\mathbf{X}$ are linearly mapped to the embedding spaces of \textit{queries} $\mathbf{Q} = \mathbf{X}\mathbf{W}_Q \in \mathbb{R}^{N_\mathrm{x} \times d_\mathrm{k}}$, \textit{keys} $\mathbf{K} = \mathbf{X}\mathbf{W}_\mathrm{K} \in \mathbb{R}^{N_\mathrm{x} \times d_\mathrm{k}}$, and \textit{values} $\mathbf{V} = \mathbf{X}\mathbf{W}_\mathrm{V} \in \mathbb{R}^{N_\mathrm{x} \times d_\mathrm{k}}$, where $\mathbf{W}_\mathrm{Q}, \mathbf{W}_\mathrm{K} \in \mathbb{R}^{d_m \times d_k}$, $\mathbf{W}_\mathrm{V} \in \mathbb{R}^{d_m \times d_v}$, and typically $d_\mathrm{k}, d_\mathrm{v} <= d_m$. Then, self-attention can be computed as follows:

\begin{equation}
    \mathrm{Att}(\mathbf{Q},\mathbf{K},\mathbf{V}) = \mathrm{Softmax}\left( \frac{\mathbf{Q}\mathbf{K}^\top}{\sqrt{d_\mathrm{k}}}\right)\mathbf{V}.
    \label{eq:self-attention}
\end{equation}

The dot-product $\mathbf{Q}\mathbf{K}^\top \in \mathbb{R}^{N_\mathrm{x} \times N_\mathrm{x}}$ is a measure of similarity. Intuitively, the larger the similarity between $\mathbf{q}_i \in \mathbf{Q}$ and $\mathbf{k}_j \in \mathbf{K}$ the more relevant the information embedded in $\mathbf{x}_j$ is for $\mathbf{x}_i$. However, this aggregation is not done in the space of $\mathbf{X}$, but in the one of the values. By applying Softmax with temperature $\sqrt{d_k}$, we come up with normalized similarities (the self-attention matrix) that weigh how much each of the values $\mathbf{v}_j$ contributes to the output representation of every other $\mathbf{v}_i$.

\noindent\textbf{Encoder module}. It consists of $E$ layers, each including \textit{Multi-Head Self-Attention} (MHSA) and \textit{Position-wise Feed-Forward Network} (PFFN) sub-layers. The MHSA sub-layer performs self-attention through multiple separate heads that map $\mathbf{X}$ to $h$ different representation sub-spaces (i.e., $\{(\mathbf{Q}_i, \mathbf{K}_i, \mathbf{V}_i) \,|\, 1 \leq i \leq h\}$). The outputs of the heads are concatenated and mapped back to a $d_\mathrm{m}$-dimensional space with another linear transformation $\mathbf{W}_\mathrm{O} \in \mathbb{R}^{(h\cdot d_\mathrm{v}) \times d_\mathrm{m}}$:

\begin{equation}
\begin{aligned}
    \mathrm{MHSA}(\mathbf{X}) = \mathrm{Concat}(\mathbf{H}_1,...,\mathbf{H}_h)\mathbf{W}_\mathrm{O}, \\
    \textrm{where } \mathbf{H}_i = \mathrm{Att}(\mathbf{Q}_i,\mathbf{K}_i,\mathbf{V}_i),
\end{aligned}
\label{eq:multihead}
\end{equation}

where $\mathbf{H}_i \in \mathbb{R}^{N_\mathrm{x} \times d_\mathrm{v}}$ is the output of the $i$\textsuperscript{th} head, and $\mathbf{Q}_i$, $\mathbf{K}_i$, and $\mathbf{V}_i$ are computed with their own associated embedding matrices (i.e., $\mathbf{W}_{\mathrm{Q}_i} \in \mathbb{R}^{d_\mathrm{m} \times d_\mathrm{k}}$, $\mathbf{W}_{\mathrm{K}_i} \in \mathbb{R}^{d_\mathrm{m} \times d_\mathrm{k}}$, and $\mathbf{W}_{\mathrm{V}_i} \in \mathbb{R}^{d_\mathrm{m} \times d_\mathrm{v}}$ with $d_\mathrm{k} = d_\mathrm{v} = d_\mathrm{m} / h$). ``Add + Norm'' is then applied to come up with $\bar{\mathbf{X}} = \mathrm{LN}(\mathbf{X} + \mathrm{MHSA}(\mathbf{X}))$, where $\bar{\mathbf{X}} \in \mathbb{R}^{N_X \times d_\mathrm{m}}$. After this, the following $\mathrm{PFFN}$ sub-layer further refines each embedding in $\bar{\mathbf{X}}$ individually (point-wise). This sub-layer is composed of two linear layers and $\mathrm{ReLU}$ activation function: $\mathrm{PFFN}(\bar{\mathbf{X}}) = \mathrm{ReLU}(\bar{\mathbf{X}}\mathbf{W}_{\mathrm{F1}})\mathbf{W}_{\mathrm{F2}}$, where $\mathbf{W}_\mathrm{F1} \in \mathbb{R}^{d_\mathrm{m} \times (4*d_\mathrm{m})}$ and $\mathbf{W}_\mathrm{F2} \in \mathbb{R}^{(4*d_\mathrm{m}) \times d_\mathrm{m}}$. Note, $\mathbf{W}_{\cdot}$ are independent for each layer, but we omit those indices for ease of notation. By applying this, $\bar{\mathbf{X}}' = \mathrm{LN}(\bar{\mathbf{X}} + \mathrm{PFFN}(\bar{\mathbf{X}}))$. 

\noindent\textbf{Decoder module}. Consisting of $D$ layers and fed with $\mathbf{Z}^0$, it substitutes MHSA with two other sub-layers. The first one, \textit{Masked Multi-Head Self-Attention} (Masked MHSA), modifies $\mathrm{Att}$ in \cref{eq:self-attention} to include a mask, $\mathbf{B} = (b_{ij}), 1 \leq i,j \leq N_\mathrm{z}$, impeding the access to certain tokens. This is added to the result of the dot-product in the numerator (and before the Softmax), as follows: $\mathbf{Q}\mathbf{K}^\top + \mathbf{B} \in \mathbb{R}^{N_\mathrm{Z} \times N_\mathrm{Z}}$, where $b_{ij} = -\infty$ iff $i < j$ (otherwise $b_{ij} = 0$). This draws attention values for the masked attention pairs to 0 when taking exponents in the Softmax. As we will see, such masking is crucial to define the auto-regressive behavior of the decoder module (avoiding tokens to attend to other tokens later in the sequence). The produced $\bar{\mathbf{Z}}$ is now passed to the \textit{Encoder-Decoder Multi-Head Cross-Attention} (MHCA) sub-layer, which leverages the memory/context produced by the encoder, namely $\mathbf{M}$ (i.e., $\bar{\mathbf{X}}'$ at encoder's $E$\textsuperscript{th} layer), into $\bar{\mathbf{Z}}$ as follows: $\mathrm{MHCA}(\bar{\mathbf{Z}}, \mathbf{M}) = \mathrm{Concat}(\textbf{J}_1, \ldots, \textbf{J}_h)\textbf{U}_\mathrm{P}$, where $\mathbf{J}_i = \mathrm{Att}(\bar{\mathbf{Z}}\mathbf{U}_{\mathrm{Q}_i}, \mathbf{M}\mathbf{U}_{\mathrm{K}_i},\mathbf{M}\mathbf{U}_{\mathrm{V}_i}) \in \mathbb{R}^{N_\mathrm{Z} \times d_v}$ is the output of the $i$\textsuperscript{th} cross-attention head, $\mathbf{U}_{\mathrm{Q}_i} \in \mathbb{R}^{N_\mathrm{Z} \times d_k}$, $\mathbf{U}_{\mathrm{K}_i} \in \mathbb{R}^{N_\mathrm{x} \times d_k}$, $\mathbf{U}_{\mathrm{V}_i} \in \mathbb{R}^{N_\mathrm{x} \times d_v}$, and $\mathbf{U}_\mathrm{P} \in\mathbb{R}^{(h\cdot d_\mathrm{v}) \times d_\mathrm{m}}$. Then, $\bar{\bar{\mathbf{Z}}} = \mathrm{LN}(\bar{\mathbf{Z}} + \mathrm{MHCA}(\bar{\mathbf{Z}}, \mathbf{M}))$. The remaining PFFN sub-layer, which is no different from the one in encoder layers, is used to produce $\bar{\bar{\mathbf{Z}}}' = \mathrm{LN}(\bar{\bar{\mathbf{Z}}} + \mathrm{PFFN}(\bar{\bar{\mathbf{Z}}}))$. Finally, in the $D$\textsuperscript{th} layer, the embeddings from the PFFN are each sent through a linear layer followed by softmax to generate the output probabilities over the words in the target vocabulary~$\mathcal{Z}$, i.e., $\hat{\mathbf{Y}} \in \mathbb{R}^{N_\mathrm{z} \times |\mathcal{Z}|}$. 

\noindent\textbf{Current Transformer trends adopted for video}. Many variations to the Transformer have become common in vision and, particularly, video. First, the use of \textit{special tokens} such as \texttt{[CLS]} (class) or \texttt{[MSK]} (mask) tokens. In video, these are parameters initialized at random and adapted during the optimization process based on the learning objective. \texttt{[CLS]} is used to condense (into a vector representation) information from the rest of token embeddings in a sequence (representing spatiotemporal patches from the video~\cite{arnab2021vivit}), and suited for high-level tasks (such as classifying the sequence globally). Using input token embeddings instead of \texttt{[CLS]} may cause the model to be biased towards them~\cite{wang2018non}. Conversely, \texttt{[MSK]} is used to replace input embeddings and signal the Transformer to reconstruct those guided by the loss and based on the remaining tokens. This forces the Transformer to learn context from the tokens and how these relate to the masked ones. Conceived for language representation learning~\cite{devlin2019bert}, this has been adopted also for video representation learning~\cite{wei2022masked, tong2022videomae}. 

Second, \textit{deviations from the canonical encoder-decoder}: encoder-only or decoder-only Transformer architectures. Encoder-only are suited to produce fixed-size outputs, i.e., augmentations of the input embeddings that can be used for more granular tasks (e.g., per-frame classification) or, when used together with \texttt{[CLS]}, to come up with a global representation (e.g., sequence-level classification). For instance, \cite{arnab2021vivit, bertasius2021spacetime, fan2021multiscale} adopted an encoder-only architecture (along with the inclusion of \texttt{[CLS]}) for video classification following~\cite{dosovitskiy2021an}. Instead, decoder-only alternatives enable auto-regressive tasks if the size of the output cannot be determined a priori just by knowing the input size (e.g., to predict a series of temporal action detections). Initially proposed by \cite{radford2018improving} in NLP, these have been also followed in the context of video in~\cite{Miech_2021_CVPR, Tan_2021_ICCV, Zhang_2021_CVPR}. Other trends originated in other fields have been followed: swapping the order of the residual connection and layer normalization~\cite{arnab2021vivit, fan2021multiscale}, although no clear general advantage of one over the other has been empirically shown yet; or replacing ReLU in the PFFN by GeLU~\cite{ging2020coot,bertasius2021spacetime,Wang_2021_ICCV,kalfaoglu2020late} following~\cite{devlin2019bert}, with only~\cite{ging2020coot} ablating this decision (finding out that GeLU was slightly outperforming ReLU on their task/data).

\noindent\textbf{Transformer limitations}. Transformers have two key limitations: first, given the pair-wise affinity computation in~\cref{eq:self-attention}, they exhibit \textit{quadratic complexity} ($\mathcal{O}(N^2)$), which will be especially problematic for video. In \cref{sec:efficient}, we will explore some works alleviating this issue by reducing the scope of the SA operation. The second limitation is the lack of \textit{inductive biases}. This is a double-edged sword, allowing for a general-purpose architecture that can handle any modality but severely complicates the learning process. While this can be solved through large quantities of data~\cite{dosovitskiy2021an}, this further adds to the computational costs of training Transformers. Throughout the three following sections, we will explore various approaches (transversal to the whole VT pipeline) to solve this issue.

\vspace{-0.2cm}
\section{Input pre-processing}
\label{sec:input}
\begin{figure}[ht!]
    \includegraphics[width=.49\textwidth]{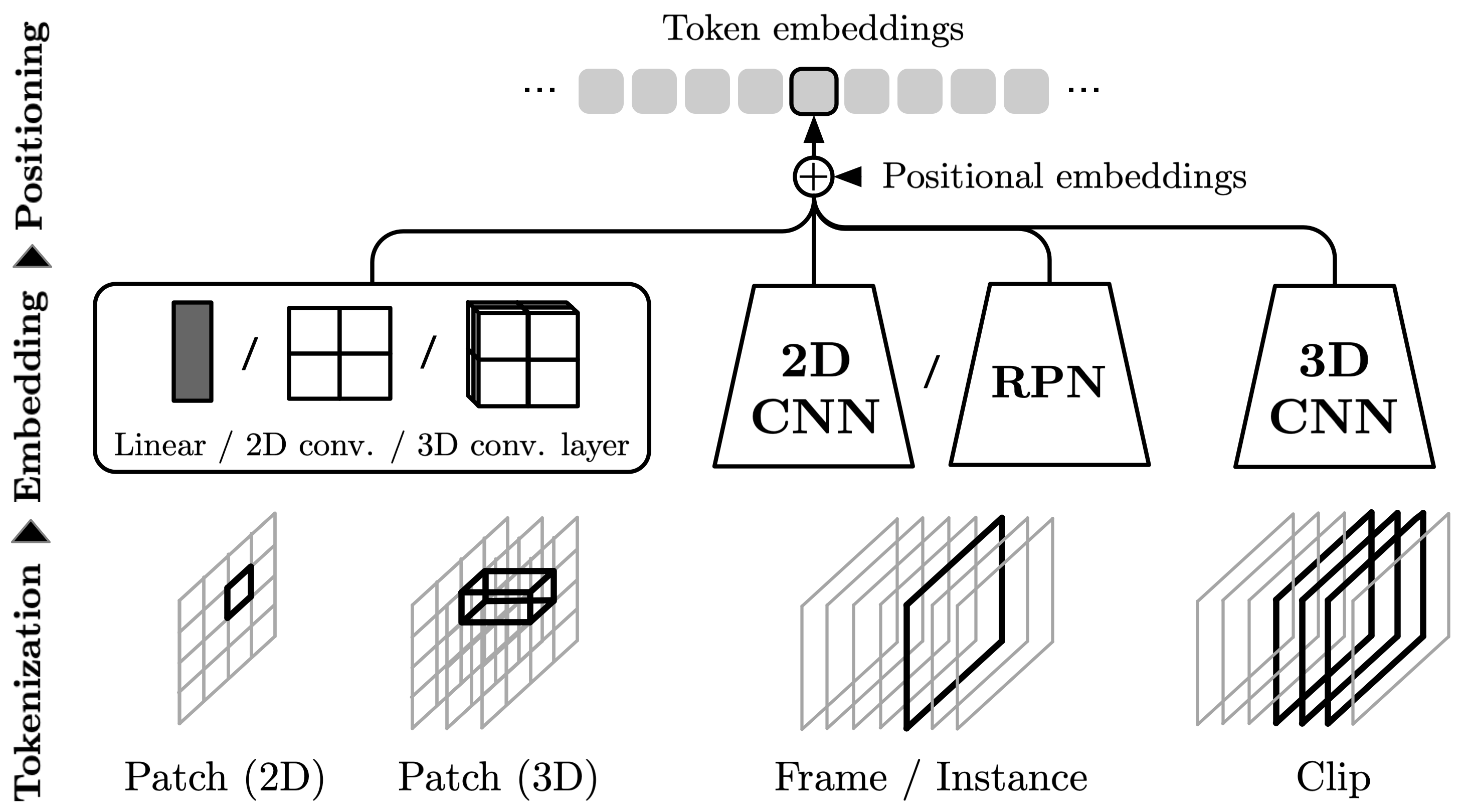}
    \vspace{-0.3cm}
    \caption{Overview of the input pre-processing step, showing tokenization and embedding strategies, as well as positioning (inclusion of positional information).}
    \label{fig:pipeline_b}
\end{figure}

Here, we review how video is processed before being input to the Transformer. This involves tokenization, embedding, and positioning (see \cref{fig:pipeline_b}). However, in the context of video, embedding often comes before tokenization: a separate network embeds the raw data into a continuous and compact representation, which can be used directly as a token or be further tokenized into more atomic units.

\vspace{-0.15cm}
\subsection{Embedding} 
\label{sec:embeddings}
In order to embed video, we find VTs following two main trends: \textit{embedding networks} or \textit{minimal embeddings}. The key difference between the two is size: while minimal embeddings are generally limited to single linear layers, large embedding networks are instantiated as full CNN architectures. Furthermore, while minimal embeddings follow the classic tokenization-then-embedding approach, full embedding networks can be used to embed full input sequences for later tokenization. In the context of video, embedding layers also function as a crucial dimensionality reduction mechanism.

\noindent\textbf{Embedding network}. Leveraging an embedding network (such as a CNN), can potentially ease the learning of the Transformer by providing strong initial features thanks to locality inductive biases. We can roughly categorize the choice of embedding network by the types of relationships they encode into spatial and spatiotemporal. Within \textit{spatial embeddings}, we find 2D CNN networks, typically ResNet variants~\cite{bb_resnet, bb_resnext}, pre-trained on large image corpora (most commonly ImageNet~\cite{imagenet_cvpr09,ridnik2021imagenetk}) to learn general filters that can extract meaningful representations of individual frames. This has been shown to work effectively in the context of video~\cite{lei2020mart,purwanto2019extreme,heo2021deepfake, gu2020pyramid,kondo2020lapformer,johnston2020self,li2020bridging, Liu_2021_ICCV}. However, 2D convolutions lack the ability to model temporal information. For this reason, we also find the use of \textit{spatiotemporal embedding} networks (e.g., in~\cite{wang2020attentionnas,purwanto2019extreme,contrastive2019chen, girdhar2019video,li2020bridging}). These are generally instantiated as 3D CNNs (such as I3D~\cite{bb_i3d} and S3D~\cite{bb_s3d}), commonly pre-trained on large video datasets such as Kinetics~\cite{carreira2017quo,carreira2019short} or HowTo100M~\cite{miech2019howto100m} to produce features involving temporal relationships. Alternatively, LSTMs~\cite{liu2019learning} or a hybrid ConvLSTM~\cite{NIPS2015_07563a3f,wang2021spatiotemporal,Weng_2021_ICCV}, can be leveraged to embed local temporal information. While spatial embedding networks are limited to per-token spatial interactions, spatiotemporal counterparts help provide initial locally-based temporal interactions.

\noindent\textbf{Minimal embeddings}. Inspired by the success of ViT~\cite{dosovitskiy2021an}, some works~\cite{arnab2021vivit, Liu_2021_ICCV, bertasius2021spacetime,Yu_2021_ICCV,jaegle2021perceiver, dosovitskiy2021an} omit deep embedding networks and subdivide the input (i.e., tokenize) and then perform embedding with only a few linear projections or convolutions. In this sense, they are guaranteed to not share information between tokens, leaving the learning of interactions between them entirely to the Transformer. Empirical studies like~\cite{bertasius2021spacetime,jaegle2021perceiver}, show that \textit{stand-alone Transformers} (i.e., without complex CNN embedding networks) are as performant as CNN counterparts, although the resulting model becomes data-hungry and computationally expensive. Given that, training and deploying VTs with minimal embeddings may benefit from architectural modifications inducing necessary biases (see~\cref{sec:architecture}). 

\vspace{-0.15cm}
\subsection{Tokenization} \label{sec:tokenization}
When dividing a video into smaller tokens to form the input sequence to the Transformer, we find several categories depending on the token input receptive field (i.e., the extent of the original input covered by a given token before being processed by the Transformer). We distinguish between patch, instance, frame, and clip tokenization (see~\cref{fig:pipeline_b}). 

\noindent\textbf{Patch-wise tokenization}. Most VTs follow ViT~\cite{dosovitskiy2021an} and employ a 2D-based patch tokenization~\cite{li2021vidtr,bertasius2021spacetime,Yu_2021_ICCV, zeng2020learning}, dividing the input video frames into regions of fixed spatial size~\cite{bertasius2021spacetime, Yu_2021_ICCV, li2021vidtr} or even multi-scale patch sizes~\cite{zeng2020learning}. Others propose using 3D patches (also regarded as \textit{cubes}) instead~\cite{liu2021swinvideo,arnab2021vivit, akbari2021vatt, fan2021multiscale, tong2022videomae}, allowing to consider local motion features within the tokens themselves. While non-overlapping patches are the most common, a few works propose using overlapping 2D~\cite{Liu_2021_ICCV} or 3D~\cite{fan2021multiscale} patches for smoother information flow between neighboring patches. Due to their access to neighboring information in the input, we also regard positions of intermediate feature maps from CNN embedding networks as patches (e.g., 2D in~\cite{Liu_2021_ICCV, Su_2021_ICCV, Wang_2021_Transformer, wang2021end} or 3D in~\cite{gavrilyuk2020actor, purwanto2019extreme}), as their exact receptive field will depend on the specific setting in which they are produced. Overall, patch-based tokenization provides finer granularity, allowing to properly model spatiotemporal interactions in the VT.

\noindent\textbf{Instance-wise tokenization}. We refer to instances as semantically meaningful (foreground) regions that extend their reach beyond small patches but still smaller than whole frames~\cite{wu2021towards, Pashevich_2021_ICCV, zhu2020actbert, girdhar2019video}. On the one hand, a \textit{Region Proposal Network} (RPN in~\cref{fig:pipeline_b}), such as a Faster R-CNN~\cite{bb_faster}, can be used to generate region proposals and their corresponding embeddings~\cite{wu2021towards}. Thus, they allow reasoning about foreground objects or region interactions. Alternatively, in ~\cite{zhu2020actbert,girdhar2019video,li2021groupformer}, this kind of tokenization is combined with other coarser tokenizations (frame- and clip-wise tokenization) allowing to form instance-context relationships. Instance-based tokenization can be regarded as a form of sparse sampling (e.g., \cite{herzig2022object, roy2021action}), potentially reducing redundancy and allowing to input relatively large temporal sequences of per-frame instance representations to the VT without running into efficiency limitations. 

\noindent\textbf{Frame-wise tokenization}. In this case, the embedding network learns initial local spatial features for each frame, and the Transformer focuses on modeling the temporal interactions among the resulting frame tokens (e.g.,~\cite{patrick2021supportset,zeng2020learning,Pashevich_2021_ICCV,li2020hero,zhou2018end,yu2021accelerated,Wang_2021_ICCV,perrett2021temporal}). This allows longer videos to be modeled (especially compared to patch tokenization), although the Transformer may have a hard time modeling fine-grained spatial interactions. However, some tasks focusing on frame-level predictions (such as video summarization~\cite{fajtl2018summarizing}) may not require them.

\noindent\textbf{Clip-wise tokenization}. Condensing the information of several frames (clip) into each individual token allows further reducing the temporal dimension of the input (e.g., in~\cite{gabeur2020mmt,zhu2020actbert,lee2021parameter,sun2019videobert,contrastive2019chen, ging2020coot}). This way, the Transformer can effectively consume more frames to cover longer temporal spans. This makes clip tokenization very suitable for long-term modeling tasks. Given the high dimensionality of clips, it is necessary to embed them into single token representations through large embedding networks: for instance, \cite{Zhang_2021_CVPR} with C3D, \cite{zhu2020actbert} with 3D ResNet-50, \cite{sun2019videobert} with S3D, \cite{kalfaoglu2020late} with R(2+1)D, or \cite{lee2021parameter} with SlowFast, to name a few. This tokenization could also be suitable for retrieval tasks, where a high-level representation of the video is required~\cite{gabeur2020mmt,zhu2020actbert}. Clip-based tokenization exacerbates the pros and cons of frame-based tokenization where fine-grained information may be lost or mixed, preventing the Transformer from disentangling it later, in favor of efficiency when handling longer videos. 

\vspace{-0.15cm}
\subsection{Positional Embeddings (PE)}\label{sec:positional_encodings}

Given that SA is an operation on sets, signaling positional information is necessary in order to exploit the spatiotemporal structure of videos. This is done via positional embeddings (PE), which can be either \textit{fixed} or \textit{learned} and then \textit{absolute} or \textit{relative}: fixed absolute~\cite{wang2021spatiotemporal, girdhar2019video, fan2021multiscale}, learned absolute~\cite{jaegle2021perceiver,zhu2020actbert,lee2021parameter}, fixed relative~\cite{lei2020mart,rakhimov2020latent}, or learned relative~\cite{lin2020bi, Weissenborn2020Scaling, liu2021swinvideo}. Absolute variants are summed to the input embeddings but can also be concatenated~\cite{wang2021end,jaegle2021perceiver,ye2021referring}, while for the relative ones, the positional information is introduced directly in the multi-head attention~\cite{wu2021rethinking}. 

Absolute embeddings are generally 1D. This naturally fits frame or clip tokenization to indicate position in the only remaining (temporal) dimension. However, when dealing with patch-wise tokenization, fixed 1D in raster order may seem counter-intuitive, as the last patch $i$-th from row $j$, will be regarded as closer to the first patch in the next row $j+1$, than to patch $i$ at row $j-1$ (or $j+1$). For this reason, 2D absolute PE~\cite{Girdhar_2021_ICCV, gavrilyuk2020actor} accounting for joint space $wh$ and time $t$ dimensions, and 3D absolute PE~\cite{jaegle2021perceiver, Yu_2021_ICCV, wang2021spatiotemporal, wang2021end} for width $w$, height $h$, and $t$ have also been proposed, disregarding~\cite{dosovitskiy2021an} who found 1D learned absolute PE to suffice -- at least for images. 

The idea behind relative PE is that the positional information added when computing attention between token $i$ and $j$ depends on their relative position, making them translation equivariant. In other words, 1D relative PE added when computing attention between tokens at positions $i$ and $j=i+k$ will be the same regardless of the value for $i$ (i.e., $-k$). Relative PEs are generally added as an additional bias term (as in~\cite{lei2020mart,shaw2018self, dai2019transformer,liu2022swinv2}) in the dot-product between $\textbf{Q}$ and $\textbf{K}$ (modifying~\cref{eq:self-attention}). We find different variants of relative PEs applied to VTs, for instance~\cite{Weissenborn2020Scaling, liu2021swinvideo, Tan_2021_ICCV} are based on decomposable attention~\cite{parikh2016decomposable}, whereas \cite{lin2020bi} follows the approach of relation-aware attention~\cite{shaw2018self}. 

\vspace{-0.15cm}
\subsection{Discussion on input pre-processing}
Most VTs employ large CNN embeddings to reduce input dimensionality (aiding with data redundancy) and to exploit their ability to produce strong representations (thanks to local inductive biases). This significantly alleviates complexity and simplifies training when employing Transformers for video tasks. The success of these methods is clearly visible in the number of works which utilize large embedding networks as opposed to minimal embeddings. While minimal embeddings are indeed lighter than large CNN counterparts, they do result in overall more costly models if used naively. As they do not provide the necessary inductive biases, these will have to be provided elsewhere (such as in the Transformer design -- see~\cref{sec:architecture} --, or during training, through large-scale (self-)supervised pre-training -- see~\cref{sec:training}). However, as we observe in~\cref{sec:performance_comparsion}, this may result in better-performing models. Regarding tokenization, it has an impact on two main factors: (1) it will affect the level at which information is modeled by the VT (longer temporal spans by using frame- or clip-based tokenization, and more fine-grained spatiotemporal modeling when employing patches); (2) it will impact the input sequence length, and consequently the computational complexity of the model. For these reasons, most works use a patch-based approach accompanied by some efficient design, or frame-based tokenization, as it provides better long-term modeling scalability.

We find that the interactions between embedding and tokenization play a crucial role in defining the abstraction level and granularity at which the Transformer can model interactions. On the one hand, large embedding networks allow to produce tokens sharing information between them, guided by interactions defined by CNN's inductive biases. In this regard, it may be desirable to leverage 3D CNNs that provide local interactions among spatiotemporally neighboring positions. On the other hand, some tokenization strategies (such as 3D patches or clips) allow the formation of fine-grained temporal interactions within the token itself. This can be further motivated by most state-of-the-art VTs employing 3D patches. In this sense, the choices of embedding network and tokenization need to be carefully considered, as they will affect the level at which spatial and temporal interactions can be formed. 

Finally, the fixed absolute PEs proposed in~\cite{vaswani2017attention} require fewer parameters than the learned counterpart. However, the latter could be learning relevant positional relations that Fourier-like approaches are unable to capture (similarly to how learned convolutional filters replaced handcrafted features). The vast majority of VTs employ these absolute variants while the use of relative counterparts is still marginal. We believe, however, that the translation equivariance these latter provide could prove useful for generalizing to unseen lengths (see~\cref{sec:discussion}). This ability would be highly useful in the video domain as it is much more prone to display inconsistent temporal lengths (and cannot be re-scaled as easily as spatial dimensions, without harming fine-grained motion modeling -- see~\cref{sec:architecture_discussion}). 

\vspace{-0.2cm}
\section{Architecture}
\label{sec:architecture}
In this section, we overview Transformer designs. The different alternatives focus on specific limitations of VTs or on better exploiting the abundant information in videos. In~\cref{sec:efficient} we analyze approaches to reduce the number of tokens accessible in a single attention operation, aiming to reduce quadratic complexity. Then, in~\cref{sec:long-term_modeling} we describe proposals to enhance the temporal modeling capabilities of VTs. Finally, in~\cref{sec:multi-view} we explore specialized designs to separately capture fine-grained and coarse-level features.

\subsection{Efficient designs} 
\label{sec:efficient}
Given the high dimensionality of video, it may be challenging to represent long time spans without potentially incurring information loss or stumbling upon the quadratic attention matrix problem. For this reason, many works decompose full attention into multiple smaller SA. This has a two-fold benefit, as it will reduce the size of individual attention matrices while infusing different inductive biases. Two main trends are observed: (1) \textit{restricted} approaches, which limit the scope of a single SA operation but maintain the sequence length throughout the network; and (2) \textit{aggregation} approaches, which focus on progressively condensing information into smaller sets of tokens. A complete overview of our proposed taxonomy for efficient video designs can be seen in \cref{fig:efficient_taxonomy}.

\begin{figure*}[ht!]
    \centering
    \subfloat[][\cref{sec:restricted_attention}]{\includegraphics[height=3.8cm]{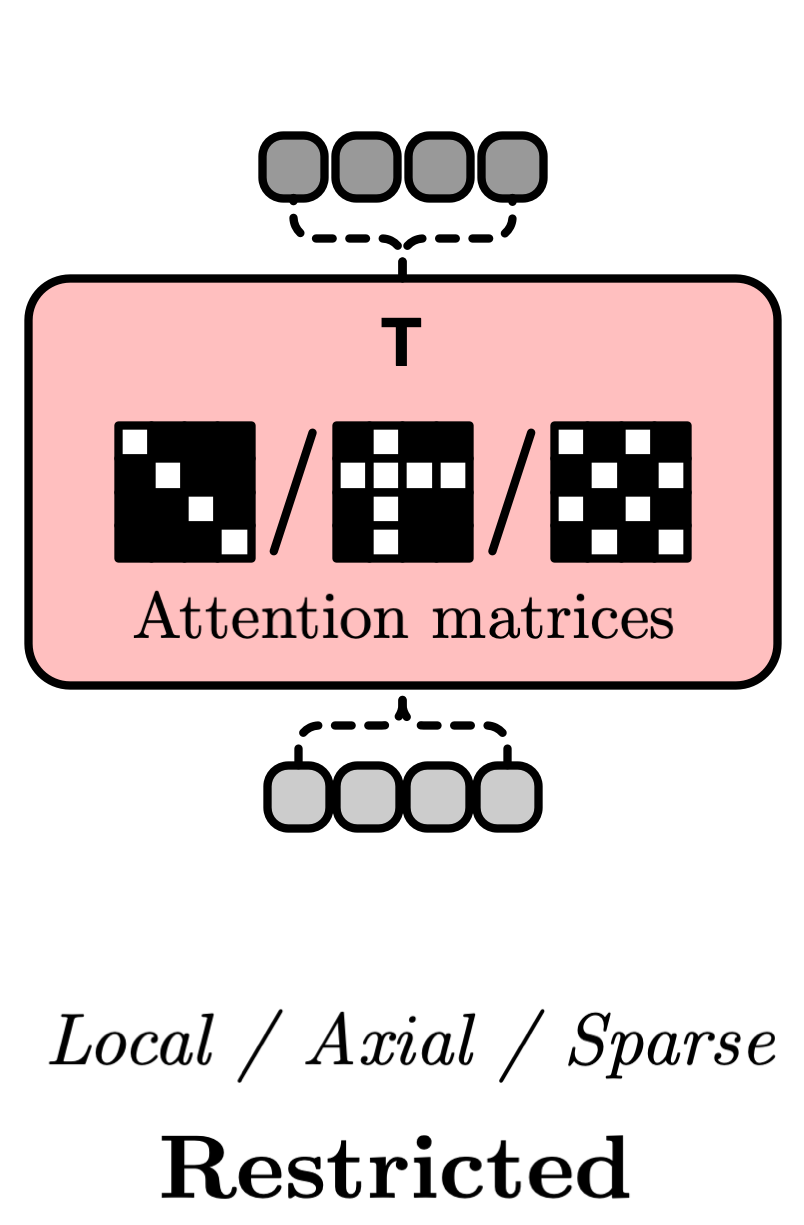}\label{fig:arch_a}}\hfill\hfill
    \subfloat[][\cref{sec:aggregation}]{\includegraphics[height=3.8cm]{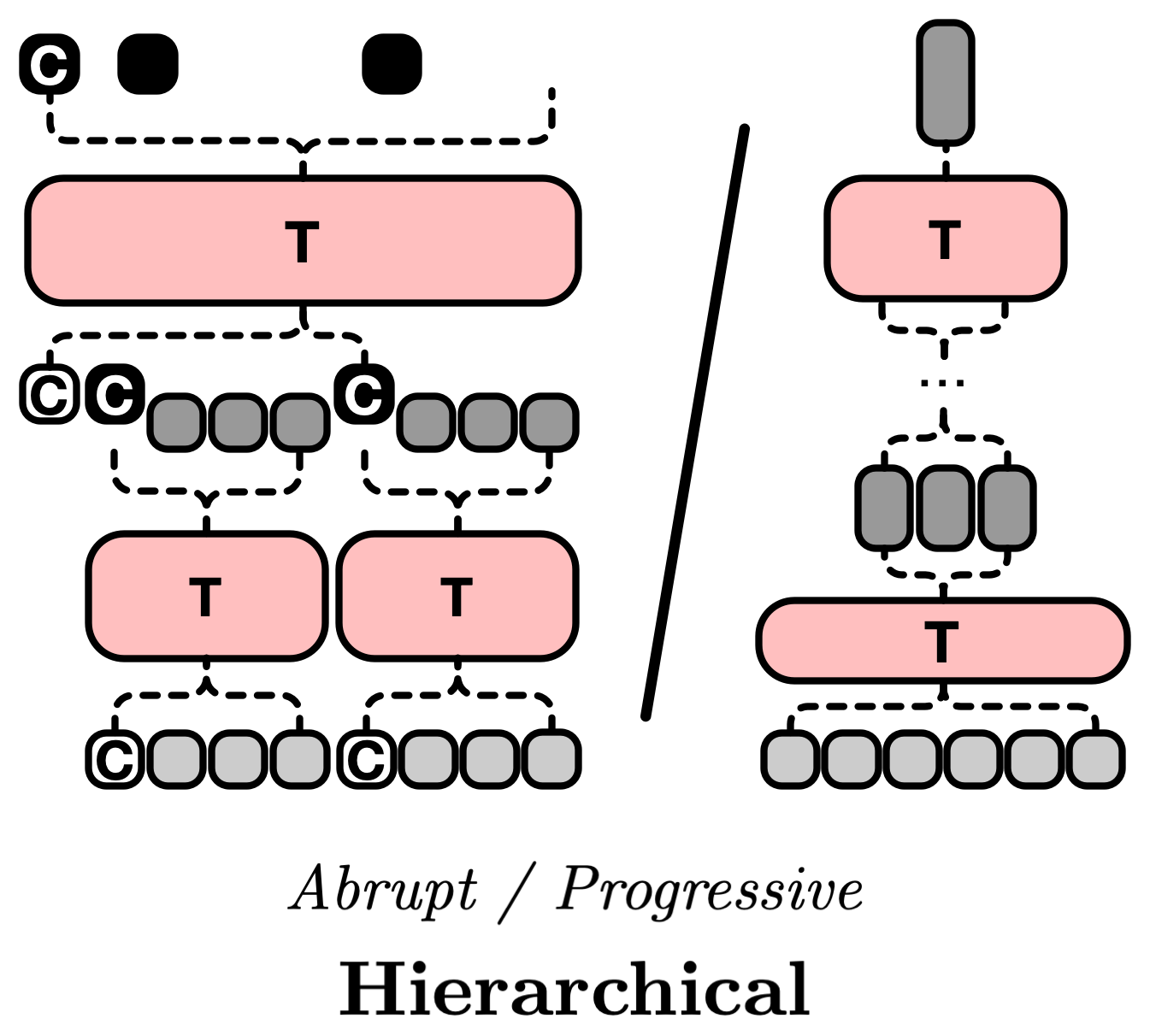}\label{fig:arch_b}}\hfill \hfill
    \subfloat[][\cref{sec:aggregation}]{\includegraphics[height=3.8cm]{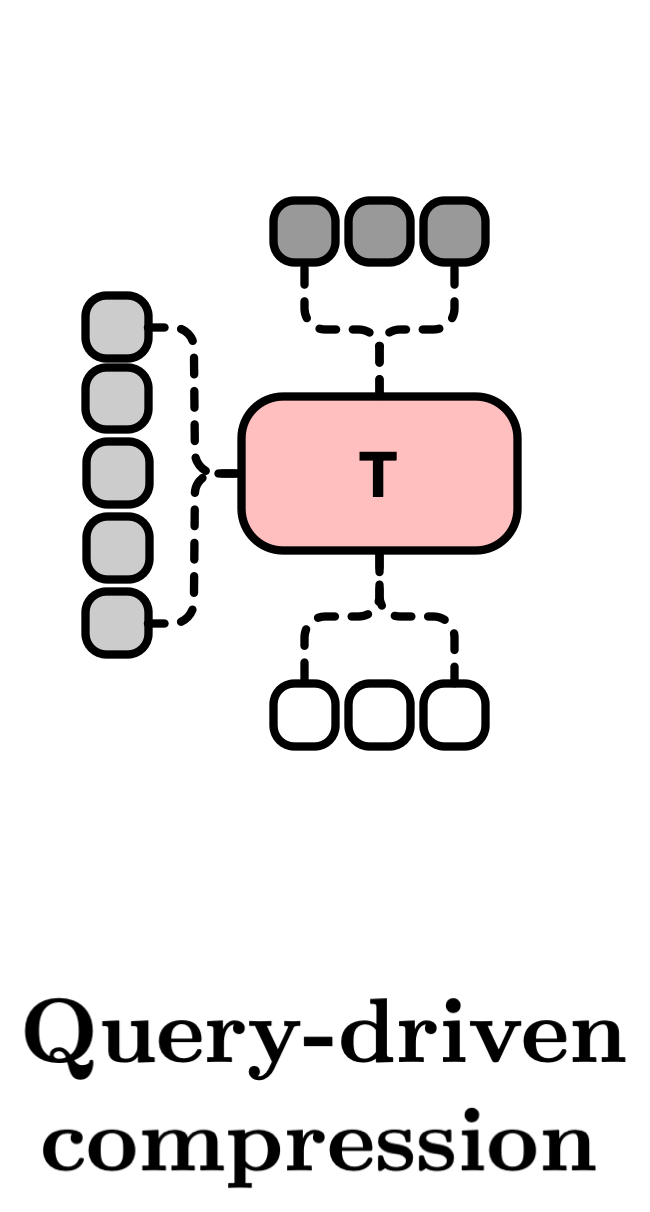}\label{fig:arch_c}}\hfill \hfill
    \subfloat[][\cref{sec:long-term_modeling}]{\includegraphics[height=3.8cm]{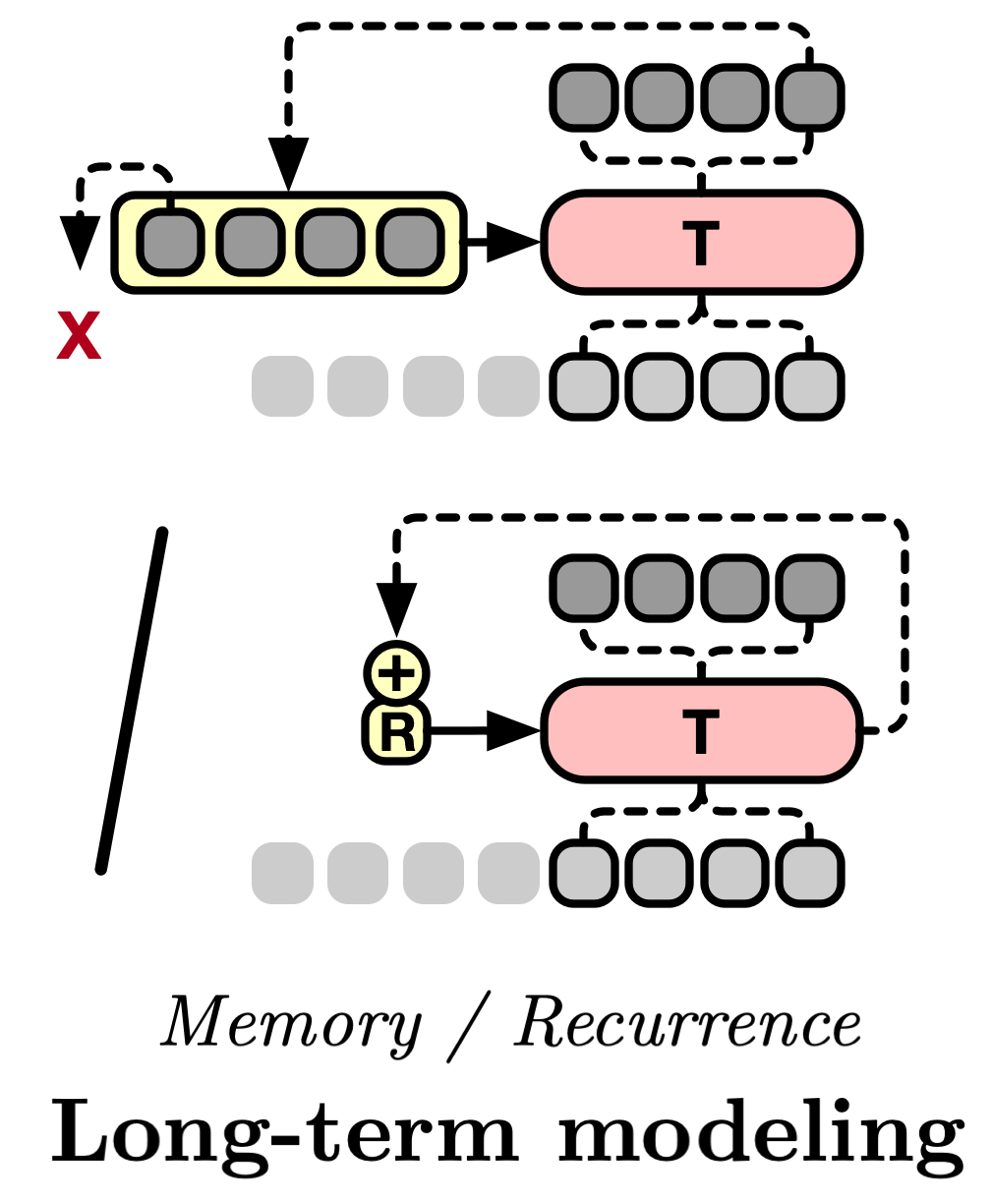}\label{fig:arch_d}}\hfill \hfill
    \subfloat[][\cref{sec:multi-view}]{\includegraphics[height=3.8cm]{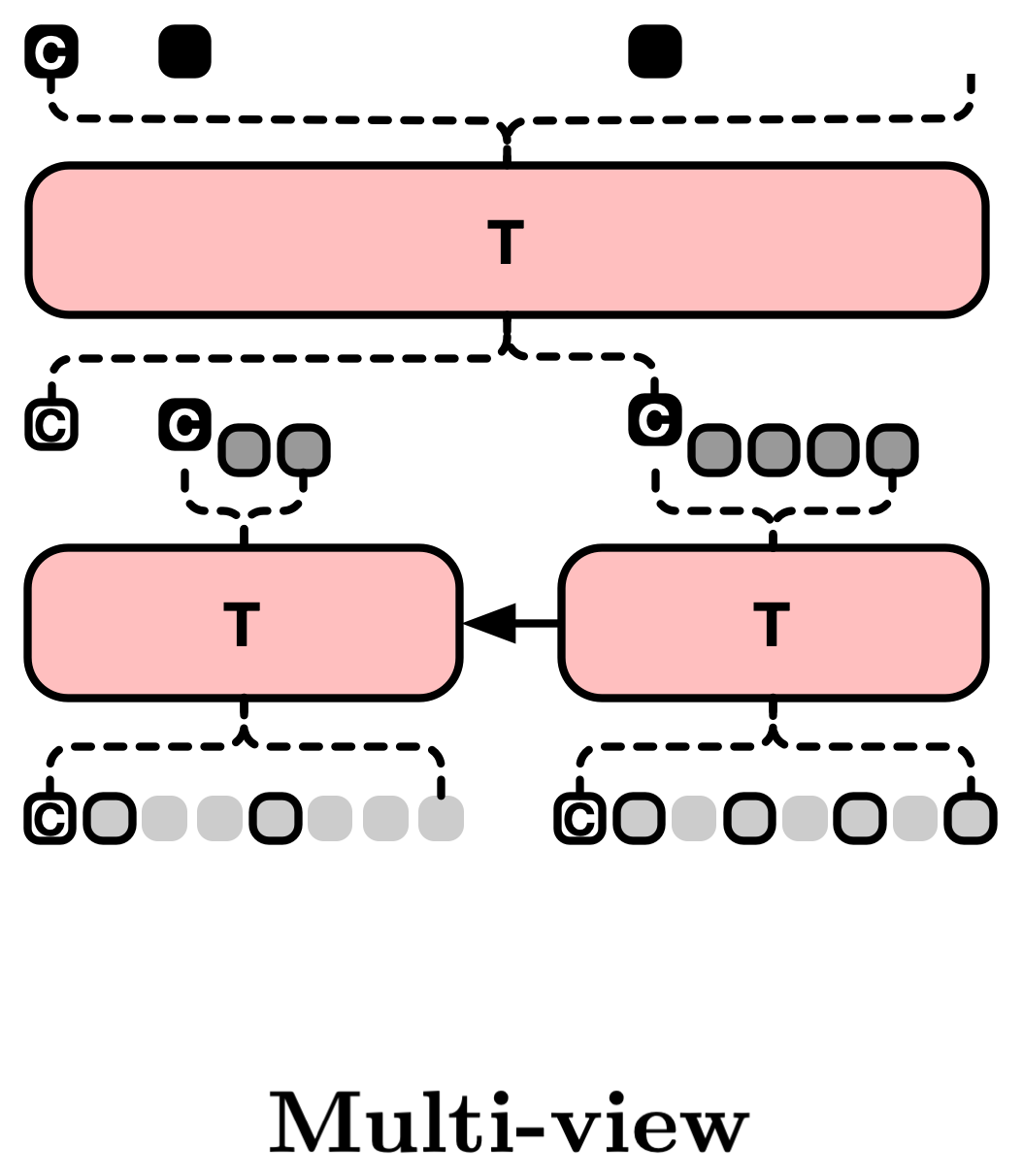}\label{fig:arch_e}}
    \caption{Visualization of the different design choices for VTs. Data tokens are in light gray (and black stroke if the token is used), whereas augmented tokens are in darker gray; those in white are initialized learnable tokens; and, [CLS] tokens are indicated with ``C'' (filled black after being augmented). Data flowing into the (T)ransformer from the side is used for cross-attention.}
    \label{fig:pipeline_c}
    \vspace{-0.25cm}
\end{figure*}

\vspace{-0.15cm}
\subsubsection{Restricted approaches} 
\label{sec:restricted_attention}
In order to approximate the full receptive field (i.e., the whole input sequence), restriction relies on stacking multiple smaller SA (similar to local filters in CNNs). We categorize restricted approaches by how subsets of tokens are selected for each SA. It can be by attending \textit{local} token neighborhoods, specific \textit{axis} (i.e., height, width or time) or \textit{sparsely} sampled subsets of tokens (see~\cref{fig:arch_a}). 

\noindent\textbf{Local approaches} are defined as the restriction by limiting attention to specific neighborhoods. Similar to \textit{sliding} filters in CNNs, the works in~\cite{neimark2021video, gu2020pyramid, bertasius2021spacetime, cong2021spatial, yang2021associating} define the neighborhoods by sampling nearby tokens given a query. Instead, ~\cite{liu2021swinvideo, liu2022swinv2, Weissenborn2020Scaling, zha2021shifted} proposed limiting SA to small \textit{fixed} windows, performing full SA separately in each of them. Relaxing the locality constraint only to time, in~\cite{yang2022temporally, bulat2021space} the fixed windows span all patches of a given frame. While sliding window local attention allows for more flexible learning (as each query has an independent local neighborhood), it has been shown to be cumbersome to implement~\cite{beltagy2020longformer}. Let $S$ and $T$ be the number of tokens in space and time respectively (i.e., $S \cdot T = N$), local approaches reduce the computational complexity of VTs from $\mathcal{O}((S \cdot T)^2)$ down to $\mathcal{O}(S \cdot T)$ assuming a small (and constant) spatiotemporal neighborhood size. These approaches gain locality biases and linear complexity at the expense of non-local receptive fields, hence will require depth to account for it. For this reason, in order to allow \textit{information to flow between windows}, we find different neighborhood sizes for each head in~\cite{gu2020pyramid, Weissenborn2020Scaling}, shifting the fixed windows on every layer in~\cite{liu2021swinvideo, liu2022swinv2} and swapping groups of features or neighborhood aggregation tokens between windows in~\cite{yang2022temporally, bulat2021space}. Instead, the use of global tokens is seen in~\cite{bertasius2021spacetime,zha2021shifted} (alternating between local and sparsely global attention), in~\cite{neimark2021video} (where the \texttt{[CLS]} token attends to and is attended by all tokens, acting as a bottleneck for non-local information) and in~\cite{bulat2021space} (which includes a global Transformer layer at the end).

\noindent\textbf{Axial approaches} define the restriction to attention by specific axes (i.e., \textit{height}, \textit{width}, or \textit{time}). These can only be applied in patch-based tokenization models, where the underlying structure of the data along the different axes is kept. \textit{Full axial attention} decomposition has been tested for VTs, either by attending over individual axes in three consecutive $\mathrm{MHA}$ sub-layers~\cite{bertasius2021spacetime}, or in a single one where each query token attends to all tokens that share with it the position in at least two axis~\cite{duke2021sstvos}. However, it is more common to decompose attention into spatial and temporal, for modeling intra-frame and inter-frame interactions respectively. Spatiotemporal decomposition reduces computational complexity from $\mathcal{O}(S^2 \cdot T^2)$ to $\mathcal{O}(S^2 \cdot T + S \cdot T^2)$. The way in which spatial and temporal attention are related in the architecture will define the granularity at which spatial, temporal, and spatiotemporal interactions of the input tokens are learned. On the one hand, allowing attention to both axes at each Transformer layer allows for \textit{spatiotemporal} relationships to form throughout layers. This can be done sequentially, through two $\mathrm{MHSA}$ sub-layers, as in~\cite{bertasius2021spacetime,arnab2021vivit} (and subsequent work~\cite{truong2022direcformer, yun2022matter, ranasinghe2022self}) or in parallel for latter combination, seen in~\cite{arnab2021vivit} through independent spatial and temporal heads and in~\cite{li2021groupformer} through  separate streams for each axis. On the other hand, entirely \textit{separating spatial from temporal} attention into consecutive modules as explored in~\cite{cong2021spatial,Girdhar_2021_ICCV}. In this sense, it is not until the latter layers that temporal modeling occurs, where it may be too late for certain spatial relationships to form. 

\noindent\textbf{Sparse approaches}. Sparse restrictions do not limit the scope of attended tokens, as opposed to local and axial approaches. Instead, given the high redundancy in video data~\cite{zhang2012slow}, sparse models provide a way to reduce unnecessary computation while maintaining a global receptive field at each layer. Sparsity can be \textit{embedded in the SA} operation by restricting it to fixed strided patterns for each query~\cite{bertasius2021spacetime, duke2021sstvos}. In other words, a given query is only allowed to attend (at most) to every other token on each axis. These are generally used to complement densely local attention. Other approaches involve some form of \textit{clustering}. This can be done through a hard assignment, where tokens get separated into groups (e.g., by k-means), allowing only attention within each of them. Intuitively, as SA contextualizes token representations through their relationships, these groupings allow attending directly to the most relevant ones for each token, discarding the ones that will contribute less. In order to allow for inter-group flow of information,~\cite{li2021groupformer} employs centroid SA, broadcasting contextualized cluster representations to each token within, whereas~\cite{zha2021shifted} uses an aggregation mechanism for later global modeling. Alternatively, in~\cite{patrick2021keeping} 
$\mathbf{Q}$ and $\mathbf{K}$ are softly assigned to a subset of maximally orthogonal ``prototypes'' sampled from $\mathbf{Q}$ and $\mathbf{K}$, performing SA in that reduced space. 

\vspace{-0.15cm}
\subsubsection{Aggregation approaches} \label{sec:aggregation} 
Aggregation-based VTs can be roughly categorized into \textit{hierarchical} and \textit{query-driven compression}. The key distinction is whether the input sequence length is reduced for all $\mathbf{Q}$, $\mathbf{K}$ and $\mathbf{V}$, or if a small set of tokens ($\mathbf{Q}$) is used to condense information from the full input sequence ($\mathbf{K}$ and $\mathbf{V}$).

\noindent\textbf{Hierarchical} designs can be further divided into abrupt or progressive. The former employ bigger neighborhoods (e.g., whole frames) and perform a single aggregation step, whereas the latter tend to work on smaller neighborhoods and involve multiple such steps (see \cref{fig:arch_b}). In both cases, the improvement in efficiency comes from the fact that deeper layers will have to process a smaller sequence length. \textit{Abrupt} approaches divide the input tokens into separate groups which are independently processed by a Transformer, to learn intra-group relationships. Then, information from each subset is aggregated, generally through a \texttt{[CLS]} token (e.g., ~\cite{arnab2021vivit,Girdhar_2021_ICCV}), although some use learnable global pooling in the form of linear~\cite{ging2020coot} or convolutional layers~\cite{ryoo2021tokenlearner}. The aggregated representations are then fed into the next stage, modeling inter-group relationships. We only find one work leveraging pure temporal hierarchy~\cite{ging2020coot}, which models frame-then-clip interactions. It is more common to employ spatiotemporal hierarchical models. These works (\!\!\!\cite{arnab2021vivit,Girdhar_2021_ICCV,neimark2021video,bulat2021space,yin2020lidar,zha2021shifted,hwang2021video}) are the aggregation equivalent of spatiotemporal axial methods: a first module (generally a ViT~\cite{dosovitskiy2021an} or Swin~\cite{liu2021swin} architecture), learns spatial patch-wise interactions, and a second one models frame-level temporal interactions. Interestingly, in~\cite{ryoo2021tokenlearner} multiple aggregation tokens are used for each frame, containing different features. As we discuss later in~\cref{sec:architecture_discussion}, these approaches may lose the ability to model fine-grained features after aggregation, potentially missing relevant temporal cues. 

\textit{Progressive} approaches, tackle this limitation by learning spatiotemporal interactions at all levels. In works such as Video Swin~\cite{liu2021swinvideo} and MViT~\cite{fan2021multiscale} (as well as their followups~\cite{li2022mvitv2,liu2022swinv2, wu2022memvit,wei2022masked,wang2022long,wang2022bevt,herzig2022object}) non-local interactions are learned at each level, whereas in~\cite{li2022uniformer} the first layers are limited to local interactions. In both cases, sequence length is progressively aggregated by local neighborhoods (i.e., through learnable local pooling) while expanding the tokens' dimensionality. While this increased model capacity for deeper layers will require more parameters (weight matrices $\mathbf{W}$ quadratically grow with the number of feature channels), it is generally compensated by smaller dimensionality in shallower layers. The work of~\cite{zha2021shifted} combines both types of hierarchy, by progressively downsampling in the spatial module, for latter aggregation and high-level temporal modeling.

\noindent\textbf{Query-driven compression}. Another aggregation-based approach consists in defining the set of queries $\mathbf{Q}$, such that $N_\mathrm{Q} \ll N$. Then, the computations are reduced from $\mathcal{O}(N^2)$ to $\mathcal{O}(N \cdot N_\mathrm{Q})$. In these, SA is performed only on the tokens that correspond to $\mathrm{Q}$, while $\mathrm{K}$ and $\mathrm{V}$ will be attended over via cross-attention. With this, the $N_\mathrm{Q}$ queries will iteratively access the whole input to distill the most useful information and aggregate it in the token embeddings corresponding to the queries. The intuition behind this is similar to how the input tokens to the decoder get refined by repeatedly cross-attending to the encoder's memory $\mathbf{M}$ (see~\cref{fig:transformer}). These queries are either an aggregated or sub-sampled version of the input data, or they are an independent set of tokens. 
\textit{Aggregating} the input into queries (e.g., through global pooling) can be used to build global streams while maintaining access to a broader low-level context within $K$ and $V$. This may be useful for tasks that require a high-level representation of the input clip (e.g., video retrieval~\cite{ging2020coot}, scene or action classification~\cite{seong2019video} or group activity recognition~\cite{li2021groupformer}). Interestingly, in~\cite{Su_2021_ICCV} this idea is developed by forming a reduced set of queries at each layer. In particular, $T$ and $S$ embeddings resulting from spatial and temporal average pooling respectively, are concatenated and used to attend the full set of keys and values. Alternatively, a \textit{sub-sampled} version of the input can be used to reason about specific regions or objects (e.g., by extracting a small set of boxes from the input clip to be used as queries~\cite{girdhar2019video,zhou2021hopper}). Using a fixed set of \textit{learnable queries} to cross-attend the input was first explored in~\cite{jaegle2021perceiver} to build a global stream, where latent embeddings are used to progressively gather information from the raw high-dimensional input. In VT literature it is more common to use these learnable queries in an object-centric fashion, extending on DETR~\cite{carion2020end} (used to detect objects at each frame) and propagating detection tokens to build recurrent Transformers (e.g., \cite{meinhardt2022trackformer,zheng2022vrdformer}). Alternatively, a set of independent \textit{text-based queries} can be defined from the text modality to aggregate relevant visual information for video question answering~\cite{kim2018multimodal}. This idea naturally extends the original Transformer, replacing the textual encoder with a video one while maintaining the auto-regressive text decoder, for video captioning~\cite{lei2020mart, li2020hero, BMT_Iashin_2020, Miech_2021_CVPR} or dense captioning~\cite{BMT_Iashin_2020, zhou2018end, yu2021accelerated} (through further event sampling).

\subsection{Long-term (temporal) modeling}
\label{sec:long-term_modeling}
Capturing long-term dynamics might be crucial for video tasks, as events observed at a given moment could potentially be only understood by looking far away in time. We here focus on works that propose dealing with long-term temporal modeling. We roughly categorize them into memory- (e.g., ~\cite{wu2022memvit, lei2020mart}) and recurrence-based approaches (e.g., \cite{yang2022recurring, meinhardt2022trackformer}). Whereas recurrent ones aggregate information into fixed-size representations, memory-based ones are variable-size and allow selective attention. In both, portions (i.e., frames/clips) of the videos are processed sequentially in a sliding window fashion to keep manageable compute and GPU memory but still ensure relevant information from past windows is within reach.

\subsubsection{Memory}
\label{sec:memory}
Naively caching many past raw (high-dimensional) input frames quickly becomes prohibitive. Instead, one can store global frame features~\cite{xu2021long, Fang_2019_CVPR} or convolutional maps late in the embedding network~\cite{yang2021associating}, intermediate embeddings across Transformer layers (e.g., those from patches~\cite{wu2022memvit}), or the Transformer's output embeddings~\cite{bozic2021transformerfusion}. In particular, when dealing with patch embeddings, aggregation might be needed before storing them~\cite{wu2022memvit}. On top of that, some works maintain several memories with different temporal reach (long/short)~\cite{xu2021long, yang2021associating}, abstraction level~\cite{wu2022memvit}, or granularity (fine/coarse)~\cite{wu2022memvit, bozic2021transformerfusion}.  

Memories are mostly \textit{accessed} via either cross-attention~\cite{Fang_2019_CVPR, xu2021long, yang2021associating} or self-attention~\cite{wu2022memvit, bozic2021transformerfusion}. By concatenating input and memory tokens sequence-wise to perform self-attention, the cost of the operation is $\mathcal{O}((N_\mathrm{M}+N_\mathrm{X})^2)$. Although manageable with small memories, cross-attention turns out to be much more affordable, with cost $\mathcal{O}(N_\mathrm{M} \cdot N_\mathrm{X})$ if we assume $N_\mathrm{X} \ll N_\mathrm{M}$. Either way, if $N_\mathrm{M}$ happens to be too large, one can \textit{reduce the number of tokens} on-the-fly when accessing them~\cite{Fang_2019_CVPR, xu2021long, wu2022memvit} by either query-driven compression~\cite{Fang_2019_CVPR, xu2021long} or progressive aggregation~\cite{wu2022memvit} -- both seen in~\cref{sec:aggregation}. On the one hand, memories leveraging query-driven compression follow a two-stage bottleneck compression: a first Transformer compresses the memory into a smaller set of tokens, whereas a second one ``decompresses'' the output of the former into a larger set but still much smaller than the original memory. In the case of~\cite{xu2021long} the second Transformer is also deeper than the one in the first stage. It also uses two separate sets of learnable tokens to perform the aggregation in both stages, while \cite{Fang_2019_CVPR} uses a hard selection of memory tokens in the first stage (obtained via \textit{Farthest Point Sampling}~\cite{qi2017pointnet++}). Besides the efficiency gained from such two-stage factorizations, we intuit differentiated underlying roles of the Transformers. While the first focuses on rough selection/compression, the second tries to recover as much information as possible, aggregating and further refining embeddings. On the other hand, progressive memory aggregation throughout the Transformer layers provides later access to finer-to-coarser details. For instance, \cite{wu2022memvit} keeps spatially aggregated $K_{(t-{M^\ell}):(t-2)}^\ell$ from previous timesteps after a learnable pooling and concatenates with lastly cached memory that is to be compressed in this iteration (i.e., $K_{t-1}^\ell$), and the current input's $K_t^\ell$ to be used in the $\ell$-th MHSA sub-layer (and analogously for $V$ embeddings).

\textit{Multiple memories} (e.g., short- and long-term) can be separately accessed and their respective memory-enhanced embeddings fused~\cite{yang2021associating} or, alternatively, a short-term memory (with fewer tokens) can drive the compression of the long-term one~\cite{xu2021long}. In multi-layer memories~\cite{wu2022memvit}, the ones in later Transformer layers implicitly access information provided by earlier ones, effectively allowing local memory accesses to approximate the full receptive field of the memory in deeper layers. As we move forward in time, memories are \textit{discarded} in a First-in First-out (FIFO) fashion~\cite{Fang_2019_CVPR, xu2021long,wu2022memvit,yang2021associating}. A notable exception is~\cite{bozic2021transformerfusion}, which leverages the self-attention weights to discard memory token(s) that are less attended by the rest. 

\vspace{-0.2cm}
\subsubsection{Recurrence}
\label{sec:recurrence}
Drawing inspiration from RNNs/LSTMs, recurrence mechanisms have also been proposed to deal with long video sequences. Here we distinguish between recurrence applied between intermediate layers in the VT~\cite{yang2022recurring,lei2020mart} and outside of it~\cite{meinhardt2022trackformer, zheng2022vrdformer}.

Within the first category, we find RViT~\cite{yang2022recurring} and MART~\cite{lei2020mart}. RViT~\cite{yang2022recurring} is essentially a ViT-like spatial Transformer that propagates the output of every self-attention sub-layer forward in time. Acting as recurrent states, these are added to the embeddings from the current time step after projecting both to its own $\textbf{Q}$, $\textbf{K}$, and $\textbf{V}$. Instead, MART~\cite{lei2020mart} leverages the embeddings alone to form $\textbf{Q}$ whereas a sequence-wise concatenation of those with the recurrent state is used to derive $\textbf{K}$ and $\textbf{V}$. Differently from RViT, the recurrent state is not the output of SA, but the result of a gating mechanism between the previous state and the current input embedding.

Recurrence can also be established outside the Video Transformer. In other words, the output embeddings from the Transformer at time $t-1$, namely $\hat{X}_{t-1}^{D}$, can be propagated to its own input at $t$. In the context of object detection, \cite{meinhardt2022trackformer, zheng2022vrdformer} propose an encoder-decoder architecture where the decoder augments a set of learnable tokens while attending to the encoder's representation of the current frame. At time $t=0$, the decoder augments an initial set of learnable tokens that will become recurrent tokens. At $t > 0$, the decoder augments the sequence-wise concatenation of the recurrent tokens at $t-1$ and added learnable tokens at $t$ to capture newly appeared objects. Trained using pairs of frames, these can still deal with long sequences during inference. One may argue that having a token for each object could be regarded as a form of memory, but from the point of view of time, the information is being recurrently aggregated into a fixed-size representation. 

\tikzset{every picture/.style={/utils/exec={\sffamily}}}

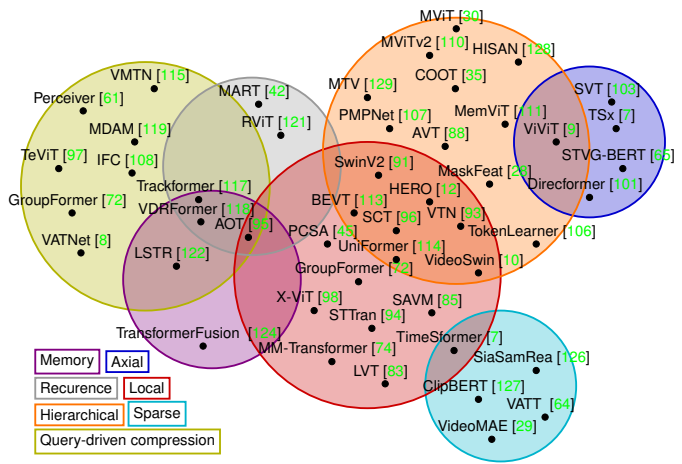
\begin{figure}    
    \pgfmathsetmacro{\rquery}{2.8}
    \pgfmathsetmacro{\rmemory}{2.0}
    \pgfmathsetmacro{\rrecurence}{2.0}
    \pgfmathsetmacro{\rsparse}{1.7}
    \pgfmathsetmacro{\rlocal}{3}
    \pgfmathsetmacro{\raxial}{1.7}
    \pgfmathsetmacro{\rhierarchical}{3}
    
    \pgfmathsetmacro{\bgopacity}{0.3}

    \definecolor{colquery}{rgb}{0.7,0.7,0.0}
    \definecolor{colmemory}{rgb}{0.5,0.0,0.5}
    \definecolor{col_recurrence}{rgb}{0.6,0.6,0.6}
    \definecolor{colsparse}{rgb}{0.0,0.7,0.8}
    \definecolor{collocal}{rgb}{0.8,0.0,0.0}
    \definecolor{colaxial}{rgb}{0.0,0.0,0.8}
    \definecolor{colhiearchical}{rgb}{1.0,0.45,0.0}

    \def\dotsize{2pt}
    \def\labelsize{\normalsize} 
    
    \hspace*{-0.25cm} 
    \resizebox{0.5\textwidth}{!}{
    \begin{tikzpicture}

        \coordinate (c_local) at        (5.0, -2.0);
        \coordinate (c_hierarchical) at (7, 0.8);
        \coordinate (c_axial) at        (10, 1);
        \coordinate (c_memory) at       (1.5, -2.1);
        \coordinate (c_sparse) at       (8.0, -4.5);
        \coordinate (c_recurency) at    (2.4, 0.45);
        \coordinate (c_query) at        (0, 0);
        
        \fill[fill=colquery, opacity=\bgopacity]   (c_query)  circle (\rquery);
        \fill[fill=colmemory, opacity=\bgopacity]   (c_memory)  circle (\rmemory);
        \fill[fill=col_recurrence, opacity=\bgopacity]   (c_recurency)  circle (\rrecurence);
        \fill[fill=colsparse, opacity=\bgopacity]   (c_sparse)  circle (\rsparse);
        \fill[fill=collocal, opacity=\bgopacity]   (c_local)  circle (\rlocal);
        \fill[fill=colaxial, opacity=\bgopacity]   (c_axial)  circle (\raxial);
        \fill[fill=colhiearchical, opacity=\bgopacity] (c_hierarchical) circle (\rhierarchical);
        
        \draw[color=colquery, very thick]   (c_query)  circle (\rquery);
        \draw[color=colmemory, very thick]   (c_memory)  circle (\rmemory);
        \draw[color=col_recurrence, very thick]  (c_recurency)  circle (\rrecurence);
        \draw[color=colsparse, very thick]   (c_sparse)  circle (\rsparse);
        \draw[color=collocal, very thick]   (c_local)  circle (\rlocal);
        \draw[color=colaxial, very thick]   (c_axial)  circle (\raxial);
        \draw[color=colhiearchical, very thick]   (c_hierarchical) circle (\rhierarchical);
        
        \node[draw, very thick, anchor=west,color=colquery] at (-2.5,-5.75) {\color{black}Query-driven compression};
        \node[draw, very thick, anchor=west,color=colhiearchical] at (-2.5,-5.15)  {\color{black}Hierarchical};
        \node[draw, very thick, anchor=west,color=col_recurrence] at (-2.5,-4.55)  {\color{black} Recurence};
        \node[draw, very thick, anchor=west,color=colmemory] at (-2.5,-3.95) {\color{black}Memory};
        \node[draw, very thick, anchor=west,color=colsparse] at (-0.4,-5.125) {\color{black}Sparse};
        \node[draw, very thick, anchor=west,color=collocal] at (-0.5,-4.55) {\color{black}Local};
        \node[draw, very thick, anchor=west,color=colaxial] at (-0.9,-3.95)  {\color{black}Axial};

        
        \filldraw[black] (c_query)++(-0.24,1.0) circle (\dotsize) node[anchor=south, xshift=-0.1cm]   {\labelsize MDAM~\cite{kim2018multimodal}}; 
        \filldraw[black] (c_query)++(-1.4,1.7) circle (\dotsize) node[anchor=south, xshift=-0.1cm]  {\labelsize Perceiver~\cite{jaegle2021perceiver}}; 
        \filldraw[black] (c_query)++(-1.45,-1.5) circle (\dotsize) node[anchor=south, xshift=-0.1cm]  {\labelsize VATNet~\cite{girdhar2019video}}; 
        \filldraw[black] (c_query)++(0.2,2.2) circle (\dotsize) node[anchor=south, xshift=-0.1cm]  {\labelsize VMTN~\cite{seong2019video}}; 
        \filldraw[black] (c_query)++(-1.95,0.4) circle (\dotsize) node[anchor=south, xshift=-0.1cm]  {\labelsize TeViT~\cite{yang2022temporally}}; 
        \filldraw[black] (c_query)++(-0.3,0.3) circle (\dotsize) node[anchor=south, xshift=-0.1cm]  {\labelsize IFC~\cite{hwang2021video}}; 
        \filldraw[black] (c_query)++(-1.65,-0.6) circle (\dotsize) node[anchor=south, xshift=-0.1cm]   {\labelsize GroupFormer~\cite{li2021groupformer} }; 

        \filldraw[black] (c_query)++(0.7,-1.8) circle (\dotsize) node[anchor=south, xshift=-0.1cm]  {\labelsize LSTR~\cite{xu2021long}}; 
        
        \filldraw[black] (c_query)++(1.2,-0.3) circle (\dotsize) node[anchor=south, xshift=-0.1cm]  {\labelsize Trackformer~\cite{meinhardt2022trackformer}}; 
        \filldraw[black] (c_query)++(1.25,-0.80) circle (\dotsize) node[anchor=south, xshift=-0.1cm]  {\labelsize VDRFormer~\cite{zheng2022vrdformer}}; 

        \filldraw[black] (c_memory)++(-0.2,-1.5) circle (\dotsize) node[anchor=south, xshift=-0.1cm]{TransformerFusion~~\cite{bozic2021transformerfusion}}; 
       
        \filldraw[black] (c_recurency)++(0.15,1.4) circle (\dotsize) node[anchor=south, xshift=-0.1cm]  {\labelsize MART~\cite{lei2020mart}}; 
        \filldraw[black] (c_recurency)++(0.65,0.7) circle (\dotsize) node[anchor=south, xshift=-0.1cm]  {\labelsize RViT~\cite{yang2022recurring}}; 

       
        \filldraw[black] (c_sparse)++(-1.05,0.80) circle (\dotsize) node[anchor=south, xshift=-0.1cm]  {\labelsize TimeSformer~\cite{bertasius2021spacetime}}; 

        \filldraw[black] (c_sparse)++(1.,-0.7) circle (\dotsize) node[anchor=south, xshift=-0.1cm]{VATT~~\cite{akbari2021vatt}}; 
        \filldraw[black] (c_sparse)++(0.8,0.35) circle (\dotsize) node[anchor=south, xshift=-0.1cm]{SiaSamRea~\cite{yu2021learning}\*}; 
        \filldraw[black] (c_sparse)++(-0.5,-0.3) circle (\dotsize) node[anchor=south, xshift=-0.1cm]{ClipBERT\cite{lei2021less}\*}; 
        \filldraw[black] (c_sparse)++(-0.2,-1.2) circle (\dotsize) node[anchor=south, xshift=-0.1cm]{VideoMAE\cite{tong2022videomae}\*}; 
        
        \filldraw[black] (c_local)++(0.4,-2.45) circle (\dotsize) node[anchor=south, xshift=-0.1cm]   {\labelsize LVT~\cite{rakhimov2020latent}}; 
        \filldraw[black] (c_local)++(1.45,-0.85) circle (\dotsize) node[anchor=south, xshift=-0.1cm]   {\labelsize SAVM~\cite{Weissenborn2020Scaling}}; 
        \filldraw[black] (c_local)++(-0.9,0.7) circle (\dotsize) node[anchor=south, xshift=-0.1cm]{\labelsize  PCSA~\cite{gu2020pyramid}}; 
        \filldraw[black] (c_local)++(-1.2,-0.8) circle (\dotsize) node[anchor=south, xshift=-0.1cm]   {\labelsize X-ViT\cite{bulat2021space}}; 
        \filldraw[black] (c_local)++(-0.8,-1.95) circle (\dotsize) node[anchor=south, xshift=-0.1cm]   {\labelsize MM-Transformer~\cite{roy2021action}}; 
        \filldraw[black] (c_local)++(0.1,-1.20) circle (\dotsize) node[anchor=south, xshift=-0.1cm]   {\labelsize STTran~\cite{cong2021spatial}}; 
        \filldraw[black] (c_local)++(-0.2,-0.15) circle (\dotsize) node[anchor=south, xshift=-0.1cm]   {\labelsize GroupFormer~\cite{li2021groupformer} }; 
        
        \filldraw[black] (c_local)++(1.4,1.65) circle (\dotsize) node[anchor=south, xshift=-0.1cm]   {\labelsize HERO~\cite{li2020hero}}; 
        \filldraw[black] (c_local)++(2.1, 1.1) circle (\dotsize) node[anchor=south, xshift=-0.1cm] {\labelsize VTN~\cite{neimark2021video}}; 
        \filldraw[black] (c_local)++(2.5,0.05) circle (\dotsize) node[anchor=south, xshift=-0.1cm]   {\labelsize VideoSwin~\cite{liu2021swinvideo}}; 
        \filldraw[black] (c_local)++(-0.3,1.4) circle (\dotsize) node[anchor=south, xshift=-0.1cm]   {\labelsize BEVT~\cite{wang2022bevt}}; 
        \filldraw[black] (c_local)++(0.65,0.35) circle (\dotsize) node[anchor=south, xshift=-0.1cm]   {\labelsize UniFormer~\cite{li2022uniformer}}; 
        \filldraw[black] (c_local)++(0.65,1.0) circle (\dotsize) node[anchor=south, xshift=-0.1cm]   {\labelsize SCT~\cite{zha2021shifted}}; 
        \filldraw[black] (c_local)++(0.25,2.25) circle (\dotsize) node[anchor=south, xshift=-0.1cm]{\labelsize SwinV2~\cite{liu2022swinv2}}; 

        \filldraw[black] (c_local)++(-2.68,0.85) circle (\dotsize) node[anchor=south, xshift=-0.1cm]   {\labelsize AOT~\cite{yang2021associating}}; 
        
        \filldraw[black] (c_axial)++(-0.75,0) circle (\dotsize) node[anchor=south, xshift=-0.1cm]{\labelsize ViViT~\cite{arnab2021vivit}};
        
        \filldraw[black] (c_axial)++(0.6,0.3) circle (\dotsize) node[anchor=south, xshift=-0.1cm]   {\labelsize TSx~\cite{bertasius2021spacetime}}; 
        \filldraw[black] (c_axial)++(0.75, -0.6) circle (\dotsize) node[anchor=south, xshift=-0.1cm]   {\labelsize STVG-BERT~\cite{Su_2021_ICCV}}; 
        \filldraw[black] (c_axial)++(0.0, -1.25) circle (\dotsize) node[anchor=south, xshift=-0.1cm]   {\labelsize Direcformer~\cite{truong2022direcformer}}; 
        \filldraw[black] (c_axial)++(0.5,0.9) circle (\dotsize) node[anchor=south, xshift=-0.1cm]   {\labelsize SVT~\cite{ranasinghe2022self}}; 
        
        \filldraw[black] (c_hierarchical)++(1.4, 2.0) circle (\dotsize) node[anchor=south, xshift=-0.1cm]{\labelsize HISAN~\cite{pramono2019hierarchical}}; 
        \filldraw[black] (c_hierarchical)++(0,2.75) circle (\dotsize) node[anchor=south, xshift=-0.1cm]{\labelsize MViT~\cite{fan2021multiscale}}; 
        \filldraw[black] (c_hierarchical)++(-0.6,2.15) circle (\dotsize) node[anchor=south, xshift=-0.1cm]{\labelsize MViTv2~\cite{li2022mvitv2}}; 
        \filldraw[black] (c_hierarchical)++(-1.5,0.5) circle (\dotsize) node[anchor=south, xshift=-0.1cm]{\labelsize PMPNet~\cite{yin2020lidar}}; 
        \filldraw[black] (c_hierarchical)++(-2.0,1.2) circle (\dotsize) node[anchor=south, xshift=-0.1cm]{\labelsize MTV~\cite{yan2022multiview}}; 
        \filldraw[black] (c_hierarchical)++(0.75,-0.75) circle (\dotsize) node[anchor=south, xshift=-0.1cm]{\labelsize MaskFeat~\cite{wei2022masked}}; 
        \filldraw[black] (c_hierarchical)++(1.8,-2.1) circle (\dotsize) node[anchor=south, xshift=-0.1cm]{\labelsize TokenLearner~\cite{ryoo2021tokenlearner}}; 
        \filldraw[black] (c_hierarchical)++(-0.25,0.1) circle (\dotsize) node[anchor=south, xshift=-0.1cm]{\labelsize AVT~\cite{Girdhar_2021_ICCV}}; 
        \filldraw[black] (c_hierarchical)++(-.02,1.4) circle (\dotsize) node[anchor=south, xshift=-0.1cm]   {\labelsize COOT~\cite{ging2020coot}}; 
        \filldraw[black] (c_hierarchical)++(1.1,0.6) circle (\dotsize) node[anchor=south, xshift=-0.1cm]{\labelsize MemViT~\cite{wu2022memvit}}; 

    \end{tikzpicture}
    }
    \vspace{-0.65cm}
    \caption{Venn diagram displaying our proposed taxonomy of efficient VT designs (best viewed in color). We describe Local, Axial and Sparse approaches in~\cref{sec:restricted_attention}, Hierarchical and Query-driven compression in~\cref{sec:aggregation}, and Memory and Recurrence in~\cref{sec:long-term_modeling}.}
    \label{fig:efficient_taxonomy}
    \vspace{.15cm}
\end{figure}

\subsection{Multi-view approaches}
\label{sec:multi-view}
Opposed to dense sampling of single views, a few VTs define multiple views of a given video to solve the task at hand in a cooperative fashion. Instance-based contrastive approaches instead employ multiple views to drive the loss (see~\cref{sec:contrastive_learning}). Note that multi-view approaches are related to multi-view sampling at inference (see~\cref{sec:performance_video_classification}), but crucially, the former leverage this technique also during training. A clear example of this parallel is~\cite{lei2021less}, which defines \textit{sparse views} by uniformly sampling video frames with a fixed stride but varying starting positions. Then, separate streams process each view and the final classification is reached by averaging predictions in a late fusion manner. This work could be seen as the sparse equivalent to fixed window local restriction. In this sense, it only incurs $\mathcal{O}(R^2k)$, where $k$ is the number of sparse sequences (i.e., $R \cdot k = N$). As weights are shared across streams, no parameter increase is incurred. 

Interestingly, many approaches define views by \textit{varying the resolution} of a given clip, while allowing interactions between them to form throughout the network (i.e., early fusion). This was first explored for video in~\cite{zeng2020learning} by using patches of different spatial size at each head, and later extended to time in~\cite{yan2022multiview} by using 3D patches instead. In the latter case, a multi-stream network is used where each stream models the same video but tokenizes with different temporal resolution (inspired by the SlowFast Network~\cite{Feichtenhofer_2019_ICCV}), allowing information flow between views through cross-attention and a final global stream (in an abrupt hierarchical fashion). In~\cite{Weng_2021_ICCV} a similar architectural setting is used, but the views are sampled from the output of progressively deeper layers of a ConvLSTM embedding network. In this sense, each view holds a smaller spatial resolution, but a bigger temporal context. Intuitively, these methods use redundancy to their advantage, helping the network become robust to missing information in single views, while each stream models a coherent representation of the full input.

\vspace{-0.15cm}
\subsection{Discussion on Architecture} 
\label{sec:architecture_discussion}
VT designs focus on reducing computational complexity and handling the redundancy of videos without compromising spatiotemporal modeling capabilities. Furthermore, restrictions imposed on VTs to make them more efficient will bias them towards favoring certain kinds of relationships. For instance, abrupt hierarchy learns temporal translation equivariance in spatial layers by modeling each frame independently, and local approaches that enforce locality biases. 
 
However, efficient designs and inductive biases do not handle redundancy. Video redundancy can be mostly attributed to appearance-based semantics varying slowly through time, even when small variations in specific pixels occur~\cite{zhang2012slow}. However, the extended information provided by these subtle changes in many consecutive frames may be crucial to properly model fine-grained motion features~\cite{Feichtenhofer_2019_ICCV}. In order to learn spatiotemporal relationships from video, this must be taken into account. Reducing spatial redundancy may be desirable, as it will allow focusing on more relevant parts of the video (e.g., through aggregation or sub-sampling of tokens). However, this requires careful consideration: removing certain information too early into the network may limit the formation of crucial temporal interactions later on. Prior works on modeling video with CNNs have shown this to be the case: early aggregation of spatial features hinders the formation of fine-grained motion features~\cite{sevilla2021only, bb_c3d, Lin_2019_ICCV}, and temporal pooling seems to hurt spatiotemporal representation learning~\cite{chen2021deep,feichtenhofer2020x3d}. With Transformers, tackling this may involve taking into account non-local neighborhoods before deciding which information is to be discarded.

Motivated by this, we derive three crucial aspects for spatiotemporal modeling: (1) explicit spatial redundancy reduction while (2) allowing to model temporal features at all levels in (3) high-fidelity temporal contexts. Different VTs exhibit varying degrees of capabilities in these three aspects. \textit{Restricted approaches} allow for low-level temporal modeling and, due to the lack of aggregation, always maintain temporal fidelity. Given their potential to overlap low- and high-level features they can be suitable for both low-level (e.g., segmentation~\cite{duke2021sstvos}) and high-level tasks (e.g., classification~\cite{bertasius2021spacetime}), but with certain limitations. Hierarchical approaches effectively exhibit (1) and (3) through aggregation on the spatial dimensions only (except in~\cite{li2022uniformer}). Particularly, for \textit{progressive hierarchy} (e.g.,~\cite{li2022mvitv2,liu2022swinv2}), the gradual increase in channel dimensionality provides deeper layers with a larger capacity to represent high-level concepts while further limiting the modeling of redundant low-level features. Furthermore, by leveraging different levels of spatiotemporal non-local contexts (e.g.,~\cite{liu2021swinvideo, fan2021multiscale}) at least in deeper layers (e.g,~\cite{li2022uniformer}), they guarantee that extended temporal fidelity is exploited before aggregation. In contrast, the \textit{abrupt counterparts} (e.g., \cite{Girdhar_2021_ICCV}) may be suffering from early aggregation. While training end-to-end may infuse temporal feedback into spatial layers (which may be sufficient for appearance-biased video benchmarks, see~\cref{sec:discussion_training}), they may lack proper motion modeling. This can be addressed by allowing to form spatiotemporal interactions before aggregation, either locally (by explicitly sharing information between neighborhoods~\cite{bulat2021space,yang2022temporally,zha2021shifted,hwang2021video} as well as by using 3D patches~\cite{arnab2021vivit, yan2022multiview}) or globally~\cite{ryoo2021tokenlearner}. \textit{Query-driven compression} approaches reduce redundancy through aggregation when used to derive global streams~\cite{ging2020coot, seong2019video, li2021groupformer}, or through sparsity when reasoning of individual objects or regions~\cite{li2021groupformer,girdhar2019video}. In both cases, the small set of queries forms high-level representations of (parts of) the input, while maintaining temporal fidelity in keys and values. However, they may exhibit a limited capability to form low-level temporal features. 
While iterative accesses may alleviate the dangers of early aggregation for high-level tasks (e.g., classification~\cite{seong2019video,jaegle2021perceiver,girdhar2019video}), low-level tasks may require to also evolve the fine-grained input representations~\cite{li2021groupformer} or to infuse them back with high-level features from the queries~\cite{Su_2021_ICCV} (similar to clustering-based sparse approaches). This is similar to the behavior exhibited by \textit{recurrent} approaches. As temporal information is collapsed into the recurrent state, they may suffer from early aggregation, which may be especially detrimental for high-level tasks~\cite{yang2022recurring}. However, these approaches may excel on applications that only require low-level reasoning of the current observation, enhanced with the forwarded high-level past context (such as for tracking~\cite{meinhardt2022trackformer}, segmentation~\cite{yang2021associating} or dense video captioning~\cite{lei2020mart}). \textit{Memory}-based approaches exhibit great capabilities for preserving the temporal resolution of the input. They can tackle redundancy through aggregation (e.g., upon storing~\cite{wu2022memvit} or dynamically on access~\cite{xu2021long}) or sparsity (either by storing only some past observations~\cite{yang2021associating}, by dropping elements in the memory according to their relevance~\cite{bozic2021transformerfusion} or only attending to a small subset of memory tokens~\cite{Fang_2019_CVPR}). Finally, \textit{multi-view} approaches working at different input resolutions explicitly allow the formation of separate coarse- and fine-level features while allowing interactions between them~\cite{Weng_2021_ICCV, zeng2020learning}. However, as redundancy is not explicitly managed, the success of these methods may be limited to computationally heavy models~\cite{yan2022multiview}. Sparse counterparts heavily downsample the input sequence, hurting temporal fidelity and requiring to compensate with other modalities~\cite{yu2021learning, lei2021less, akbari2021vatt}.

\vspace{-0.2cm}
\section{Training a Transformer}
\label{sec:training}
The two main limitations of Transformers will heavily influence the way in which they are trained. 
On the one hand, large-scale pre-training aids Transformers to overcome their lack of inductive biases~\cite{dosovitskiy2021an,chen2021outperform,devlin2019bert}, but recent studies suggest that self-supervised pre-training (see~\cref{sec:pretext_tasks}) alleviates the need for large supervised datasets~\cite{tong2022videomae,wei2022masked}. On the other hand, some solutions to the lack of inductive biases aggravate computational costs. CNN embedding networks add to the memory footprint and potentially overflow GPU memory when training, especially if done end-to-end. Avoiding overfitting big models requires strong regularization~\cite{xu2019overfitting} and lots of data~\cite{zhai2021scaling}, which is further problematic when handling several stages of training that require more time and compute. Finally, leveraging self-supervised tasks is computationally heavy, especially for video. 

\vspace{-0.15cm}
\subsection{Training regime}
\label{sec:training_regime}
We next explore how VTs are trained, from a lens of embedding networks and pre-training. Pre-training involves one or more training stages before transferring the network to a downstream task, for which the model is either fine-tuned or linearly probed (training a few linear layers on top of the frozen Transformer). 

\noindent\textbf{End-to-End training with minimal embeddings}. 
End-to-end training of deep neural networks has proven to outperform multiple-stage algorithms. To ease memory limitations while allowing for end-to-end training of the Transformer, it is common to use minimal embeddings. Some train in a supervised fashion~\cite{bertasius2021spacetime,jaegle2021perceiver,arnab2021vivit,Weissenborn2020Scaling,fan2021multiscale,zha2021shifted}, directly for a downstream task on large datasets, such as Kinetics-700~\cite{carreira2019short}, or ImageNet21K~\cite{ridnik2021imagenetk}. However, all these leveraged efficient architectures and thanks to the inductive biases these designs provide, the network will pick up on relevant patterns faster, and more capacity can be given to Transformer layers. Other works aiming for smaller datasets train aided by some data augmentation~\cite{yuan2020temporal,Weng_2021_ICCV,Yu_2021_ICCV} or self-supervised losses~\cite{zeng2020learning,Liu_2021_ICCV,Girdhar_2021_ICCV,Miech_2021_CVPR,akbari2021vatt} on medium to large datasets. Stand-alone Transformers seem to be able to learn without large CNN embeddings if aided by the inductive biases that efficient designs, data augmentation, or self-supervised losses provide. Still, most of these require multiple stages of training either through large datasets or computationally heavy self-supervised techniques.

\noindent\textbf{End-to-End with embedding networks}. Other works train Transformer and deep CNN embedding layers end-to-end either with a pre-trained embedding network~\cite{Su_2021_ICCV,chen2021transformer,kim2018multimodal}, fine-tuning just the later layers~\cite{contrastive2019chen, seong2019video}, or training end-to-end from scratch~\cite{wang2021spatiotemporal,luo_sychro_2020, Wang_2021_Transformer,Patrick_2021_ICCV}. Some were able to train end-to-end by capping Transformers to 1$\sim$4 layers \cite{li2021vidtr,kalfaoglu2020late,yin2020lidar}, suggesting that just a few Transformer layers after a large embedding network may be enough to boost performance. Some others' success is attributable to leveraging efficient designs (e.g., local SA~\cite{gu2020pyramid}) or weight sharing~\cite{lee2021parameter} -- that reduces the effective number of parameters to be stored in memory (discussed later in~\cref{sec:discussion}). Finally,~\cite{wang2020end,girdhar2019video} report having substantial computational resources available, which allowed them to fit in memory both, a large embedding network and a big Transformer. Empirical studies on both image~\cite{ramachandran2019standalone,dai2021coatnet} and video~\cite{li2022uniformer} Transformers have consistently found improvements when training Transformers and CNN embedding layers end-to-end. This may further be seen in works reporting improved CNN-based results alone after being trained as the embedding net of a Transformer~\cite{lee2021parameter,contrastive2019chen,Sun_2021_ICCV}, pointing towards CNNs benefiting from long-term temporal feedback provided by the Transformer layers. 

\noindent\textbf{Frozen embedding networks}. The most common approach by far for VTs is leveraging some large pre-trained and frozen CNN embedding network. These are then used for feature extraction, which further boosts cost-effectiveness, as they can be pre-computed. Transformer layers are then trained for a downstream task on those features. Compared to end-to-end training from scratch, it is often cheaper and more efficient to employ SOTA models, which have been carefully tuned to perform well on some supervised task. While it is definitely common to use medium to large datasets (as in~\cite{lei2020mart,zhou2018end,Pashevich_2021_ICCV,BMT_Iashin_2020,perrett2021temporal,Zhang_2021_CVPR,yu2021accelerated,iashin2020multi}), with this approach, many video works~\cite{purwanto2019extreme,pramono2019hierarchical,camgoz2020sign,chen2019bert4sessrec,purwanto2019three,gavrilyuk2020actor,camgoz2020multi,Tan_2021_ICCV,Wang_2021_ICCV,li2020bridging} are still able to train the Transformer on small datasets ($<$10k training samples). Nevertheless, these approaches are limited by the quality of the pre-trained features and could be biased towards the task they were trained on (which is generally supervised).

\noindent\textbf{Pre-trained Transformers}. Video-based pre-training has proven to work best for video classification tasks~\cite{wei2022masked, wang2022long}, maybe due to the distribution gap, as pre-training only on images does not provide any motion cues. Nevertheless, image-based pre-training may provide stronger spatial features, given the higher variability of appearance and number of categories (providing better semantics regarding objects) compared to video datasets (where many consecutive frames contain similar appearance statistics). It is for this reason that we find many VTs leveraging image pre-trained Transformers, commonly on some ImageNet variant~\cite{imagenet_cvpr09, ridnik2021imagenetk}. This is generally done in one of two fashions. On the one hand, some works~\cite{arnab2021vivit,Girdhar_2021_ICCV, yan2022multiview, neimark2021video, yang2021associating, bulat2021space} leverage a pre-trained image Transformer (generally ViT~\cite{dosovitskiy2021an} or Swin~\cite{liu2021swin}) as the spatial stage of an abrupt hierarchical VT, training the later temporal layers from scratch. On the other hand, a pre-trained image Transformer can be directly adapted by using 3D patches to factor time in (as well as inflating linear embeddings and positional biases to account for this change) before fine-tuning for video~\cite{liu2021swinvideo, liu2022swinv2, fan2021multiscale, wei2022masked}. Finally, some object-centric approaches (e.g.,~\cite{wang2020end,zhou2021hopper,wang2021end, meinhardt2022trackformer}) leverage pre-trained Transformer-based object detectors (e.g., DETR~\cite{carion2020end}) as initialization. 

\vspace{-0.15cm}
\subsection{Self-supervised pretext tasks}
\label{sec:pretext_tasks} 
Harvesting large annotated datasets incurs additional labeling costs, and may further influence towards human-induced annotation biases~\cite{rodrigues2018deep, chen2021understanding}. \textit{Self-supervised learning} (SSL) has been recently shown to alleviate data needs for an equivalent supervision-based pre-training (e.g.,~\cite{tong2022videomae, wei2022masked}), while providing more robust~\cite{hendrycks2019using} and general features~\cite{kim2021selfreg, grill2020bootstrap, chen2021exploring}. Despite the great success of SSL in both NLP~\cite{devlin2019bert} and Image Transformers~\cite{he2022masked}, they are not as widespread in the video domain, which could be attributed to the large costs involved in such a process. Therefore, we next analyze the benefits and limitations of SSL for VTs, so as to motivate further research in this area. 

Traditional time-related pretext tasks (see~\cite{schiappa2022self} for a complete review) are rarely used in the context of VTs. They are generally limited to shuffling the input sequence, and training the network to correctly reorder it~\cite{li2020hero, yun2022matter, guo2022cross, truong2022direcformer}, effectively learning coherent temporal dynamics. However, these have not found as much success~\cite{schiappa2022self} compared to (1) \textit{Instance-based learning} and (2) \textit{Masked Token Modeling} (MTM), which we explore next. The former learns sequence-level representations that are invariant to different spatiotemporal perturbations, whereas the latter mask individual token representations of the input and tries to reconstruct them. 

\subsubsection{Instance-based learning}
\label{sec:contrastive_learning}
Instance-based approaches for VTs leverage contrastive losses (generally \textit{InfoNCE}~\cite{oord2018representation}) to make representations of whole sequences invariant to certain augmentations. These approaches define one anchor $\textbf{x}$, a positive sample $\textbf{x}^+$ and a set of $G$ negative samples to contrast against $\{\textbf{x}^-_g\}$, where $1 \leq g \leq G $. These tasks force representations for the positive pair to be similar, while it drives apart representations for the negative (dissimilar) pairs. Minimizing InfoNCE can be seen as maximizing a lower bound on the mutual information between $\textbf{x}$ and $\textbf{x}^+$~\cite{oord2018representation}. These losses have also been used in the context of cross-modal matching for VTs, as explored in~\cite{multimodalSurvey2}. Hence, we only briefly discuss them in the context of video retrieval in the Supplementary and focus here on their uses for video only.

\noindent\textbf{View mining}. Positive pairs tend to be differently augmented versions (generally regarded as \textit{views}) of the same sample. In VTs (and generally for video), it is customary to apply spatial augmentations (e.g., random cropping, color jittering, horizontal flips, or Gaussian blur) consistently through time (i.e., applying the same augmentation to all frames~\cite{qian2021spatiotemporal}). By aligning multiple views' representations, the model learns to be invariant to such perturbations. However, spatial augmentations alone are not enough for video SSL~\cite{schiappa2022self}, and generating temporal views needs to be done carefully. For instance, reversing or randomly shuffling a clip may make the model invariant to temporal causality. In VTs (similar to other video literature~\cite{feichtenhofer2021large}), it is common to use multiple temporal~\cite{wu2021towards, Sun_2021_ICCV, wang2022long, guo2022cross} or spatiotemporal~\cite{ranasinghe2022self,Patrick_2021_ICCV} crops of a given video to form the positive pairs, with varying temporal spans~\cite{wang2022long, Patrick_2021_ICCV, ranasinghe2022self} and frame-rates (i.e, speed)~\cite{ranasinghe2022self, wang2022long}, whereas negatives are sampled among the rest of training videos. Learning invariance to such changes may be useful for high-level tasks where a wide abstract understanding of the video is enough. Nevertheless, this could disregard local view-dependent information in favor of redundant cross-view information~\cite{purushwalkam2020demystifying}, favoring the formation of appearance-biased features (see~\cref{sec:discussion_training}). In order to tackle this, some VTs use multiple global and local potentially overlapping views as positives~\cite{ranasinghe2022self, Patrick_2021_ICCV, wang2022long}, which may allow for better modeling of part-whole relationships. Intuitively, this forces global views to preserve information in the local ones, while maintaining global context awareness in local views. Alternatively, the alignment task can be relaxed, skewing away from learning absolute invariance to changes between views. One example is seen in~\cite{Sun_2021_ICCV}, which conditions alignment on the temporal shift between crops. Another example is seen in VT works introducing asymmetries in the networks computing the different views' representations: using additional predictors~\cite{wang2022long}, momentum encoders~\cite{ranasinghe2022self, wang2022long} (originally proposed in~\cite{He_2020_CVPR}, are believed to behave as network ensembles~\cite{Caron_2021_ICCV}), and even CNNs~\cite{guo2022cross} (probably helping infuse some locality bias from CNN representations into the Transformer). Introducing some of these asymmetries has indeed been found to boost downstream performance on image~\cite{tian2021understanding} and video tasks~\cite{feichtenhofer2021large}. Intuitively, they may be relaxing the alignment task into a more predictive setting, allowing features to be aware of context, not so much invariant to it. 

\noindent\textbf{Negative sampling}. One crucial limitation of contrastive approaches is their need for large negative sets~\cite{chen2020simple}. These are generally mined from the batch, which can be very limiting in the context of full video representations, as it may not always be possible to hold enough different instances in a batch. VTs tackle this through large memory banks that store representations of past batches~\cite{wang2022long,shao2021temporal,Liu_2021_Hit} (which may further serve as regularizers, due to storing sample representations from past iterations produced by the same model with slightly different weights) or through hard negative mining (forcing the network to learn small nuances in the views by trying to separate somewhat similar samples, measured by feature representation distances~\cite{lee2021parameter,Li_2021_CVPR}). Finally, we also find works dropping negatives altogether. One example is seen in~\cite{ranasinghe2022self}, which formulates learning as instance-based classification, where every positive view has to be classified in the same pseudo-class. Another example is the work in~\cite{yu2021learning}, where multiple sparse views of the input are independently processed and the aggregated prediction is used to distill the consensus into single view streams. 

\subsubsection{Masked Token Modeling}
\label{sec:mtm}
MTM draws inspiration from the \textit{Masked Token Prediction} task proposed in BERT~\cite{devlin2019bert}. It randomly replaces some input tokens with a learnable \texttt{[MSK]} token and the network is trained to predict (classify) the replaced tokens. This forces the Transformer to learn contextualized representations of the input. However, different from language tokens, visual tokens cannot be easily mapped to a discrete and limited-size vocabulary so as to pose MTM as a classification task. For perspective, a pixel codebook would require $255^3 \approx 16\mathrm{M}$ distinct elements, whereas BERT employed a vocabulary of 30K. Furthermore, posing it as a classification task would disregard the distance of the prediction to the actual ground truth value, distracting the network with high-frequency details of the data which could be irrelevant. To solve this issue in the context of VTs, we roughly find three families of approaches, categorized by the type of target: (1) working at \textit{feature} level either through regression~\cite{li2020hero, cheng2020videotrm, Li_2021_CVPR, wei2022masked} or contrasting~\cite{li2020hero,lee2021parameter,contrastive2019chen,chen2019bert4sessrec}, as well as (2) \textit{quantization} of visual tokens~\cite{sun2019videobert, wang2022bevt}. Interestingly, some works have actually found success (3) regressing the original token in \textit{pixel} space~\cite{tong2022videomae, liu2022swinv2}. We also find VTs classify token contents~\cite{zhu2020actbert, wu2021towards}, but as these require manual annotations, we do not delve into them here.

\noindent\textbf{Feature-based MTM} works regress a feature-based representation of the masked tokens. This can be posed as a prediction (e.g., using an MSE loss)~\cite{li2020hero, cheng2020videotrm, Li_2021_CVPR} or as a contrastive task~\cite{li2020hero,lee2021parameter,Xiao_2022_CVPR, chen2019bert4sessrec}. The target token representation is obtained from the input embedding network (e.g.,~\cite{li2020hero,lee2021parameter,Xiao_2022_CVPR}) or from an external encoder~\cite{contrastive2019chen}. In this sense, by requiring an additional network, these models potentially incur additional compute and memory costs. In order to avert this,~\cite{wei2022masked} proposes using the HOG features of the masked region, which are cheaper to compute and can be pre-computed. Interestingly, the work of~\cite{Girdhar_2021_ICCV} uses causal masked SA instead of replacing tokens with \texttt{[MSK]}. In this sense, all tokens are tasked with solving feature-based MTM by trying to predict the next token representation (in a predictive coding setting~\cite{rao1999predictive,oord2018representation}).

\noindent\textbf{Quantization-based MTM} involves discretizing video tokens to a limited codebook, generally requiring some pre-trained network to define it. For instance, in~\cite{sun2019videobert} an S3D~\cite{bb_s3d} followed by hierarchical k-means is used for both embedding the tokens prior to the Transformer and the discrete (cluster assignation) pseudo label for the prediction, whereas in~\cite{wang2022bevt} a VQ-VAE~\cite{ramesh2021zero} is used instead, but only to generate the ground truth for the masked tokens. The use of quantization makes it possible for these models to optimize the network with a classification objective, akin to NLP counterparts. Similar to many feature-based MTM, however, these approaches also require an additional pre-trained network.

\noindent\textbf{Pixel-based MTM} directly regresses the pixel space for masked regions~\cite{tong2022videomae, liu2022swinv2}. They do not require any further networks or computing additional features, making them very simple to implement. However, pixels as targets have been argued to focus on irrelevant high-frequency details of data, which could be detrimental for high-level tasks~\cite{han2019video}. Nevertheless, this may be more nuanced and require further research, as we discuss next in~\cref{sec:discussion_training}.

\vspace{-0.15cm}
\subsection{Discussion on training strategies}
\label{sec:discussion_training}
Training stand-alone VTs requires balancing solutions to the lack of inductive biases with potentially limited computational budgets. This implies factoring in large datasets, SSL, and efficient designs while accounting for the large dimensionality of videos, properly sized clips, batches, and architectures. VTs are dominated by fully supervised training aided by \textit{large frozen CNN embeddings} (which are not so common in other fields, such as NLP), and disregard pre-training of Transformer layers. On the one hand, long-range temporal interactions provided by Transformers boost CNN's performance in many application scenarios\cite{purwanto2019extreme,Pashevich_2021_ICCV,BMT_Iashin_2020,pramono2019hierarchical,camgoz2020sign,kondo2020lapformer,gavrilyuk2020actor,Tan_2021_ICCV,lin2020bi}. On the other hand, the embedding network provides initial representations and dimensionality reduction, alleviating Transformers' training limitations. Nevertheless, this approach caps the potential of Transformers to model spatiotemporal interactions and depends on the transferability of the pre-trained embedding features (e.g., distribution or task shift). 

The canonical pre-training then fine-tuning paradigm acts as a \textit{smart form of initialization}. Skewing from it may allow avoiding catastrophic forgetting~\cite{mccloskey1989catastrophic} while achieving more generalizable features. For example, by incorporating self-supervised auxiliary losses during fine-tuning, as done by some VTs~\cite{Girdhar_2021_ICCV, yun2022matter}. In~\cite{li2020hero} a training schedule is proposed that samples a different (self-)supervised task at each batch, showing improved results for video retrieval as more tasks are added. Alternatively, recent works (e.g.,~\cite{wang2022bevt, Girdhar_2022_CVPR, zhang2021co}) deviate from the trend of image-based pre-training and achieve promising results by optimizing for image and video tasks in a joint manner.

SSL is not as widespread for VTs when compared to supervised or image-based initialization. However, we believe VTs could greatly benefit from large-scale unlabeled videos, as well as from the inductive biases SSL provides. In this sense, we see great promise in the current developments on SSL that are better suited to train visual Transformers. MTM could be seen from the lens of generative-based pre-training as it bears great resemblance with CNN-based inpainting~\cite{pathak2016context}. We believe that the success of MTM may be attributable to Transformers providing explicit granularity through tokenization. In order to \textit{conquer} the complex global task of inpainting large missing areas of the input, MTM \textit{divides} it into smaller local predictions. This is especially true in both 2D- and 3D-based patch tokenization approaches~\cite{wei2022masked, tong2022videomae, wang2022bevt}. Intuitively, the model needs an understanding of both global appearance and motion semantics as well as low-level local patterns to properly gather the necessary context to solve token-wise predictions. This may allow VTs to learn more holistic representations (i.e., better learning of part-whole relationships). Nevertheless, given the high redundancy of videos it could be trivial for the network to find shortcuts, borrowing information from neighboring spatiotemporal positions instead. It has been found that high masking ratios (e.g., 40\%-60\% in MaskFeat~\cite{wei2022masked} or even 75\%-90\% in VideoMAE~\cite{tong2022videomae}), especially compared to NLP (15\%-20\% in BERT~\cite{devlin2019bert}) or images (20\%-50\% in MAE~\cite{he2022masked}), indeed force the network to capitalize on global relationships of the data, as seen by improved performance on high-level semantic tasks (see~\cref{sec:performance_comparsion}). Furthermore, ablations in~\cite{wei2022masked,tong2022videomae} suggest that the masking strategy can also impact the learning of such shortcuts, showing that masking blocks of tokens in space consistently through time helps to avoid them. Regarding the choice of target for MTM, quantized and feature-based seem to work best for video~\cite{wei2022masked} (albeit requiring an additional pre-trained network). Pixel-based provide the cheapest target but are generally discarded arguing they may fixate on irrelevant high-frequency details of data. However, the generally used MSE loss has been shown to disregard such details~\cite{zhang2018unreasonable,oprea2020review, pathak2016context}, so further research may be needed. Finally, we highlight HOG features, which provide the best compute/performance trade-off (see~\cref{sec:performance_comparsion}), as they are cheap to compute while providing partial invariance to various deformations. 

Despite requiring large batches for negative mining, instance-based contrastive approaches have consistently shown potential for high-level video tasks~\cite{schiappa2022self}. 
By contrasting differently spatiotemporal augmented views, the network learns invariance to appearance perturbations, spatial scale, and occlusions, as well as changes of perspective or illumination naturally present in video~\cite{gordon2020watching,wang2015unsupervised}. However, the model may also become invariant to temporal translation and deformation, effectively disregarding fine-grained motion dynamics and biasing it towards appearance-based static cues (which is enough for appearance-biased datasets  (e.g., UCF101 or Kinetics) where the presence of certain objects may suffice to predict an action class~\cite{huang2018makes,buch2022revisiting}). As we discussed in~\cref{sec:contrastive_learning}, re-introducing motion modeling requires relaxing the alignment task through network asymmetries (e.g.,~\cite{guo2022cross, ranasinghe2022self}) or careful sampling techniques (e.g.,~\cite{Patrick_2021_ICCV, wang2022long}) to balance part-whole relationship learning. However, compared to MTM, it is easier for these approaches to overlook low-level view-dependent temporal information, crucial for proper motion modeling~\cite{yuan2022contextualized,qian2021spatiotemporal}. 

In this context, we see promise in combining instance-based contrastive learning and MTM, both in multi-task scenarios~\cite{contrastive2019chen,Sun_2021_ICCV,wu2021towards} as well as feature-based contrastive MTM~\cite{li2020hero,lee2021parameter,Xiao_2022_CVPR,contrastive2019chen} (as opposed to direct regression). These latter could combine the holistic feature learning of MTM while potentially accounting for the uncertainty of modeling missing information through contrastive losses (as the model is not tasked with explicit hard prediction~\cite{han2019video}). For instance, in~\cite{li2020hero}, this alternative is found to outperform L2 feature regression in the context of video-moment retrieval. These approaches remain, to the best of our knowledge, unexplored in the context of patch-based tokenization, where the cardinality of the negative set would be much larger than for instance-based approaches (allowing for many hard negatives from the same sequence as well as easy negatives from all other sequences in the batch). Nevertheless, it is still unclear what these models are actually learning, so future research is needed for proper interpretation of SSL features, which currently are mostly evaluated based on their success on downstream performance~\cite{schiappa2022self, jing2020self}. 

\vspace{-0.25cm}
\section{Performance on video classification}
\label{sec:application}
The task of video classification has attracted the most research in Transformers for video, given the generality of the task and availability of large datasets for training and evaluation, things that allow for more comprehensive performance analysis. Next, we overview the particularities of video classification  (\cref{sec:performance_video_classification}) and then analyze VT state-of-the-art performance on it~(\cref{sec:performance_comparsion}).

\subsection{Video classification}
\label{sec:performance_video_classification}
Video classification aims to predict the class of a given input sequence of frames. For the task, a VT will encode descriptive high-level global representations of a given sample. Then, some linear layers followed by a softmax provide a class-score probability distribution. The category with maximum probability should match the ground-truth class. VTs competing to become state-of-the-art in classification tend to be standalone (i.e., use minimal embedding), and thus will be the ones we cover. Next, we present the benchmarking datasets, experimental protocols, and details on the sampling of the clips.

\noindent\textbf{Evaluation datasets}. The most popular dataset is the large-scale \textit{Kinetics-400} (K400)~\cite{carreira2017quo}, consisting of 306K 10-second clips and 400 manually annotated human actions categories with at least 400 examples per class. Kinetics-600 (K600) -- an extension of K400 with 495K clips and 600 classes -- is only used for pre-training, but not for evaluation. K400/K600 are however known to be appearance-biased~\cite{yun2022matter}. To better assess the modeling of more complex temporal dynamics, most works are also evaluated on \textit{Something-Something v2} (SSv2)~\cite{goyal2017something, mahdisoltani2018effectiveness}. SSv2 is an egocentric human action dataset where some of the categories can only be distinguished by having an understanding of the arrow of time (e.g., ``Moving [sth] away from [sth]'' vs ``Moving [sth] closer to [sth]''). SSv2 consists of 220K videos of duration ranging from 2 to 6 seconds and 174 fine-grained categories.

\noindent\textbf{Experimental protocols}. We find two learning protocols being followed: training from scratch or pre-training the model first. \textit{Training from scratch} is rarer because of the size of the models (especially, their larger variants). When following \textit{pre-training}, the weights learned during the first stage are used to initialize the model that is to be trained in the downstream dataset/task. Common pre-training strategies for video classification are (a) image-based pre-training on ImageNet, and either (b) supervised or (c) a self-supervised video pre-training (generally on video datasets larger than the downstream one, e.g., K600 for evaluating K400 and K400/K600 for SSv2). After initialization, the models are trained on the downstream dataset, fine-tuning existing weights and adapting new ones.

\noindent\textbf{Clip sampling}. Models are fed with trimmed video clips. These are relatively short, with a number of frames $T'$ typically 8 to 64 frames and fixed spatial resolution $S' = H'\times W'$ pixels (often $H' = W' = 224$, hence shortened to $S' = 224\textsuperscript{2}$, see ``Input'' in~\cref{tab:perf_k400,tab:perf_ssv2}). However, to make sense of these numbers, and especially $T'$, it is crucial to consider the temporal stride $\tau$, i.e., the step between clip frames when sampling them from the original video. A larger $\tau$ extends the temporal span of the clip w.r.t. the video without incurring extra computation costs, while also reducing the redundancy among otherwise nearby sampled frames. For instance, with $\tau = 4$ and $T' = 64$, a clip covers a temporal span equivalent to a densely ($\tau=1$) sampled clip of 256 frames. Importantly, $\tau$ must be chosen factoring in the temporal resolution of videos (e.g., $\sim$25 FPS in K400) and the fine-grained motion information one is willing to sacrifice in favor of context. 

\noindent\textbf{Views}. The clips generated can be regarded as \textit{temporal views} (related to the views described in~\cref{sec:pretext_tasks}, which are used for some methods during pre-training). During training, one temporal view per video is gathered at a random temporal position. These are constructed with fixed size $T' \times S'$ and stride $\tau$. For inference, most models follow a multi-view approach: the classification decision for the video is achieved by averaging the prediction obtained from different spatiotemporal crop views. 

\vspace{-0.15cm}
\subsection{Comparison among state-of-the-art models}
\label{sec:performance_comparsion}
To draw comparisons we consider the factors defined by the columns of~\cref{tab:perf_k400,tab:perf_ssv2}. Among those, the most interesting one to study is perhaps the pre-training strategy, which will drive the rest of the section, separately analyzing K400 and SSv2.

\noindent\textbf{K400: training from scratch}. Doing so, we only find MViTv2~\cite{li2022mvitv2} and its predecessor MViT~\cite{fan2021multiscale}. The main difference between the two is the inclusion of an extra residual pooling connection and the use of relative positional encoding. With these, ``MViTv2-B 32@3'' (82.9\%) performs better than its older counterpart ``MViT-B 32@3'' by +2.7\%. In fact, it also surpasses ``MViT-B 64@4'' -- which has an increased temporal receptive field (2.6x) -- by +1.7\%. Later in~\cite{wei2022masked}, the same authors explored different initialization strategies and showed overfitting of the larger variants of MViTv2 when not using effective initialization. This can be seen for ``MViTv2-L↑'', with an increased spatial resolution (from 224 to 312) and compute (from 51 MP to 218 MP), performing worse (-0.7\%) than ``MViTv2-B 32@3'. Although this is to be expected, the smaller variants are still able to learn from scratch successfully -- as we will see later, even better than 3D ConvNets. Given the need for pre-training of larger models, we next discuss the two most popular strategies in the context of K400 and demonstrate its large positive effect (e.g., ``MViTv2↑ 32@3'' boosts its results from 82.2\% to 85.3\% by leveraging image-based pre-training).

\begin{table}[t]
\setlength{\tabcolsep}{2pt}
\adjustbox{max width=\columnwidth}{
\begin{threeparttable}
\begin{tabular}{|c|c|lr|c|c|c|c|}
\hline
\hfill                                                       & \textbf{Pre-train}                           & \textbf{Name}                       & \textbf{Ref.}                          & \textbf{Input}                          & \textbf{TF} × $v_t$ × $v_s$ & \textbf{MP.} & \textbf{Acc.} \\ \hline
\multirow{4}{*}{\rotatebox[origin=c]{90}{\textbf{ConvNets}}}           & \multirow{2}{*}{-}                 & SlowFast (R101+NL)         & \cite{bb_slowfast}            & 16@8 × 256\textsuperscript{2}  & 0,23 × 10 × 3      & 60  & 79,8 \\
                                   \cline{3-8}
                                                             &                                    & X3D-XXL                    & \cite{feichtenhofer2020x3d}   & 16@5 × 312\textsuperscript{2}  & 0,19 × 10 × 3      & 20  & 80,4 \\
                                                             \cline{2-2}\cline{3-8}
                                                             & \multirow{2}{*}{\begin{tabular}[c]{@{}c@{}}IG65\\ (video)\end{tabular}}            & R(2+1)D-152                & \cite{ghadiyaram2019large}    & 32@1 × 128\textsuperscript{2}  & 0,25 × 10 × 1      & 118 & 81,3 \\
                                   \cline{3-8}
                                                             &                                    & ir-CSN-152                 & \cite{tran2019video}          & 32@2 × 224\textsuperscript{2}  & 0,10 × 10 × 3      & NA  & \textbf{82,6} \\ \hline\hline
\multirow{6}{*}{\rotatebox[origin=c]{90}{\textbf{Scratch}}}           & \multirow{6}{*}{-}                 & MViT-S                     & \multirow{3}{*}{\cite{fan2021multiscale}}      & 16@4 × 224\textsuperscript{2}  & 0,03 × 5 × 1       & 26  & 76,0 \\
                                                             &                                    & MViT-B                     &       & 32@3 × 224\textsuperscript{2}  & 0,17 × 5 × 1       & 37  & 80,2 \\
                                                             &                                    & MViT-B                     &       & 64@4 × 224\textsuperscript{2}  & 0,46 × 3 × 3       & 37  & 81,2 \\
                                   \cline{3-8}
                                                             &                                    & MViTv2-S                   & \multirow{2}{*}{\cite{li2022mvitv2}}           & 16@4 × 224\textsuperscript{2}  & 0,06 × 5 × 1       & 35  & 81,0 \\
                                                             &                                    & MViTv2-B                   &            & 32@3 × 224\textsuperscript{2}  & 0,23 × 5 × 1       & 51  & \textbf{82,9} \\
                                                             &                                    & MViTv2-L↑                  & \cite{wei2022masked}          & 32@3 × 312\textsuperscript{2}  & 2,06 × 5 × 3       & 218 & 82,2 \\ \hline\hline
\multirow{27}{*}{\rotatebox[origin=c]{90}{\textbf{Image pre-tr. (I)}}} & \multirow{3}{*}{IN}                & UniFormer-B                & \multirow{2}{*}{\cite{li2022uniformer}}        & 16@4 × 224\textsuperscript{2}  & 0,10 × 4 × 1       & 50  & 82,0 \\
                                                             &                                    & UniFormer-B                &         & 32@4 × 224\textsuperscript{2}  & 0,26 × 4 × 3       & 50  & 83,0 \\
                                                             \cline{3-8}
                                                             & \multirow{17}{*}{IN21}                & Swin-B                     & \cite{liu2021swinvideo}       & 32@2 × 224\textsuperscript{2}  & 0,28 × 4 × 3       & 88  & 80,6 \\
                                                             \cline{2-2}\cline{3-8}
                                                             &                                    & SCT-L                      & \cite{zha2021shifted}         & 24@10 × 224\textsuperscript{2} & 0,34 × 4 × 3       & 60  & 83,0 \\
                                                             \cline{3-8}
                                                             &                                    & Swin-B                     & \multirow{2}{*}{\cite{liu2021swinvideo}}       & 32@2 × 224\textsuperscript{2}  & 0,28 × 4 × 3       & 88  & 82,7 \\
                                                             &                                    & Swin-L↑                    &        & 32@2 × 384\textsuperscript{2}  & 2,11 × 10 × 5      & 200 & 84,9 \\
                                                             \cline{3-8}
                                                             &                                    & TS                         & \cite{bertasius2021spacetime} & 8@16 × 224\textsuperscript{2}  & 0,20 × 1 × 3       & 121 & 78,0 \\
                                                             \cline{3-8}
                                                             &                                    & ViViT-L-FE                 & \cite{arnab2021vivit}         & 32@2 × 224\textsuperscript{2}  & 3,98 × 1 × 3       & 352 & 81,7 \\
                                                             \cline{3-8}
                                                             &                                    & VTN-3 (Aug)                & \cite{neimark2021video}       & 250@1 × 224\textsuperscript{2}   & 4,22 × 1 × 1       & 114 & 79,8 \\
                                                             \cline{3-8}
                                                             &                                    & DirecFormer                & \cite{truong2022direcformer}  & 8@32 × 224\textsuperscript{2}  & 0,20 × 1 × 3       & 124 & 82,8 \\
                                                             \cline{3-8}
                                                             &                                    & Mformer                    & \multirow{2}{*}{\cite{patrick2021keeping}}     & 96@3 × 224\textsuperscript{2}  & 0,96 × 10 × 3      & NA  & 81,1 \\
                                                             &                                    & Mformer↑                   &      & 64@4 × 336\textsuperscript{2}  & 1,19 × 10 × 3      & NA  & 80,2 \\
                                                             \cline{3-8}
                                                             &                                    & X-ViT                      & \multirow{2}{*}{\cite{bulat2021space}}         & 16@1 × 224\textsuperscript{2}  & 0,28 × 1 × 3       & 92  & 80,2 \\
                                                             &                                    & X-ViT                      &          & 16@1 × 224\textsuperscript{2}  & 0,28 × 2 × 3       & 92  & 80,7 \\
                                                             \cline{3-8}
                                                             &                                    & MTV-B                      & \multirow{2}{*}{\cite{yan2022multiview}}       & 32@2 × 224\textsuperscript{2}  & 0,4 × 4 × 3        & 310 & 81,8 \\
                                                             &                                    & MTV-B↑                     &        & 32@2 × 320\textsuperscript{2}  & 0,96 × 4 × 3       & 310 & 82,4 \\
                                                             \cline{3-8}
                                                             &                                    & RViT-XL                    & \cite{yang2022recurring}      & 64@NA × 224\textsuperscript{2} & 11,90 × 3 × 3      & 108 & 81,5 \\
                                                             \cline{3-8}
                                                             &                                    & MViTv2-S                   & \multirow{2}{*}{\cite{wei2022masked}}          & 16@4 × 224\textsuperscript{2}  & 0,07 × 10 × 1      & 36  & 82,6 \\
                                                             &                                    & MViTv2-L↑                  &           & 32@3 × 312\textsuperscript{2}  & 2,06 × 5 × 3       & 218 & 85,3 \\
                                                             &                                    & MViTv2-L↑                  & \cite{li2022mvitv2}           & 40@3 × 312\textsuperscript{2}  & 2,83 × 5 × 3       & 218 & 86,1 \\
                                                             \cline{2-2}\cline{3-8}
                                                             & (IN-21 + P) & \multirow{2}{*}{SwinV2-G↑}                  & \multirow{2}{*}{\cite{liu2022swinv2}}          & \multirow{2}{*}{8@NA × 384\textsuperscript{2}}  & \multirow{2}{*}{NA × 4 × 3}         & \multirow{2}{*}{3 K} & \multirow{2}{*}{\textbf{86,8}} \\
                                                             & (SSL)            &          &  &         &  &  & \\
                                                             \cline{2-2}\cline{3-8}
                                                             & \multirow{5}{*}{JFT}               & ViViT-L-FE                 & \multirow{2}{*}{\cite{arnab2021vivit}}         & 32@2 × 224\textsuperscript{2}  & 3,98 × 1 × 3       & 352 & 83,5 \\
                                                             &                                    & ViViT-H                    &        & 32@2 × 224\textsuperscript{2}  & 3,98 × 4 × 3       & 352 & 84,9 \\
                                                             \cline{3-8}
                                                             &                                    & MTV-L                      & \multirow{2}{*}{\cite{yan2022multiview}}       & 32@2 × 224\textsuperscript{2}  & 1,50 × 4 × 3       & NA  & 84,3 \\
                                                             &                                    & MTV-H                      &        & 32@2 × 224\textsuperscript{2}  & 3,71 × 4 × 3       & NA  & 85,8 \\
                                                             \cline{3-8}
                                                             &                                    & TokenLearner & \cite{ryoo2021tokenlearner}   & 64@1 × 256\textsuperscript{2}  & 4,08 × 4 × 3       & 450 & 85,4 \\ \hline\hline
\multirow{10}{*}{\rotatebox[origin=c]{90}{\textbf{Video pre-tr. (V)}}} & \multirow{8}{*}{K400 (SSL)}        & LSTCL (Swin-B*)            & \cite{wang2022long}           & 16@8 × 224\textsuperscript{2}  & 0,36 × 5 × 1       & 88  & 81,5 \\
                                                            \cline{3-8}
                                                             &                                    & MaskFeat-S                 & \multirow{4}{*}{\cite{wei2022masked}}          & 16@4 × 224\textsuperscript{2}  & 0,07 × 10 × 1      & 36  & 82,2 \\
                                                             &                                    & MaskFeat-L↑                &          & 32@3 × 312\textsuperscript{2}  & 2,06 × 5 × 3       & 218 & 86,3 \\
                                                             &                                    & MaskFeat-L↑                &           & 40@3 × 312\textsuperscript{2}  & 2,83 × 4 × 3       & 218 & 86,4 \\
                                                             &                                    & MaskFeat-L↑↑               &           & 40@3 × 352\textsuperscript{2}  & 3,79 × 4 × 3       & 218 & 86,7 \\
                                                            \cline{3-8}
                                                             &                                    & VideoMAE (ViT-B)           & \multirow{3}{*}{\cite{tong2022videomae}}       & 16@4 × 224\textsuperscript{2}  & 0,18 × 5 × 3       & 87  & 80,9 \\
                                                             &                                    & VideoMAE (ViT-L)           &        & 16@4 × 224\textsuperscript{2}  & 0,60 × 5 × 3       & 305 & 84,7 \\
                                                             &                                    & VideoMAE↑ (ViT-L)          &        & 32@4 × 320\textsuperscript{2}  & 3,96 × 5 × 3       & 305 & 85,8 \\
                                                             \cline{2-2}\cline{3-8}
                                                             & \multirow{2}{*}{K600 (SSL)}        & MaskFeat-L                 & \multirow{2}{*}{\cite{wei2022masked}}          & 16@4 × 224\textsuperscript{2}  & 0,34 × 10 × 1      & 218 & 85,1 \\
                                                             &                                    & MaskFeat-L↑↑               &           & 40@3 × 352\textsuperscript{2}  & 3,79 × 4 × 3       & 218 & \textbf{87,0} \\ \hline\hline
\multirow{4}{*}{\rotatebox[origin=c]{90}{\textbf{I+V}}}               & IN +                    & \multirow{2}{*}{SVT (TS)}                   & \multirow{2}{*}{\cite{ranasinghe2022self}}     & 8@NA × 224\textsuperscript{2} +                     & \multirow{2}{*}{0,20 × 1 × 3}       & \multirow{2}{*}{121} & \multirow{2}{*}{78,1} \\                                                                          
& K400 (SSL)                    &                    &      & 64@NA × 96\textsuperscript{2}                      &        &  &  \\
                                                            \cline{2-2}\cline{3-8}
                                                             & IN (SSL) + & BEVT                       & \multirow{2}{*}{\cite{wang2022bevt}}           & 16@NA × 224\textsuperscript{2} & 0,28 × 4 × 3       & 88  & 80,6 \\
                                                             &     K400 (SSL)                                & BEVT (Dall-E tknzr.)       &           & 16@NA × 224\textsuperscript{2} & 0,28 × 4 × 3       & 88  & 81,1 \\
\hline
\end{tabular}
\begin{tablenotes}
\item ↑: increased spatial resolution.
\item ``IN21 + P'': extension of IN21 with a private private dataset (70M images in total).
\end{tablenotes}
\end{threeparttable}}
\caption{Accuracy (top-1) on Kinetics-400. ``Input'': temporal and spatial size of the views; ``TF'': TFLOPs; $v_t $ and $v_s$: the number of temporal and spatial views; ``MP'': parameters ($\times10^6$); and ``Pre-train'': pre-training strategy.} 
\label{tab:perf_k400}
\end{table}

\noindent\textbf{K400: image pre-training}. The majority of VTs pre-train on either ImageNet-1K (``IN''), ImageNet-21K (``IN21''), or JFT-300M (``JFT''). IN and IN21 consist of 1K and 21K classes and over 1.2M and 14M examples respectively, whereas JFT is a non-public dataset with 300M multi-label images and 18,291 non-mutually-exclusive labels. Other works have been using their own image datasets or extending public ones. For instance, ``Video-SwinV2-G''~\cite{liu2022swinv2} (86.8\%), being the best performing model, extended IN21 (14M images) with a private dataset of images (``P'' in Tab.~\ref{tab:perf_k400}), totaling 70M samples. Close second is ``MViTv2-L↑ 40@3''~\cite{li2022mvitv2} (86.1\%), with weights pre-trained exclusively on IN21 while only dropping by -0.7\% with respect to the first one, but with 14x fewer parameters. The third is ``MTV-H''\cite{yan2022multiview} (85.8\%), this one pre-trained on JFT with 300M images. Unfortunately, in this work, the authors used JFT to pre-train their largest models (``MTV-L'' and ``MTV-H'') and IN21 to train ``MTV-B''/``MTV-B (320)'', therefore not validating the actual contribution of JFT w.r.t. IN-21K for any of the variants; making difficult to discern the actual contribution of the model scaling. Also, TokenLearner~\cite{ryoo2021tokenlearner} completely relies on JFT for all the experiments, with its best model ``TokenLearner 16at18 (L/10)'' (85.4\%) coming fourth. It was ViViT~\cite{arnab2021vivit} that showed how the same model variant trained on JFT, ``ViViT-L (JFT)'' (83.5\%), was considerably improving upon the same variant pre-trained on IN-21K (``ViViT-L''), by +1.8\%. It is, hence, of great merit that ``MViTv2-L↑ 40@3'' (86.1\%) still surpasses, respectively by +0.3\% and +0.7\%, the results of ``MTV-H'' and ``TokenLearner 16at18 (L/10)''. It is true that compared to those, MViTv2 variant utilizes larger spatial (312\textsuperscript{2}, versus 224\textsuperscript{2} and 256\textsuperscript{2} pixels) and temporal receptive field (120 vs 64 frames), but the number of TFLOPs and the amount of pre-training data to process are still both lower: 14 MP versus 300 MP in JFT for MTV and TokenLearner, and 42 TFLOPs versus 47 and 48 TFLOPs.

In terms of cost-effectiveness, we find ``UniFormer-B''~\cite{li2022uniformer} (83.0\%), ``SCT-L'' (83.0\%)~\cite{zha2021shifted}, ``Direcformer''~\cite{truong2022direcformer} (82.8\%) -- this one based on \cite{bertasius2021spacetime}-, and ``MViTv2-S'' (82.6\%)~\cite{wei2022masked}. These models only suffer a drop between -3.1\% and -3.3\% of accuracy but between 10x and 70x less FLOPs w.r.t. ``MViTv2-L↑ 40@3''.

\noindent\textbf{K400: video (self-supervised) pre-training}. An emerging trend in the literature is to perform all SSL pre-training, fine-tuning and evaluation on the same dataset~\cite{wei2022masked, tong2022videomae, wang2022long, wang2022bevt}. ``MaskFeat-L↑ 40@3''~\cite{wei2022masked} reaches 86.4\%, thus showing the contribution of MaskFeat (SSL) pre-training compared with supervised training on the same architecture, i.e., MViTv2, by +0.3\%. That result of MaskFeat is also only -0.2\% behind the best image-based pre-trained model (i.e., ``Video-SwinV2-G''). Then, ``MaskFeat-L↑↑ 40@30'' by switching K400 with K600 and slightly increasing the spatial resolution from 312 to 352 (still lower than 382 of ``Video-SwinV2-G''), the model obtains state-of-the-art results (87\%), outperforming any of the image pre-trainings. VideoMAE~\cite{tong2022videomae} comes second in this category consisting of a ViT backbone with 3D inflation of the patch embeddings. This outperforms all image-based pre-trained models, except for ``Video-SwinV2-G''. Thus it seems learning motion priors during pre-training has a very positive effect on performance when targeting video classification.

\noindent\textbf{K400: ConvNets}. For the sake of completeness, we compare VTs to 3D ConvNets, which were state-of-the-art right until VTs managed to surpass them. See how ``MViTv2-S'' (81.1\%) trained from scratch, exceeds the performance of comparable ConvNets: ``SlowFast R101+NL'' (79.8\%) and ``X3D-XXL'' (80.4\%). This might be attributable to the higher temporal fidelity of MViTv2 being more profitable than extra context -- at least on short videos. The number of views for testing was also higher for both (30 versus 5 in '' MViTv2-S'' ). Nonetheless, it also consumes 18x - 22x fewer TFLOPs and works on a smaller spatial resolution (224 only, versus 256 or 312). Switching to ``MViTv2-B 32@4'' (82.9\%), we see how trained from scratch this model does better than ConvNets pre-trained on the very-large weakly-annotated video dataset IG65 (i.e., `R(2+1)D-152''~\cite{ghadiyaram2019large} (81.3\%) and ``ir-CSN-152''~\cite{tran2019video} (82.6\%)), even when using half the views.

Moving to the study of SSv2, we found none of the works train from scratch. Another thing to note is the number of temporal views used because of the shorter duration of SSv2 videos compared to Kinetics. Despite that, the temporal dynamics are harder to capture as we will see next.

\noindent\textbf{SSv2: image pre-training}. Although less common than for K400, there are works that pre-train on image datasets. Among the ones using IN, ``DirecFormer''~\cite{truong2022direcformer} (64.9\%) is the one performing the best. It surpasses its own backbone (``TS''~\cite{bertasius2021spacetime}) in all the variants by forcing the learning of temporal order of shuffled input frames via auxiliary SSL. ``TIME''~\cite{yun2022matter} is another one using auxiliary SSL ablated with different VT backbones. This one, not so much competing in performance with larger model variants, still points out the effectiveness of temporal guidance for image-based pre-trained models when transferred to the downstream video task. Finally, trained on IN-21K, ``X-ViT'' (66.4\%)~\cite{bulat2021space} is the absolute winner in this category. Unfortunately, by focusing on efficiency alone, it is not able to compete with heavier models that are supervisedly pre-trained on K400.

\begin{table}[t]
\setlength{\tabcolsep}{2pt}
\adjustbox{max width=\columnwidth}{
\begin{threeparttable}
\begin{tabular}{|c|c|lr|c|c|c|c|c|}
\hline
\hfill                                                       & \textbf{Pre-train}                           & \textbf{Name}                       & \textbf{Ref.}                          & \textbf{Input}                          & \textbf{TF} × $v_t$ × $v_s$ & \textbf{MP.} & \textbf{Acc.} \\ \hline
\multirow{2}{*}{\rotatebox[origin=c]{90}{\textbf{CN}}}           & \multirow{2}{*}{IN}                 & \multirow{2}{*}{TDN (R101)}         & \multirow{2}{*}{\cite{bb_slowfast}}            & 8@1 × 256\textsuperscript{2} + & \multirow{2}{*}{0,2 × 1 × 3}      & \multirow{2}{*}{198}  & \multirow{2}{*}{\textbf{69,6}} \\
           &                 &        &            & 16@1 × 256\textsuperscript{2}  &      &  & \\
\hline\hline
\multirow{11}{*}{\rotatebox[origin=c]{90}{\textbf{Image pre-tr. (I)}}}             & \multirow{5}{*}{IN}                                                            & TS*                         & \multirow{4}{*}{\cite{yun2022matter}}                       & 8@NA × 224\textsuperscript{2}                  & NA × 1 × 3                       & 121                  & 62,1                  \\
                                                                         &                                                                                & Mformer*                    &                       & 8@NA × 224\textsuperscript{2}                  & NA × 1 × 3                      & NA                   & 63,8                  \\
                                                                         &                                                                                & TIME (TS*)                  &                       & 8@NA × 224\textsuperscript{2}                  & NA × 1 × 3                       & 121                  & 63,7                  \\
                                                                         &                                                                                & TIME (Mformer*)             &                       & 8@NA × 224\textsuperscript{2}                  & NA × 1 × 3                       & NA                   & 64,7                  \\
                                                            \cline{3-8}
                                                                         &                                                                                & DirecFormer                 & \cite{truong2022direcformer}               & 8@32 × 224\textsuperscript{2}                  & 0,20 × 1 × 3                  & 124                  & 64,9                  \\
                                                            \cline{2-2}\cline{3-8}
                                                                         & \multirow{6}{*}{IN21}                                                          & TS                          & \multirow{3}{*}{\cite{bertasius2021spacetime}}              & 8@16 × 224\textsuperscript{2}                 & 0,20 × 1 × 3                  & 121                  & 59,5                  \\
                                                                         &                                                                                & TS-HR                       &              & 16@16 × 448\textsuperscript{2}                & 1,70 × 1 × 3                  & 121                  & 62,2                  \\
                                                                         &                                                                                & TS-L                        &               & 96@4 × 224\textsuperscript{2}                 & 2,38 × 1 × 3                  & 121                  & 62,4                  \\
                                                            \cline{3-8}
                                                                         &                                                                                & ViViT-L                     & \cite{arnab2021vivit}                      & 32@2 × 224\textsuperscript{2}                  & 3,98 × 1 × 3                  & 352                  & \textbf{65,9}                  \\
                                                            \cline{3-8}
                                                                         &                                                                                & X-ViT                       & \multirow{2}{*}{\cite{bulat2021space}}                      & 16@NA × 224\textsuperscript{2}                 & 0,28 × 1 × 3                  & 92                   & 66,2                  \\
                                                                         &                                                                                & X-ViT                       &                       & 32@NA × 224\textsuperscript{2}                 & 0,42 × 1 × 3                  & 92                   & 66,4                  \\
                                                            \hline\hline
\multirow{11}{*}{\rotatebox[origin=c]{90}{\textbf{Video pre-tr. (V)}}}             & \multirow{3}{*}{K400}                                                          & MViT-B                      & \multirow{2}{*}{\cite{fan2021multiscale}}                   & 32@3 × 224\textsuperscript{2}                  & 0,17 × 1 × 3                  & 37                   & 67,1                  \\
                                                                         &                                                                                & MViT-B                      &                   & 64@4 × 224\textsuperscript{2}                  & 0,46 × 1 × 3                  & 37                   & 67,7                  \\
                                                            \cline{3-8}
                                                                         &                                                                                & MViTv2-B                    & \cite{li2022mvitv2}                        & 32@3 × 224\textsuperscript{2}                  & 0,23 × 1 × 3                  & 51                   & 70,5                  \\
                                                            \cline{2-2}\cline{3-8}
                                                                         & \multirow{2}{*}{K600}                                                          & MViT-B                      & \multirow{2}{*}{\cite{fan2021multiscale}}                   & 32@3 × 224\textsuperscript{2}                  & 0,17 × 1 × 3                  & 37                   & 67,8                  \\
                                                                         &                                                                                & MViT-B-24                   &                   & 32@3 × 224\textsuperscript{2}                  & 0,24 × 1 × 3                  & 53                   & 68,7                  \\
                                                            \cline{2-2}\cline{3-8}
                                                                         & \multirow{2}{*}{K400 (SSL)}                                                    & LSTCL (Swin-B*)             & \cite{wang2022long}                        & 16@8 × 224\textsuperscript{2}                  & 0,36 × 5 × 1                  & 88                   & 67,0                  \\
                                    \cline{3-8}
                                                                         &                                                                                & MaskFeat-L↑                 & \cite{wei2022masked}                       & 40@3 × 312\textsuperscript{2}                  & 2,83 × 4 × 3                  & 218                  & 74,4                  \\
                                                            \cline{2-2}\cline{3-8}
                                                                         & K600 (SSL)                                                                     & MaskFeat-L↑                 & \cite{wei2022masked}                       & 40@3 × 312\textsuperscript{2}                  & 2,83 × 1 × 3                  & 218                  & 75,0                  \\
                                                            \cline{2-2}\cline{3-8}
                                                                         & \multirow{3}{*}{SSv2 (SSL)}                                                    & VideoMAE (ViT-B)            & \multirow{3}{*}{\cite{tong2022videomae}}                    & 16@2 × 224\textsuperscript{2}                  & 0,18 × 2 × 3                  & 87                   & 70,6                  \\
                                                                         &                                                                                & VideoMAE (ViT-L)            &                     & 16@2 × 224\textsuperscript{2}                  & 0,60 × 2 × 3                  & 305                  & 74,2                  \\
                                                                         &                                                                                & VideoMAE (ViT-L)            &                     & 32@2 × 320\textsuperscript{2}                  & 1,44 × 1 × 3                  & 305                  & \textbf{75,3}                  \\
                                                            \hline\hline
\multirow{22}{*}{\rotatebox[origin=c]{90}{\textbf{Image + video pre-tr. (I + V)}}} & \multirow{2}{*}{IN + K400}                                                     & UniFormer-B                 & \multirow{2}{*}{\cite{li2022uniformer}}                     & 16@4 × 224\textsuperscript{2}                  & 96,67 × 1 × 3                 & 50                   & 70,4                  \\
                                                                         &                                                                                & UniFormer-B                 &                     & 32@4 × 224\textsuperscript{2}                  & 259,00 × 1 × 3                & 50                   & 71,2                  \\
                                                            \cline{2-2}\cline{3-8}
                                                                         & \multirow{11}{*}{IN21 + K400}                                                  & Swin-B                      & \cite{liu2021swinvideo}                    & 32@2 × 224\textsuperscript{2}                  & 0,28 × 1 × 3                  & 88                   & 69,6                  \\
                                                                         \cline{3-8}
                                                                         &                                                                                & X-ViT                       & \cite{bulat2021space}                      & 16@1 × 224\textsuperscript{2}                  & 0,28 × 1 × 3                  & 92                   & 67,2                  \\
                                                                         \cline{3-8}
                                                                         &                                                                                & MViTv2-B                    & \multirow{2}{*}{\cite{li2022mvitv2}}                        & 32@3 × 224\textsuperscript{2}                  & 0,23 × 1 × 3                  & 51                   & 72,1                  \\
                                                                         &                                                                                & MViTv2-L↑                   &                        & 40@3 × 312\textsuperscript{2}                  & 2,83 × 1 × 3                  & 218                  & \textbf{73,3}                  \\
                                                                         \cline{3-8}
                                                                         &                                                                                & Mformer                     & \multirow{2}{*}{\cite{patrick2021keeping}}                   & 96@3 × 224\textsuperscript{2}                  & 0,96 × 1 × 3                  & NA                   & 67,1                  \\
                                                                         &                                                                                & Mformer↑                    &                  & 64@4 × 336\textsuperscript{2}                  & 1,19 × 1 × 3                  & NA                   & 68,1                  \\
                                                                         \cline{3-8}
                                                                         &                                                                                & MTV-B                       & \multirow{2}{*}{\cite{yan2022multiview}}                   & 32@2 × 224\textsuperscript{2}                  & 0,40 × 4 × 3                  & 310                  & 67,6                  \\
                                                                         &                                                                                & MTV-B↑                      &                    & 32@2 × 320\textsuperscript{2}                  & 0,96 × 4 × 3                  & 310                  & 68,5                  \\
                                    \cline{3-8}
                                                                         &                                                                                & RViT-XL                     & \cite{yang2022recurring}                   & 64@NA × 224\textsuperscript{2}                 & 35,70 × 1 × 3                 & 108                  & 67,9                  \\
                                                                    
                                    \cline{3-8}
                                                                         &                                                                                & MViTv2-S                    & \cite{li2022mvitv2}                        & 16@4 × 224\textsuperscript{2}                  & 0,06 × 1 × 3                  & 35                   & 68,2                  \\
                                    \cline{3-8}
                                                                         &                                                                                & ORViT MF-L                  & \cite{herzig2022object}                    & 32@4 × NA                   & 1,26 × 1 × 3                  & 148                  & 69,5                  \\
                                                            \cline{2-2}\cline{3-8}
                                                                         & \multirow{2}{*}{IN + K600}                                                     & UniFormer-B                 & \multirow{2}{*}{\cite{li2022uniformer}}                     & 16@4 × 224\textsuperscript{2}                  & 96,67 × 1 × 3                 & 50                   & 70,2                  \\
                                                                         &                                                                                & UniFormer-B                 &                     & 32@4 × 224\textsuperscript{2}                  & 259,00 × 1 × 3                & 50                   & 71,2                  \\
                                                            \cline{2-2}\cline{3-8}
                                                                         & IN21 + K600                                                                    & SCT-L                       & \cite{zha2021shifted}                      & 24@10 × 224\textsuperscript{2}                 & 0,34 × 4 × 3                  & 60                   & 68,1                  \\
                                                            \cline{2-2}\cline{3-8}
                                                                         & \multirow{2}{*}{IN + K400 (SSL)}                                               & \multirow{2}{*}{SVT (TS)}   & \multirow{2}{*}{\cite{ranasinghe2022self}} & 8@NA × 224\textsuperscript{2} +                     & \multirow{2}{*}{0,20 × 1 × 3} & \multirow{2}{*}{121} & \multirow{2}{*}{59,2} \\
                                                                         &                                                                                &                             &                                            &                                               64@NA × 96\textsuperscript{2} &                               &                      &                       \\
                                                            \cline{2-2}\cline{3-8}
                                                                         & IN21 + K400 (SSL) & MaskFeat-L &   \cite{wei2022masked}     & 40@3 × 224\textsuperscript{2} & 2,83 × 1 × 3 & 218 & \textbf{73,3} \\
                                                            \cline{2-2}\cline{3-8}
                                                                         &  \multirow{2}{*}{\begin{tabular}[c]{@{}c@{}}IN (SSL) +\\   K400 (SSL)\end{tabular}}                                                                              & BEVT                        & \multirow{2}{*}{\cite{wang2022bevt}}                        & 16@NA × 224\textsuperscript{2}                 & 0,32 × 1 × 3                  & 88                   & 70,6                  \\
                                                                         &                                                                                & BEVT (Dall-E tknzr.)        &                         & 16@NA × 224\textsuperscript{2}                 & 0,32 × 1 × 3                  & 88                   & 71,4 \\
\hline
\end{tabular}
\begin{tablenotes}
\item *: re-implementation.
\item ↑: increased spatial resolution.
\end{tablenotes}
\end{threeparttable}}
\caption{Accuracy (top-1) in Something-Something v2. See caption in~\cref{tab:perf_k400}.}
\label{tab:perf_ssv2}
\end{table}


\noindent\textbf{SSv2: video (supervised) pre-training}. It is quite common to reuse supervisedly trained checkpoints on Kinetics by transferring them to SSv2 for fine-tuning. These have often also been pre-trained on IN-1K/IN-21 before Kinetics. However, to better disentangle video and image contributions, we focus first on video-only pre-training models, and concretely on those relying on K600. Looking at ``MViT-B 32@3'' (67.8\% pre-trained on K600) and ``MViT-B 64@4'' (67.7\% pre-trained on K400), with a temporal receptive field of 96 to 128 frames respectively, we see there is no improvement in SSv2 by extending temporal context, but slightly better performance when keeping finer temporal resolution (stride 3 instead of 4). Even more interesting is that the deeper ``MViT-B-24 32@3'' (with 24 layers) outperforms +0.9\% upon the 12-layered 32@3 variant. This suggests more complex temporal dynamics might require not necessarily increasing temporal resolution, but higher abstract spatiotemporal semantics being modeled. That or advancements in architectural designs to better model those without going deeper, as done by ``MViTv2-B 32@3'' (70.5\%) also with 12 layers. Finally, if we have a look at models that leverage image-based pre-training before Kinetics, we find further improvement (e.g., ``MViTv2-B 32@3'' from 70.5\% to 72.1\%).  What seems to be not as useful, according to ``UniFormer'' variants, is to switch from K400 to K600.

\noindent\textbf{SSv2: video (self-supervised) pre-training}. The only model pre-training on SSv2 is VideoMAE (``VMAE''), which turns out to be the best performing one. In particular, ``VMAE (ViT-L) 32@2'' (75.3\%) slightly improves upon ``MaskFeat-L↑ 40@3'', those being self-supervisedly pre-trained on K400 (74.4\%) or K600 (75.0\%). It does so with almost half the temporal context, half the FLOPs, and -- importantly -- with fewer data. All in all, VideoMAE and MaskFeat seem to point out pixel- and feature-based MTM approaches compare favorably with ``SVT'' (instance-based invariance learning) or ``BEVT'' (quantization-based MTM) despite the latter also using image-based pre-training.

\subsection{Discussion on performance}
\label{sec:performance_discussion}

We have introduced the task of video classification and analyzed the performance of state-of-the-art models on Kinetics-400 and Something-Something v2. Our main finding was that the pre-training strategy was the biggest factor influencing the performance of VTs for video classification, thus the following discussion will address three questions related to this: (1) \textit{Can Video Transformers be trained from scratch?}, (2) \textit{Which is the best pre-training strategy?}, and (3) \textit{How can we effectively model stronger spatiotemporal dynamics?}. 

For the smallest models, \textit{training from scratch} seems to be doable. In particular, MViT~\cite{fan2021multiscale} and MViTv2~\cite{li2022mvitv2} are able to, respectively, compete with and slightly surpass 3D ConvNets trained from scratch. In fact, MViTv2 even outperforms those pre-trained on very large weakly-annotated video datasets (e.g., IG-65M). In particular, we attribute the success of those to the locality bias they infused (via the local pooling-based progressive aggregation discussed in \cref{sec:aggregation}), which allows these models to go deeper without exploding in computational complexity while still keeping their self-attention operation global. However, training from scratch seems to be the least desirable strategy to follow.

Among pre-training strategies, video-based ones, either supervised (e.g., on K400/K600 before fine-tuning SSv2) or self-supervised, are superior to image-based pre-training alone. Image-based supervised pre-trained models seem to be able to partially compensate the lack of temporal modeling with appearance diversity by leveraging huge -- often non-public -- image datasets (e.g., JFT~\cite{yan2022multiview, ryoo2021tokenlearner, arnab2021vivit} or  extensions of IN21~\cite{liu2022swinv2}). Alternatively, image-based self-supervised learning only competes with video pre-training when leveraging prohibitively large models~\cite{liu2022swinv2}. However, parting from learned very diverse and general appearance features will not harm the modeling of time in later stages, but serve as a good initialization for subsequent video-based pre-training (e.g., all those works that combine IN/IN21 and K400/K600 before fine-tuning on SSv2) or fine-tuning stages with temporal SSL auxiliary losses (e.g., ~\cite{truong2022direcformer,yun2022matter}). On the other hand, we can see how self-supervised video pre-training surpasses supervised regimes. In particular, MaskFeat~\cite{wei2022masked} (with MViTv2~\cite{li2022mvitv2} backbone) and VideoMAE~\cite{tong2022videomae} (with a plain ViT~\cite{arnab2021vivit}) outperform those pre-trained on video in a supervised way. 

For the successful modeling of spatiotemporal patterns, Masked Token Modeling stands out (see~\cref{sec:discussion_training}). Concretely, MaskFeat (feature-based MTM) obtains the best results on K400 and is second best on SSv2, which is dominated by VideoMAE (pixel-based MTM). Interestingly, these models do not require extra data or manual annotations to surpass all other models, being able to self-supervisedly pre-train on the evaluation dataset itself. Unfortunately, instance-based invariance learning (e.g.,~\cite{ranasinghe2022self,wang2022long}), being that popular for image representation learning, heavily underperforms compared to MTM for video classification.

Apart from the importance of pre-training, other findings in~\cref{sec:performance_comparsion} we might want to highlight are: first, that the modeling of the complex spatiotemporal dynamics seems to benefit more from deeper models and temporal granularity than extended temporal spans; second, that naive adoptions of image and NLP models (e.g., VTN~\cite{neimark2021video}, which leverages the image-based ViT~\cite{beltagy2020longformer} to model space and the language-based Longformer~\cite{beltagy2020longformer} to model time) might not work that well; and, third, that although joint self-supervised learning on image and video (i.e., BEVT~\cite{wang2022bevt}) is promising, it still has a long way to go.

\vspace{-0.25cm}
\section{Final Discussion}
\label{sec:discussion}
In this survey, we have comprehensively analyzed trends and advances in leveraging Transformers to model video.

\noindent\textbf{Complexity}. Given the inherent complexity of Transformers and the great dimensionality of videos, most changes focus on handling the computational burden. This is done transversally across the various stages of the VT pipeline. 
We find this is most generally addressed with frozen embedding networks, easing Transformer learning through the provided inductive biases and reducing input dimensionality. The Transformer in this context is used to enhance these representations through long-range interactions, which seems enough to boost performance in many areas of application. However, this trend alone may be limiting the potential of Transformers to learn non-local low-level motion cues. We are excited to see novel VT designs (e.g., MViT~\cite{fan2021multiscale}) which greatly reduce complexity thanks to the inductive biases embedded in the Transformer itself (sometimes becoming lighter than CNN counterparts, see~\cref{sec:performance_comparsion}). We also see great promise in MTM when separating the representation learning from the reconstruction which is done by an additional decoder discarded after pre-training~\cite{tong2022videomae}. This separation allows the (deeper) encoder to only leverage unmasked tokens, which greatly alleviates training complexity when using large masking ratios. Crucially, this sacrifices the possibility to leverage certain designs for the VT, as the input structure is lost (e.g., local or hierarchical approaches may not find enough tokens in a given neighborhood to learn valuable representations).

\noindent\textbf{Spatial redundancy and temporal fidelity}. Modeling temporal interactions requires special considerations not present when only modeling appearance (i.e., with image Transformers). On the one hand, the highly redundant appearance information in videos~\cite{zhang2012slow,tong2022videomae} makes it difficult to model information-rich representations that avoid repeatedly representing similar or same sub-representations. It has been proven that pure attentional models lose expressivity with depth, collapsing towards uniform attention in deeper layers \cite{dong2021attention,dosovitskiy2021an, jaegle2021perceiver}. It further seems that this smoothing of the attention matrix is accompanied by highly uniform token representations and even redundant weight matrices~\cite{chen2022principle}. Proper handling of video redundancy is crucial in VTs, where we hypothesize these observations may get exacerbated. On the other hand, few exceptions aside, many current designs and SSL approaches directly inherit from image approaches without careful consideration of the nuances that come with time, making them strongly biased to learn appearance features. As we have seen, allowing temporal features to form at both low- and high-level while accounting for the necessary temporal fidelity is also critical. In this sense, reducing redundancy for video should mostly target appearance features. 

\noindent\textbf{Key advancements on VTs}. Regarding \textit{architectural choices}, we find progressive hierarchical approaches to stand out. They carefully consider non-local temporal contexts before spatial aggregation. This effectively tackles the redundancy problem while avoiding early aggregation problems that hinder the learning of fine-grained motion features. However, to properly handle long-range interactions without losing temporal fidelity, memory-based approaches with adequate sampling or aggregation techniques may be crucial. Regarding \textit{self-supervised learning}, MTM forces to leverage global spatiotemporal semantic contexts through high masking ratios when solving local token-wise predictions. By doing so, it is driven to learn both motion and appearance cues necessary to solve the task. Nevertheless, we look forward to further developments in sampling techniques for instance-based contrastive approaches that skew from appearance biases toward motion-specific features.

\noindent\textbf{Inductive biases}. As we have seen, inductive biases are a pivotal aspect for all aspects of VTs. They alleviate the need for data by providing stronger cues for the Transformer to pick up faster. Frozen \textit{embedding networks} could be regarded as infusing task-specific biases, as the Transformer is bounded to learn on the provided representations, which in turn are dependant on the pre-training auxiliary task. Some examples include detected bounding boxes of objects~\cite{Girdhar_2021_ICCV, herzig2022object}, higher-level (action) features~\cite{zhu2020actbert}, or scene, motion, OCR, and facial features, among others~\cite{gabeur2020mmt}. We have also seen how most \textit{architectural designs} infuse some inductive biases to aid in training the Transformer. However, in this regard, VT literature so far is limited when considering infusing motion-specific biases that help the network to pick up relevant spatiotemporal cues. Just two works deviate from this trend. Motionformer~\cite{patrick2021keeping} proposes trajectory attention to reason about aggregated object or region representations through implicit motion paths in both time and space. Differently, OrViT \cite{herzig2022object} leverages separate motion and appearance streams. The former learns trajectories of individual objects or regions that later get added to patch-wise token representations of the whole video appearance, effectively infusing motion into it. Finally, besides locality biases or invariance to perturbations induced by different \textit{training losses}, we deem it interesting to highlight works infusing causality biases by training the network to sort shuffled video sequences~\cite{truong2022direcformer,yun2022matter}. Furthermore, the work in~\cite{guo2022cross} combines the benefits of both CNNs and Transformers for video learning through a siamese distillation setting, effectively inducing CNN locality biases into the Transformer. 

\subsection{Generalization}
It has been shown that vision Transformers are robust to various perturbations~\cite{Bhojanapalli_2021_ICCV, mahmood2021robustness}, suggesting they may be better able to form abstract semantic representations~\cite{zhang2021delving}, probably due to their ability to leverage non-local contexts~\cite{paul2022vision}. These findings point towards Transformers favoring out-of-distribution (OOD) generalization~\cite{HendrycksLWDKS20}. Few VTs have studied this on OOD data~\cite{zhu2020actbert,patrick2021supportset,Weng_2021_ICCV,Liu_2021_ICCV,shao2021temporal,liu2019learning} or evaluated the learned features in other settings~\cite{ging2020coot,sun2019videobert,Yu_2021_ICCV,Tan_2021_ICCV}, showing consistent results. Nevertheless, the issue of generalization of video may entail studying other aspects that are still under-researched. For instance, we hypothesize that generalizing to varied frame sampling rates may require further training or conditioning the network on said rates such that it may become robust. We observed, however, that some existing work may display capabilities to generalize to unseen sequence lengths.

\noindent\textbf{Unseen sequence length}. One issue to account for when processing sequences of unseen length is positional encodings. While we expect them to generalize to shorter sequences, they may have trouble when dealing with longer ones (which may be desirable to provide extended temporal fidelity during deployment), as no positional information is present to account for them. We find few VTs showing that PEs can easily be extended by fine-tuning the model on longer sequence lengths~\cite{akbari2021vatt,fan2021multiscale}. Recent VT works have also seen promising results when leveraging input conditioned RPEs~\cite{li2022uniformer} or by learning a small network that computes log-scale relative positional biases~\cite{liu2022swinv2}. These advances pose a great potential to easily generalize to unseen sequence lengths. Similarly, long-range modeling architectures could also handle sequences of any given length, as they process inputs sequentially within fixed windows, but they may require RPEs~\cite{wu2022memvit}. 

\noindent\textbf{Multi-modality}. Video is inherently multi-modal (i.e., contains visual and auditory information), which could be leveraged to learn more general representations. The lack of inductive biases makes Transformers very versatile tools to handle any modality. It has been found that high-level semantic features learned by language-based Transformers generalize to other modalities~\cite{sung2022vl,lu2021pretrained}. In the context of VTs we find VideoBERT~\cite{sun2019videobert}, where a pre-trained language BERT~\cite{devlin2019bert} model is used as initialization, showing promising results in this direction. Lately, there has been a great interest to use these architectures to solve multi-modal tasks~\cite{multimodalSurvey2}. We hypothesize that the lack of inductive biases may allow Transformers to learn shared multi-modal representation spaces that exhibit better generalization capabilities. When targeting video-only tasks (e.g., tracking, segmentation, classification) we see potential in multi-modal SSL to learn such spaces. We find a few VTs leveraging instance-based multi-modal learning approaches~\cite{ging2020coot, lee2021parameter, contrastive2019chen, li2020hero} to align representations from various modalities. For instance,~\cite{akbari2021vatt} successfully performs heavy downsampling of video by aligning it with audio and textual modalities; or the model in~\cite{Patrick_2021_ICCV} which learns to attend to the spatial sources of audio within the video by aligning audio with visual crops. Interestingly, this alignment is further enforced in some works by sharing weights between Transformer streams modeling different modalities~\cite{akbari2021vatt}, sometimes even showing improved results compared to not sharing~\cite{lee2021parameter}. As pointed out in~\cite{schiappa2022self}, this has proven to be very useful for video (at least in the context of classification) outside of Transformers, especially when pairing video with audio or text.    

\subsection{Future work}
VTs are still in their infancy and despite seeing clear trends, much more research is needed. First of all, we find a severe lack of explainability tools that properly assess the kind of spatiotemporal representations that different designs and self-supervised losses provide. Overlaying head-specific attention heat-maps of the first layer over a given input may provide some ad-hoc explanations on what the model deems relevant~\cite{serrano2019attention,jain2019attentionNot,wiegreffe2019attentionNotNot}. Even if some VTs have explored this direction (e.g.,~\cite{Pashevich_2021_ICCV,patrick2021supportset,jaegle2021perceiver,neimark2021video,Patrick_2021_ICCV,bozic2021transformerfusion,Girdhar_2021_ICCV, li2020hero}), this technique may prove overly cumbersome for video, as it requires inspecting such per-sample activations for multiple full video sequences. Possible future venues could analyze the learned patterns of attention preferred by different heads (as in~\cite{naseer2021intriguing}), which may clue on relevant design choices that favor such patterns. Besides, the aforementioned versatility of Transformers could be used to probe the model through textual descriptions (as done for images in~\cite{thrush2022winoground}). Furthermore, we see an interesting future direction in analyzing whether video-based features would also generalize to other modalities. For instance by following a similar approach as in~\cite{sung2022vl} and tuning a few adapter layers to map other modalities into the video representation space. Beyond current MTM approaches, other traditional losses could be adapted to the token granularity, such as 3D jigsaw puzzles~\cite{kim2019self}. Regarding instance-based methods, adapting recent developments to images such as Barlow Twins~\cite{zbontar2021barlow} or VicReg~\cite{bardes2022vicreg} which focus on preserving view-dependent information, may prove beneficial to video modeling. Nevertheless, further research is still needed to alleviate the computational burden of self-supervision in video. Finally, key advancements in architectural choices and training techniques for VTs are mostly limited to high-level tasks, hindering analysis of the contributions they provide for general video representation learning. Furthermore, VTs have barely tackled generative tasks such as frame prediction~\cite{Weissenborn2020Scaling,rakhimov2020latent} or inpainting~\cite{Liu_2021_ICCV,zeng2020learning}. We believe that token granularity and long-range modeling capabilities of Transformers could benefit these tasks. However, given the high complexity, they entail and the tendency of Transformers to disregard high-frequency details may pose severe challenges. We hope our contributions in this paper will entice further research in many different areas of application and boost our current understanding of Video Transformers.


\section*{Acknowledgments}
{\normalsize This work was funded by the Pioneer Centre for AI, DNRF grant number P1, by the Spanish project PID2019-105093GB-I00, by ICREA under the ICREA Academia program, and by Milestone Research Programme at Aalborg University.}

\ifCLASSOPTIONcaptionsoff
  \newpage
\fi

\vspace{-0.25cm}
\bibliographystyle{IEEEtran}
\bibliography{./biblio}


\newpage
\clearpage
\section{Supplementary} 

The supplementary material includes the following: the general table with a general overview of the most relevant Video Transformers surveyed in \cref{sec:general_table} and details about specific Transformer trends for different video tasks in a more application-oriented manner in \cref{sec:task_specific_designs}.

\subsection{General table}
\label{sec:general_table}

The general table overviews the most relevant Video Transformers surveyed. Note that due to its length, the table has been split into two subtables, \cref{tab:general_table_a} and \cref{tab:general_table_b}.

\begin{table*}[t!]
\setlength{\tabcolsep}{2pt}
\adjustbox{max width=\textwidth}{
\begin{threeparttable}
\begin{tabular}{|l|lr|l|cccc|llcc|c|}
\cline{2-13}
\multicolumn{1}{c|}{} & \multirow{2}{*}{\textbf{Name}} & \multirow{2}{*}{\textbf{Ref.}} & \multirow{2}{*}{\textbf{Yr.}} & \multicolumn{4}{c|}{\textbf{Architecture}} & \multicolumn{4}{c|}{\textbf{Input}} & \textbf{Train.} \\
\cline{5-13}
\multicolumn{1}{c|}{} & \hfill & \hfill & \hfill & \textbf{Arch.} & \textbf{Aggr.} & \textbf{Restr.} & \textbf{Long-t.} & \textbf{Backbone} & \textbf{Embd.} & \textbf{Tknz.} & \textbf{Pos.} & \textbf{SSL} \\
\hline
\multirow{35}{*}{\rotatebox[origin=c]{90}{\textbf{Classification}}} & TimeSformer       & \cite{bertasius2021spacetime}     & \textquotesingle 21 & E  & - & LAS   & -  & -                                               & Minimal Embedding                                                                                 & P     & LA        & - \\
& PE                & \cite{lee2021parameter}           & \textquotesingle 21 & E  & - & -     & -  & -                                               & SlowFast\cite{bb_slowfast}, RN-50\cite{bb_resnet}                                                 & C     & LA        & P \\
& CBT               & \cite{contrastive2019chen}        & \textquotesingle 19 & E  & - & -     & -  & -                                               & S3D\cite{bb_s3d}                                                                                  & C     & -         & P \\
& ViViT             & \cite{arnab2021vivit}             & \textquotesingle 21 & E  & H & A     & -  & ViT~\cite{dosovitskiy2021an}                    & Minimal Embedding                                                                                 & P     & LA        & - \\
& ELR               & \cite{purwanto2019extreme}        & \textquotesingle 19 & E  & - & -     & -  & -                                               & I3D\cite{bb_i3d}                                                                                  & P     & -         & - \\
& FAST              & \cite{Yu_2021_ICCV}               & \textquotesingle 21 & E  & - & -     & -  & -                                               & Minimal Embedding                                                                                 & P     & LA        & - \\
& VATNet            & \cite{girdhar2019video}           & \textquotesingle 19 & E  & Q & -     & -  & -                                               & I3D\cite{bb_i3d}, Faster R-CNN (RP only)\cite{bb_faster}                                          & P + I & FA        & - \\
& VATT              & \cite{akbari2021vatt}             & \textquotesingle 21 & E  & - & S     & -  & -                                               & Minimal Embedding                                                                                 & P     & LA        & P \\
& MViT              & \cite{fan2021multiscale}          & \textquotesingle 21 & E  & H & -     & -  & -                                               & Minimal Embedding                                                                                 & P     & LA        & - \\
& SCT               & \cite{zha2021shifted}             & \textquotesingle 21 & E  & H & L     & -  & -                                               & Minimal Embedding                                                                                 & P     & LA        & - \\
& CATE              & \cite{Sun_2021_ICCV}              & \textquotesingle 21 & E  & - & -     & -  & -                                               & SlowFast\cite{bb_slowfast} (Slow br.)                                                             & C     & -         & P \\
& LapFormer         & \cite{kondo2020lapformer}         & \textquotesingle 20 & E  & - & -     & -  & -                                               & RN-50\cite{bb_resnet}                                                                             & P     & FA        & - \\
& TRX               & \cite{perrett2021temporal}        & \textquotesingle 21 & E  & - & -     & -  & -                                               & RN-50\cite{bb_resnet}                                                                             & F     & FA        & - \\
& LTT               & \cite{kalfaoglu2020late}          & \textquotesingle 20 & E  & - & -     & -  & -                                               & R(2+1)D\cite{bb_3dresnet}                                                                         & F     & LA        & - \\
& Actor-T           & \cite{gavrilyuk2020actor}         & \textquotesingle 20 & E  & - & -     & -  & -                                               & I3D\cite{bb_i3d}, HRNet\cite{bb_hrnet}                                                            & I     & FA        & - \\
& STiCA             & \cite{Patrick_2021_ICCV}          & \textquotesingle 21 & E  & - & -     & -  & -                                               & R(2+1)D-18\cite{bb_3dresnet}, RN-9\cite{bb_resnet}                                                & F     & LA        & A \\
& GroupFormer       & \cite{li2021groupformer}          & \textquotesingle 21 & ED & Q & L     & -  & -                                               & I3D\cite{bb_i3d}                                                                                  & I + F & LA        & - \\
& Video Swin        & \cite{liu2021swinvideo}           & \textquotesingle 21 & E  & H & L     & -  & -                                               & Minimal Embedding                                                                                 & P     & LR        & - \\
& VTN               & \cite{neimark2021video}           & \textquotesingle 21 & E  & H & L     & -  & ViT~\cite{dosovitskiy2021an}                    & Minimal Embedding                                                                                 & P     & LA        & - \\
& Video-Swin-V2     & \cite{liu2022swinv2}              & \textquotesingle 22 & E  & H & L     & -  & -                                               & RN-50\cite{bb_resnet}                                                                             & P     & LR        & P \\
& MTV               & \cite{yan2022multiview}           & \textquotesingle 22 & E  & H & -     & -  & ViT~\cite{dosovitskiy2021an}                    & Minimal Embedding                                                                                 & P     & LA        & - \\
& Motionformer      & \cite{patrick2021keeping}         & \textquotesingle 21 & E  & - & -     & -  & -                                               & Minimal Embedding                                                                                 & P     & LA        & - \\
& X-ViT             & \cite{bulat2021space}             & \textquotesingle 21 & E  & - & L     & -  & ViT~\cite{dosovitskiy2021an}                    & Minimal Embedding                                                                                 & P     & LA        & - \\
& ObjTr             & \cite{wu2021towards}              & \textquotesingle 21 & E  & - & -     & -  & -                                               & Faster R-CNN~\cite{bb_faster}, RN-101~\cite{bb_resnet}                                            & I     & FA + LA   & - \\
& MViTv2            & \cite{li2022mvitv2}               & \textquotesingle 22 & E  & H & -     & -  & -                                               & Minimal Embedding                                                                                 & P     & LR        & - \\
& MaskFeat          & \cite{wei2022masked}              & \textquotesingle 22 & E  & H & -     & -  & MViT~\cite{li2022mvitv2}                        & Minimal Embedding                                                                                 & P     & LR        & P \\
& LSTCL             & \cite{wang2022long}               & \textquotesingle 22 & E  & - & -     & -  & Swin~\cite{liu2021swinvideo}                    & Minimal Embedding                                                                                 & P     & LA        & P \\
& RViT              & \cite{yang2022recurring}          & \textquotesingle 22 & E  & - & -     & R  & ViT~\cite{dosovitskiy2021an}                    & Minimal Embedding                                                                                 & P     & LA        & - \\
& Direcformer       & \cite{truong2022direcformer}      & \textquotesingle 22 & E  & - & A     & -  & TimeSformer~\cite{bertasius2021spacetime}       & Minimal Embedding                                                                                 & P     & LA        & - \\
& VideoMAE          & \cite{tong2022videomae}           & \textquotesingle 22 & E  & - & S*    & -  & ViT~\cite{dosovitskiy2021an}                    & Minimal Embedding                                                                                 & P     & LA        & P \\
& BEVT              & \cite{wang2022bevt}               & \textquotesingle 22 & E  & H & L     & -  & Swin~\cite{liu2021swinvideo}                    & Minimal Embedding                                                                                 & P     & LA        & P \\
& TIME       & \cite{yun2022matter}              & \textquotesingle 22 & E  & - & -     & -  & Motionformer~\cite{patrick2021keeping}          & Minimal Embedding                                                                                 & P     & LA        & A \\
& TokenLearner      & \cite{ryoo2021tokenlearner}       & \textquotesingle 21 & E  & H & -     & -  & ViT~\cite{dosovitskiy2021an}                    & Minimal Embedding                                                                                 & P     & LA        & - \\
& SVT               & \cite{ranasinghe2022self}         & \textquotesingle 22 & E  & - & A     & -  & TimeSformer~\cite{bertasius2021spacetime}       & Minimal Embedding                                                                                 & P     & LA        & P \\
& UniFormer         & \cite{li2022uniformer}            & \textquotesingle 22 & E  & H & L     & -  & -                                               & Minimal Embedding                                                                                 & P     & LA + LR * & - \\
\hline
\multirow{9}{*}{\rotatebox{90}{\textbf{Captioning}}} & ActBERT           & \cite{zhu2020actbert}             & \textquotesingle 20 & E  & - & -     & -  & -                                               & R(2+1)D\cite{bb_3dresnet}, Faster R-CNN \cite{bb_faster}                                          & I + C & LA        & P \\
& HERO              & \cite{li2020hero}                 & \textquotesingle 20 & E  & H & -     & -  & -                                               & RN-101\cite{bb_resnet}, SlowFast\cite{bb_slowfast}                                                & F     & FA        & P \\
& MART              & \cite{lei2020mart}                & \textquotesingle 20 & ED & - & -     & R  & -                                               & RN-200\cite{bb_resnet}, BN Inception\cite{bb_bninception}                                         & F     & FR        & - \\
& VideoBERT         & \cite{sun2019videobert}           & \textquotesingle 19 & E  & - & -     & -  & -                                               & S3D\cite{bb_s3d}                                                                                  & C     & LA        & P \\
& E2E-DC            & \cite{zhou2018end}                & \textquotesingle 19 & ED & - & -     & -  & -                                               & RN-200\cite{bb_resnet}, BN Inception\cite{bb_bninception}                                         & F     & FA        & - \\
& BMT               & \cite{BMT_Iashin_2020}            & \textquotesingle 20 & ED & - & -     & -  & -                                               & I3D\cite{bb_i3d}                                                                                  & F     & FA        & - \\
& AMT               & \cite{yu2021accelerated}          & \textquotesingle 21 & ED & - & -     & -  & -                                               & RN-200\cite{bb_resnet}, BN-Inception\cite{bb_bninception}                                         & F     & FA        & - \\
& MDVC              & \cite{iashin2020multi}            & \textquotesingle 20 & ED & - & -     & -  & -                                               & I3D\cite{bb_i3d}                                                                                  & F     & FA        & - \\
& RLM               & \cite{li2020bridging}             & \textquotesingle 20 & D  & - & -     & -  & -                                               & I3D\cite{bb_i3d}                                                                                  & C     & FA        & - \\
\hline
\multirow{9}{*}{\rotatebox[origin=c]{90}{\textbf{Retrieval}}} & HiT               & \cite{Liu_2021_Hit}               & \textquotesingle 21 & E  & - & -     & -  & -                                               & S3D\cite{bb_s3d}, SENet-154\cite{bb_senet}                                                        & F + C & LA        & T \\
& COOT              & \cite{ging2020coot}               & \textquotesingle 20 & E  & H & -     & -  & -                                               & RN-152\cite{bb_resnet}; ResNext-101\cite{bb_resnext};  I3D\cite{bb_i3d}                           & F     & -         & T \\
& MMT               & \cite{gabeur2020mmt}              & \textquotesingle 20 & E  & - & -     & -  & -                                               & S3D\cite{bb_s3d}, DenseNet-101\cite{bb_densenet}, RN-50\cite{bb_resnet}, SENet-154\cite{bb_senet} & P + F & FA        & T \\
& Support-set       & \cite{patrick2021supportset}      & \textquotesingle 21 & E  & - & -     & -  & -                                               & RN-152\cite{bb_resnet}, R(2+1)D-34                                                                & F     & -         & T \\
& TCA               & \cite{shao2021temporal}           & \textquotesingle 21 & E  & - & -     & -  & -                                               & iMAC\cite{gordo2017end}, L-3-iRMAC\cite{kordopatis2019visil}                                      & F     & -         & T \\
& MDMMT             & \cite{dzabraev2021mdmmt}          & \textquotesingle 21 & E  & - & -     & -  & -                                               & CLIP\cite{bb_clip}                                                                                & F     & LA        & T \\
& Fast and Slow     & \cite{Miech_2021_CVPR}            & \textquotesingle 21 & D  & - & -     & -  & -                                               & TSM RN-50\cite{bb_tsmRN}                                                                          & P     & -         & T \\
& ClipBERT          & \cite{lei2021less}                & \textquotesingle 21 & E  & - & S*    & -  & -                                               & RN-50\cite{bb_resnet}                                                                             & P     & LA        & - \\
& CACL              & \cite{guo2022cross}               & \textquotesingle 22 & E  & - & -     & -  & -                                               & RN-50\cite{bb_resnet}                                                                             & F     & LA        & P \\
\hline
\multirow{7}{*}{\rotatebox[origin=c]{90}{\textbf{Tracking}}} & Hopper            & \cite{zhou2021hopper}             & \textquotesingle 21 & ED & - & -     & -  & -                                               & ResNeXt-101\cite{bb_resnext}, DETR\cite{carion2020end}                                            & I + F & LA        & - \\
& DTT               & \cite{yu2021high}                 & \textquotesingle 21 & ED & - & -     & -  & -                                               & RN-50\cite{bb_resnet}                                                                             & P     & LA        & - \\
& TrDIMP            & \cite{Wang_2021_Transformer}      & \textquotesingle 21 & ED & - & -     & -  & -                                               & RN-50\cite{bb_resnet}                                                                             & P     & -         & - \\
& TransT             & \cite{chen2021transformer}        & \textquotesingle 21 & E  & - & -     & -  & -                                               & RN-50\cite{bb_resnet}                                                                             & P     & FA        & - \\
& STARK             & \cite{yan2021learning}            & \textquotesingle 21 & ED & - & -     & -  & -                                               & RN-50\cite{bb_resnet}                                                                             & P     & FA        & - \\
& Trackformer       & \cite{meinhardt2022trackformer}   & \textquotesingle 22 & ED & Q & -     & MR & -                                               & RN-50\cite{bb_resnet}                                                                             & P     & FA        & - \\
& VDRFormer         & \cite{zheng2022vrdformer}         & \textquotesingle 22 & ED & Q & -     & MR & -                                               & RN-101\cite{bb_resnet}                                                                            & P     & FA        & - \\
\hline
\multirow{6}{*}{\rotatebox[origin=c]{90}{\textbf{Low-level}}} & ET-Net            & \cite{Weng_2021_ICCV}             & \textquotesingle 21 & ED & - & -     & -  & -                                               & ConvLSTM\cite{shi2015convolutional}                                                               & P     & FA        & T \\
& STTN              & \cite{zeng2020learning}           & \textquotesingle 20 & ED & - & -     & -  & -                                               & 2D CNN (custom)                                                                                   & P     & -         & T \\
& FuseFormer        & \cite{Liu_2021_ICCV}              & \textquotesingle 21 & ED & - & -     & -  & -                                               & I3D\cite{bb_i3d}                                                                                  & P     & -         & T \\
& SAVM              & \cite{Weissenborn2020Scaling}     & \textquotesingle 20 & ED & - & L     & -  & -                                               & Minimal Embeddings                                                                                & P     & LR        & T \\
& VLT               & \cite{rakhimov2020latent}         & \textquotesingle 20 & ED & - & -     & -  & -                                               & VQ-VAE\cite{oord2017neural}                                                                       & P     & FR        & T \\
& TransformerFusion & \cite{bozic2021transformerfusion} & \textquotesingle 21 & E  & - & S*    & M  & -                                               & RN-18\cite{bb_resnet}                                                                             & F     & LA        & - \\
\hline
\multirow{6}{*}{\rotatebox[origin=c]{90}{\textbf{Segmentation}}} & VisTR             & \cite{wang2021end}                & \textquotesingle 21 & ED & - & -     & -  & -                                               & RN-50\cite{bb_resnet}                                                                             & P     & FA        & - \\
& MFN               & \cite{wang2021spatiotemporal}     & \textquotesingle 21 & E  & - & -     & -  & -                                               & 3D CNN (custom)                                                                                   & P     & FA        & - \\
& CMSANet           & \cite{ye2021referring}            & \textquotesingle 21 & E  & - & -     & -  & -                                               & DeepLab-101\cite{bb_deeplab}                                                                      & P     & FA        & - \\
& IFC               & \cite{hwang2021video}             & \textquotesingle 22 & ED & Q & -     & -  & -                                               & RN-101\cite{bb_resnet}                                                                            & P     & FA        & - \\
& TeViT             & \cite{yang2022temporally}         & \textquotesingle 22 & ED & Q & -     & -  & MsgShifT~\cite{yang2022temporally, wang2022pvt} & Minimal Embedding                                                                                 & P     & FA        & - \\
& AOT               & \cite{yang2021associating}        & \textquotesingle 21 & E  & - & L     & MR & Swin~\cite{liu2021swinvideo}                    & MobileNet-V2\cite{bb_mobilenetv2}                                                                 & P     & FA + RL   & - \\
\hline
\end{tabular}
\begin{tablenotes}
 \item *: Non-attentional sparsity (e.g., input level).
\end{tablenotes}
\end{threeparttable}}
\caption{General overview of relevant Video Transformers surveyed. In \textit{Architecture}, ``Arch.'': architecture, that is Encoder (E), Decoder (D), or Encoder-Decoder (ED); ``Aggr.'', aggregation strategy, either Hierarchical (H) or Query-driven compression (Q); ``Restriction'', can be Local (L), Axial (A), Sparse (S), or a mix. ``Long-t.'': long-term temporal modeling, Memory (M), Recurrence (R), or a both. In \textit{Input}, ``Backbone'' refers to Transformer backbone;  ``Embd.'', the Embedding Network; ``Tknz', the tokenization strategy, patch- (P), instance- (I), frame- (F), or clip-wise (C); and ``Pos.'', the positional embedding, can be Fixed Absolute (FA), Fixed Relative (FR), Learned Absolute (LA), Learned Relative (LR), or a combination. (Continuation in \cref{tab:general_table_b})}
\label{tab:general_table_a}
\end{table*}

\begin{table*}[t!]
\setlength{\tabcolsep}{2pt}
\adjustbox{max width=\textwidth}{
\begin{threeparttable}
\begin{tabular}{|l|lr|l|cccc|llcc|c|}
\cline{2-13}
\multicolumn{1}{c|}{} & \multirow{2}{*}{\textbf{Name}} & \multirow{2}{*}{\textbf{Ref.}} & \multirow{2}{*}{\textbf{Yr.}} & \multicolumn{4}{c|}{\textbf{Architecture}} & \multicolumn{4}{c|}{\textbf{Input}} & \textbf{Train.} \\
\cline{5-13}
\multicolumn{1}{c|}{} & \hfill & \hfill & \hfill & \textbf{Arch.} & \textbf{Aggr.} & \textbf{Restr.} & \textbf{Long-t.} & \textbf{Backbone} & \textbf{Embd.} & \textbf{Tknz.} & \textbf{Pos.} & \textbf{SSL} \\
\hline
\multirow{4}{*}{\rotatebox[origin=c]{90}{\textbf{O.D.}}} & PCSA              & \cite{gu2020pyramid}              & \textquotesingle 20 & E  & - & L     & -  & -                                               & MobileNet-V3\cite{bb_mobilenetv3}                                                                 & P     & -         & - \\
& TCTR              & \cite{yuan2021temporal}           & \textquotesingle 21 & ED & - & -     & -  & -                                               & RN-50\cite{bb_resnet}                                                                             & P     & FA        & - \\
& PMPNet            & \cite{yin2020lidar}               & \textquotesingle 20 & ED & - & -     & -  & -                                               & GraphCNN (custom)                                                                                 & P     & -         & - \\
& ORViT             & \cite{herzig2022object}           & \textquotesingle 22 & ED & - & -     & -  & -                                               & Faster R-CNN\cite{bb_faster}, RN-50 \cite{bb_resnet}                                              & P + I & LR        & - \\
\hline
\multirow{4}{*}{\rotatebox[origin=c]{90}{\textbf{Summ.}}} & H-MAN             & \cite{liu2019learning}            & \textquotesingle 19 & E  & - & -     & -  & -                                               & VAE-GAN\cite{mahasseni2017unsupervised}                                                           & F     & -         & - \\
& VasNet            & \cite{fajtl2018summarizing}       & \textquotesingle 19 & E  & - & -     & -  & -                                               & GoogLeNet\cite{bb_googlenet}                                                                      & F     & FA        & - \\
& BiDAVS            & \cite{lin2020bi}                  & \textquotesingle 20 & E  & - & -     & -  & -                                               & GoogLeNet\cite{bb_googlenet}                                                                      & F     & LR        & - \\
& VMTN              & \cite{seong2019video}             & \textquotesingle 19 & E  & Q & -     & -  & -                                               & ResNet-18\cite{bb_resnet}, SENet-101\cite{bb_senet}                                               & P     & FA        & - \\
\hline
\multirow{6}{*}{\rotatebox[origin=c]{90}{\textbf{Localiz.}}} & HISAN             & \cite{pramono2019hierarchical}    & \textquotesingle 19 & E  & - & -     & -  & -                                               & Faster R-CNN\cite{bb_faster}                                                                      & I + F & -         & - \\
& STVGBert          & \cite{Su_2021_ICCV}               & \textquotesingle 21 & E  & Q & -     & -  & -                                               & RN-101\cite{bb_resnet}                                                                            & P     & -         & - \\
& MeMViT            & \cite{wu2022memvit}               & \textquotesingle 22 & E  & H & -     & M  & MViTv2~\cite{li2022mvitv2}                      & Minimal Embedding                                                                                 & P     & LR        & - \\
& MSAT              & \cite{Zhang_2021_CVPR}            & \textquotesingle 21 & E  & - & -     & -  & -                                               & C3D\cite{bb_c3d}                                                                                  & C     & FA        & - \\
& RTD-Net           & \cite{Tan_2021_ICCV}              & \textquotesingle 21 & D  & - & -     & -  & -                                               & I3D\cite{bb_i3d}                                                                                  & F     & LR        & - \\                    
& LSTR              & \cite{xu2021long}                 & \textquotesingle 21 & ED & Q & -     & M  & -                                               & RN-50\cite{bb_resnet}                                                                             & F     & FA        & - \\
\hline
\multirow{4}{*}{\rotatebox[origin=c]{90}{\textbf{Others}}} & SiaSamRea         & \cite{yu2021learning}             & \textquotesingle 21 & E  & - & S*    & -  & ClipBERT~\cite{lei2021less}                     & RN-50\cite{bb_resnet}                                                                             & P     & LA        & A \\
& Perceiver         & \cite{jaegle2021perceiver} P      & \textquotesingle 21 & E  & Q & -     & -  & -                                               & Minimal Embedding                                                                                 & P     & LA        & - \\
& AVT               & \cite{Girdhar_2021_ICCV}          & \textquotesingle 21 & E  & H & -     & -  & ViT~\cite{dosovitskiy2021an}                    & Minimal Embedding                                                                                 & P     & LA        & A \\
& OadTR             & \cite{Wang_2021_ICCV}             & \textquotesingle 21 & ED & - & -     & -  & -                                               & RN-200\cite{bb_resnet}, BN-Inception\cite{bb_bninception}                                         & F     & LA        & - \\
& STTran            & \cite{cong2021spatial}            & \textquotesingle 21 & ED & - & L     & -  & -                                               & RN-101 F R-CNN\cite{ bb_faster}                                                                   & I + F & LA        & - \\
& E.T.              & \cite{Pashevich_2021_ICCV}        & \textquotesingle 21 & E  & - & -     & -  & -                                               & Faster R-CNN\cite{bb_faster}, Mask R-CNN\cite{bb_mask}                                            & F     & FA        & - \\
& SMT               & \cite{Fang_2019_CVPR}             & \textquotesingle 19 & ED & Q & S* & M  & -                                                  & RN-18\cite{bb_resnet}                                                                             & F     & FA        & - \\
& JSLT              & \cite{camgoz2020sign}             & \textquotesingle 20 & ED & - & -     & -  & -                                               & InceptionV4\cite{bb_inceptionv4}                                                                  & F     & FA        & - \\
& MSLT              & \cite{camgoz2020multi}            & \textquotesingle 20 & ED & - & -     & -  & -                                               & InceptionV4\cite{bb_inceptionv4}                                                                  & F     & FA        & - \\
& SBL               & \cite{luo_sychro_2020}            & \textquotesingle 20 & ED & - & -     & -  & -                                               & RN-18\cite{bb_resnet}                                                                             & F     & -         & - \\
& MDAM              & \cite{kim2018multimodal}          & \textquotesingle 19 & E  & Q & -     & -  & -                                               & RN-152\cite{bb_resnet}                                                                            & F     & FA        & - \\
& PSAC              & \cite{li2021vidtr}                & \textquotesingle 21 & E  & - & -     & -  & -                                               & Minimal Embedding                                                                                 & P     & FA        & - \\
& BTH               & \cite{Li_2021_CVPR}               & \textquotesingle 21 & E  & - & -     & -  & -                                               & VGG-16\cite{bb_vgg}                                                                               & F     & FA        & P \\
& BERT4SessRec      & \cite{chen2019bert4sessrec}       & \textquotesingle 20 & E  & - & -     & -  & -                                               & GoogLeNet\cite{bb_googlenet}                                                                      & C     & FA        & P \\
& Dyadformer        & \cite{Curto_2021_ICCV}            & \textquotesingle 21 & E  & - & -     & -  & -                                               & R(2+1)D-152\cite{bb_3dresnet}                                                                     & C     & FA        & - \\
& MM-Transformer    & \cite{roy2021action}              & \textquotesingle 22 & ED & - & L     & -  & -                                               & Mask R-CNN~\cite{bb_mask}                                                                         & I     & FA        & - \\
\hline
\end{tabular}
\begin{tablenotes}
 \item *: Non-attentional sparsity (e.g., input level)
\end{tablenotes}
\end{threeparttable}}
\caption{(Continuation of \cref{tab:general_table_a})}
\label{tab:general_table_b}
\end{table*}


\subsection{Task-specific designs}
\label{sec:task_specific_designs}

In this section, four major subsections review specific designs of the following tasks: Action classification in \cref{sec:classification_task}, Video translation (e.g., captioning) in \cref{sec:video_translation_task}, Retrieval in \cref{sec:retrieval_task}, and Object-centric tasks (e.g., detection and tracking) in \cref{sec:object_centric_tasks}. This is followed by short summary subsections regarding the remaining tasks: Low-level in \cref{sec:low_level_tasks}, Segmentation in \cref{sec:segmentation_task}, Summarization in \cref{sec:summarization_task}, and Others in \cref{sec:other_tasks}. 

\subsubsection{Classification}
\label{sec:classification_task}

Regarding video classification, few works rely on pure Transformers \cite{arnab2021vivit, liu2021swinvideo, bertasius2021spacetime, zha2021shifted,  akbari2021vatt} that for the most part focus on efficiency: both \cite{arnab2021vivit} and \cite{bertasius2021spacetime} test various space-time decompositions, whereas~\cite{arnab2021vivit} also tests tokenization strategies (2D vs 3D patches). They found that a pre-trained ViT~\cite{dosovitskiy2021an} encoding 2D patches with a temporal encoder on top performed the best. The works of \cite{liu2021swinvideo} and \cite{zha2021shifted} propose different types of restricted attention: the former restricts locally in shifting windows and the latter by only attending to previous frame's patches after having exchanged information with another efficient attention mechanism~\cite{kitaev2019reformer}. In \cite{fan2021multiscale} they opt for 3D patches whose receptive field is enlarged across stages by subsequently merging token embeddings. Others pursue building very deep Transformers by maintaining a very compact latent representation~\cite{jaegle2021perceiver}. These larger Transformers for classification require large labeled datasets for fully-supervised training~\cite{fan2021multiscale,liu2021swinvideo} or heavily rely on self-supervised pre-training~\cite{sun2019videobert,lee2021parameter}. For multi-modal datasets, encoder fusion~\cite{sun2019videobert} or hierarchical encoder fusion is utilized~\cite{contrastive2019chen}.

Several other works rely on larger (usually CNN-based) backbones~\cite{sun2019videobert, lee2021parameter, contrastive2019chen, perrett2021temporal, kalfaoglu2020late, gavrilyuk2020actor, Patrick_2021_ICCV, Sun_2021_ICCV}, facilitating the training on smaller datasets. When equipped with these backbones, shallow encoders can serve as mere pooling operators~\cite{kalfaoglu2020late, gavrilyuk2020actor, Patrick_2021_ICCV, purwanto2019extreme}. For detection backbones, Transformers are also a natural way to fuse information among detections~\cite{li2021groupformer} or to allow them to attend over a larger visual context~\cite{girdhar2019video}. Although mostly used in pure Transformers, efficient designs have been explored for these kinds of works as well, e.g. weight sharing~\cite{lee2021parameter}.

\subsubsection{Video translation}
\label{sec:video_translation_task}

The translation task intends to map the raw input video to an output signal of a potentially different nature and with an arbitrary (a priori unknown) length. Although the output could be another video, it is often a signal in another modality (e.g., language) or simply a sequence of discrete symbols. The most popular instantiation of translation is \textit{video captioning}~\cite{lei2020mart, li2020hero, BMT_Iashin_2020, Miech_2021_CVPR} that consists in producing natural language descriptions of what is globally going on in the video. When producing separate captions for different video subparts independently, this is referred to as \textit{dense video captioning}~\cite{zhou2018end, yu2021accelerated}. A more specialized type of video captioning is \textit{sign-language translation}~\cite{camgoz2020sign, camgoz2020multi}. Additional other forms of translation are: \textit{video reasoning}~\cite{zhou2021hopper}, which extends the task of captioning by allowing a natural language prompt along with the video; \textit{video-language dialogue} systems, which add to reasoning the requirement of back and forth communication with an external agent while reasoning about the visuals~\cite{li2020bridging}; \textit{temporal (or spatiotemporal) action localization}~\cite{Tan_2021_ICCV} to produce a list of, respectively, temporal begin and end times or a ``tube'' of bounding boxes containing the human actions in the video; or \textit{robot video-based navigation}~\cite{Fang_2019_CVPR}, in which the video -- and perhaps other sensory inputs -- are translated to the next action (a sequence of next actions) to take.

VTs tackling translation typically leverage encoder-decoder architectures, in which video is passed through the encoder and served as context to the decoder -- similarly to~\cite{vaswani2017attention}, only that the encoder is a video encoder instead of a language one. Task-specific modifications of this design are found for dense video captioning~\cite{BMT_Iashin_2020, zhou2018end, yu2021accelerated}, where a temporal proposal generator is attached after the encoder to tell the decoder where/when in the sequence it has to focus. \cite{BMT_Iashin_2020} is a two-stage method where the proposals are generated in the first stage. In the second stage, proposals are used to cut temporal clips from the video that need to be re-encoded (to avoid information from the different proposals being mixed up) and fed to the decoder to produce the per-clip captions. Slightly different is \cite{zhou2018end}, which instead of clipping the videos, converts the proposals to differentiable masks with a masking function whose parameters are trained also getting a back-propagation signal from the decoder. Still, the different masks have to be applied to the video and yet again re-forwarded through the encoder. \cite{yu2021accelerated} eliminates the re-forwarding by making the most of local self-attention, which limits the leak of information across the encoded proposals. \cite{Tan_2021_ICCV} tackled temporal action detection by relying only on a Transformer Decoder. Inspired by ~\cite{carion2020end}, proposals are not generated by the encoder or an external module after it but are sourced from a set of learnable token embeddings input to the decoder. The decoder augments these tokens and, later, two heads are in charge of classifying those into actions and regressing their temporal position and length -- similarly to a YOLO-like network~\cite{redmon2018yolov3}. 

In most of those works, the decoder module maintained its canonical form, although there are works that propose small variations. One has to do with the first and foremost SA sublayer. \cite{li2020bridging} removes the decoder's SA layer before the CA, whereas \cite{yu2021accelerated} substitutes it by a moving average -- both to make the models computationally lighter. Another one is to modify the input received by the CA sublayer. \cite{zhou2018end, BMT_Iashin_2020} receive the outputs of the encoder layers but at their respective depth, instead of only the one from the encoder's layer. The disadvantage of this is assuming the encoder and the decoder require the same number of layers. In \cite{camgoz2020multi}, the decoder also receives multiple inputs but from separate encoders. The decoder deals with those in different parallel CA sublayers and averages their outputs. There are also designs that go without a Transformer encoder, replacing it with an external non-Transformer module~\cite{Tan_2021_ICCV} or relying entirely on the decoder~\cite{li2020bridging, lei2020mart}. \cite{li2020bridging} follow the prompt-based input of GPT-2~\cite{radford2019language} and feds $n$ video features as the first tokens in the decoding sequence and decodes the caption starting at the $n+1$-th input embedding. In particular, \cite{lei2020mart} prompts not only the visual features but the current language sentence features to generate the next sentence in a paragraph. All in all, prompting is the generalization of the original shifting operation in~\cite{vaswani2017attention}, where the decoding starts at a shifted position to account for the start token.

\subsubsection{Video retrieval}
\label{sec:retrieval_task}
The task of retrieval consists in recovering a piece of information associated to a particular query. Those associations can be video-video pairs~\cite{shao2021temporal} or pairs composed of different modalities (video with, most often, language~\cite{ging2020coot, Liu_2021_Hit, Miech_2021_CVPR, patrick2021supportset} or language plus audio~\cite{gabeur2020mmt, dzabraev2021mdmmt}). Retrieval relies on a distance metric among the representations of the queries and the retrieval candidates. The representations are learned during training using the pairs to minimize the distances between the representations of the corresponding pairs while repelling from the query the non-corresponding candidates' representations in a joint space. This can be done through classification, by extending BERT's \textit{Next Sentence Prediction} to a cross-modal matching task, forcing the network to find co-occurrent information in both modalities~\cite{lee2021parameter,sun2019videobert,zhu2020actbert}. Alternatively, this can be naturally extended into a contrastive setting. In retrieval, it is common to use two anchors (which form the positive pair) and two negative sets, one from each modality. In VT literature we find these losses instantiated through a combined hinge loss~\cite{li2020hero,ging2020coot} or \textit{Bi-directional Max-Margin}~\cite{gabeur2020mmt,dzabraev2021mdmmt,patrick2021supportset}, which enforce similarity for true pairs to be higher than that of negative pairs, by at least a given margin. Alternatively, InfoNCE is also used~\cite{Liu_2021_Hit,Patrick_2021_ICCV}, normalizing the similarity score of positive pairs by that of a set of negative pairs, effectively forcing the network to learn similar representations for correctly paired samples and vice-versa for negative ones. While the most common approach is to align final output representations, some works leverage hierarchical contrastive losses, which also align intermediate feature representations~\cite{ging2020coot,Liu_2021_Hit}. During inference, the aligned representations are fixed, so the task simply becomes a search (e.g., K-Nearest Neighbors) to find the top-k examples most similar to a given query within the database of candidates' pre-computed representations.

One interesting variation of this pipeline is \cite{Miech_2021_CVPR}, in which the alignment is performed on the outputs of a siamese two-stream video-and-language CNN for faster retrieval instead. Then, a decoder-only Transformer fed with the text as input and CNN-based video features as context re-ranks the previously top-k retrieved elements using the decoding likelihood score. In a similar spirit, \cite{patrick2021supportset} also leverages the likelihood of a language-based decoder-only Transformer during training as a loss that measures how well the query language caption can be reconstructed from the weighted combination of the features from all the non-corresponding videos in the batch. Those weights are based on the similarity of the query caption with the captions of those other videos. \cite{gabeur2020mmt} aligns at the same time video, audio, and recognized speech with a language caption. The language-video, language-audio, and language-speech similarities are aggregated before contrastive alignment with a mixture of weights governed by the content of the caption (e.g., the language-video similarity is given more weight if the caption refers to something that is more salient in the video than in the other modalities).

\subsubsection{Object-centric tasks: tracking and object detection}
\label{sec:object_centric_tasks}
Tasks such as object detection, tracking, and segmentation are inherently object-centric in nature, and recent work~\cite{herzig2022object, hwang2021video, yang2022temporally, meinhardt2022trackformer, zheng2022vrdformer} within these tasks have begun to leverage temporally coherent object representations. As object-centric approaches tend to focus on per-object outputs, a large part of the information within a given frame is redundant (as mentioned in Sec. \ref{sec:efficient}), therefore leveraging known and relevant content from previous frames can be used to focus the global attention to object relevant cues. As such, these approaches typically leverage memory or recurrency (as described in \ref{sec:long-term_modeling}) to correlate object information temporally. 
 
In the former recent work~\cite{hwang2021video, yang2022temporally} leverage a set of ''messenger´´ tokens to relay contextual information between frames. IFC-transformer~\cite{hwang2021video} processes the relationship in an isolated encoder, whereas TeViT~\cite{yang2022temporally} shifts the tokens between frame sequences to achieve object-specific information aggregation, to accumulate temporal information across different steps sequentially, resulting in a hierarchical-like approach for temporal information sharing. Other work~\cite{yang2021associating} however, performs both long-term and short-term information sharing in parallel, subsequently concatenated. Due to varied framerate and inter-frame changes in content, smoothness cannot be guaranteed through long-term alone, thus short-term attention is computed on a smaller spatiotemporal neighborhood, to ensure smooth and continuous predictions between frames.
With regards to recurrency, other approaches~\cite{meinhardt2022trackformer,zheng2022vrdformer} leverage object-specific tokens that are derived from an outputted bounding box, and the spatial+size information is then recurrently propogated\cite{meinhardt2022trackformer,zheng2022vrdformer} or used to produce region-specific attention for each concurrently between frames\cite{herzig2022object}. 
Inspired by recent works in Vision transformers (particularly DETR and its variants)~\cite{zheng2022vrdformer, meinhardt2022trackformer}, leverages the bounding box predictions from each frame to augment the decoder queries by concatenating detection tokens from previous frames to the existing learned-fixed tokens, in addition to storing each detection in memory for increased robustness to occlusions in the video sequence. 

As can be observed in vision transformers, architectures that leverage the object features to aggregate contextual information such as \cite{pramono2019hierarchical, herzig2022object} attempt to enhance existing representations with more focus on object-centric information. Where ORViT~\cite{herzig2022object} leverages auxiliary bounding box information in each transformer layer, whereas the GroupFormer~\cite{li2021groupformer} leverages bounding boxes to isolate objects for a separate object-specific action classification branch. Unlike the recurrent and memory style approach these types of approaches don't seem to aim for an efficient design in terms of computational cost, but rather efficient in the sense of information-rich representation, that leverages object-centric information in addition to global context information.

\subsubsection{Low-level tasks}
\label{sec:low_level_tasks}
Given the high dimensionality of video data, video generation tasks are quite challenging, and not many video Transformers try to address them. In particular, \cite{Weissenborn2020Scaling,rakhimov2020latent} tackle future frame prediction, \cite{Weng_2021_ICCV} generates grayscale video from event-based videos and~\cite{zeng2020learning,Liu_2021_ICCV} perform video inpainting. Most of these propose to embed a Transformer within some type of convolutional auto-encoder to evolve representations between encoder and decoder~\cite{Weng_2021_ICCV,zeng2020learning,Liu_2021_ICCV}. The only exception is \cite{Weissenborn2020Scaling}, which performs local attention and generates video autoregressively one pixel channel at a time. Interestingly, \cite{Liu_2021_ICCV} outperforms \cite{zeng2020learning} in all tested benchmarks for inpainting by using an overlapping patch tokenization strategy.

\subsubsection{Segmentation}
\label{sec:segmentation_task}
Most work in segmentation leverage temporal relations to refine intermediate feature representations\cite{wang2021spatiotemporal, wang2021end, Yu_2021_ICCV}. Most notably, \cite{wang2021end} leverages the Transformers' ability to view the entire sequence, to include an auxiliary loss where representations of individuals are matched temporally, effectively teaching the network to implicitly track objects and leverage temporal fine-grained information.
Alternatively, \cite{ye2021referring} leverages a novel word-visual attention mechanism allowing a textual query to attend to specific content in multiple spatial scales and perform segmentation based on the said query.

\subsubsection{Summarization}
\label{sec:summarization_task}
Few works have used Transformers for the task of video summarization by predicting frame-wise importance scores. We find two key trends when solving this task through VTs: the use of RNNs as an initial step~\cite{liu2019learning,sung2020video} and using individual frames to attend to aggregated subsets of the video either from a GRU~\cite{sung2020video} or by using a masked Transformer~\cite{lin2020bi}.

\subsubsection{Other tasks}
\label{sec:other_tasks}
Transformers have also been applied for action anticipation~\cite{Girdhar_2021_ICCV, Wang_2021_ICCV}, sign-language translation~\cite{camgoz2020sign, camgoz2020multi}, visual-question answering~\cite{kim2018multimodal, li2021vidtr}, autonomous driving~\cite{prakash2021multi}, robot navigation~\cite{Fang_2019_CVPR}, visual-language navigation~\cite{Pashevich_2021_ICCV}, personality recognition~\cite{Curto_2021_ICCV}, lip reading~\cite{luo_sychro_2020},  dynamic scene graph generation~\cite{cong2021spatial}, and multimedia recomendation~\cite{chen2019bert4sessrec}. As not many video Transformers have tackled this, it is too early to ascertain specific trends, so we simply list them here for completeness.


\end{document}